\documentclass[letterpaper]{article} 
\usepackage{aaai24}  
\usepackage{times}  
\usepackage{helvet}  
\usepackage{courier}  
\usepackage[hyphens]{url}  
\usepackage{graphicx} 
\usepackage{natbib}  
\usepackage{caption} 
\usepackage{algorithm}
\usepackage{algorithmic}

\usepackage{amsmath}

\usepackage{amssymb}
\usepackage{mathtools}
\usepackage{amsthm}

\usepackage{booktabs}       
\usepackage[export]{adjustbox}
\usepackage{subfigure}
\usepackage{paralist}
\usepackage{multirow}
\usepackage{xcolor}

\usepackage{newfloat}
\usepackage{listings}
\DeclareCaptionStyle{ruled}{labelfont=normalfont,labelsep=colon,strut=off} 
\floatstyle{ruled}
\newfloat{listing}{tb}{lst}{}
\floatname{listing}{Listing}
\title{Composite Active Learning:\\Towards Multi-Domain Active Learning with Theoretical Guarantees}

\author{
    Guang-Yuan Hao\textsuperscript{\rm 1 \rm4},
    Hengguan Huang\textsuperscript{\rm 3},
    Haotian Wang\textsuperscript{\rm 5}\thanks{Work done during PhD at Rutgers.},
    Jie Gao\textsuperscript{\rm 2},
    Hao Wang\textsuperscript{\rm 2} 
}
\affiliations{
    \textsuperscript{\rm 1}Hong Kong University of Science and Technology  
    \textsuperscript{\rm 2}Rutgers University\\
    \textsuperscript{\rm 3}National University of Singapore
    \textsuperscript{\rm 4}Mohamed bin Zayed University of Artificial Intelligence
    \textsuperscript{\rm 5}JD Logistics \\
    guangyuanhao@outlook.com, hengguan.huang@gmail.com, wanghaotian18@jd.com,
     \{jg1555,     hw488\}@cs.rutgers.edu
}

\usepackage{bibentry}

\def\I{{\bf I}}

\def\x{{\bf x}}

\def\z{{\bf z}}

\def\0{{\bf 0}}
\def\1{{\bf 1}}

\def\TM{{\mathcal T}}
\def\UM{{\mathcal U}}

\def\OM{{\mathcal O}}
\def\HM{{\mathcal H}}

\def\LM{{\mathcal L}}
\def\MM{{\mathcal M}}

\def\SM{{\mathcal S}}

\def\alp{\mbox{\boldmath$\alpha$\unboldmath}}

\def\argmax{\mathop{\rm argmax}}

\newtheorem{theorem}{Theorem}
\newtheorem{lemma}{Lemma}

\numberwithin{theorem}{section}
\numberwithin{lemma}{section}
\numberwithin{remark}{section}
\numberwithin{cor}{section}
\numberwithin{proposition}{section}

\newcommand{\tabref}[1]{Table~\ref{#1}}
\newcommand{\secref}[1]{Sec.~\ref{#1}}
\newcommand{\figref}[1]{Fig.~\ref{#1}}
\newcommand{\lemref}[1]{Lemma~\ref{#1}}
\newcommand{\thmref}[1]{Theorem~\ref{#1}}

\newcommand{\eqnref}[1]{Eqn.~\ref{#1}}
\newcommand{\algref}[1]{Alg.~\ref{#1}}

\renewcommand{\hat}{\widehat}
\renewcommand{\frac}{\tfrac}
\begin{document}

\maketitle

\begin{abstract}
Active learning (AL) aims to improve model performance within a fixed labeling budget by choosing the most informative data points to label. 
Existing AL focuses on the single-domain setting, where all data come from the same domain (e.g., the same dataset). 
However, many real-world tasks often involve multiple domains. For example, in visual recognition, it is often desirable to train an image classifier that works across different environments (e.g., different backgrounds), where images from each environment constitute one domain. 
Such a multi-domain AL setting is challenging for prior methods because they (1) ignore the similarity among different domains when assigning labeling budget and (2) fail to handle distribution shift of data across different domains. 
In this paper, 
we propose the first general method, dubbed composite active learning (CAL), for multi-domain AL. 
Our approach explicitly considers the domain-level and instance-level information in the problem; 
CAL first assigns domain-level budgets according to domain-level importance, which is estimated by optimizing an upper error bound that we develop; 
with the domain-level budgets, CAL then leverages a certain instance-level query strategy to select samples to label from each domain. 
Our theoretical analysis shows that our method achieves a better error bound compared to current AL methods. Our empirical results demonstrate that our approach significantly outperforms the state-of-the-art AL methods on both synthetic and real-world multi-domain datasets. Code is available at \url{https://github.com/Wang-ML-Lab/multi-domain-active-learning}.
\end{abstract}

\section{Introduction}\label{sec:intro}


The performance of machine learning models, especially supervised learning ones, largely hinges on the amount of labeled data available. However, in practice, labels are often expensive, tedious, or time-consuming to obtain. 
Active learning (AL) tackles this problem by `actively' choosing the most informative data points (e.g., data with the most uncertain model predictions) to label, thereby achieving higher accuracy with the same labeling budget. 

\begin{figure}
\begin{center}
\vskip -0.0cm
\includegraphics[width=0.55\linewidth]{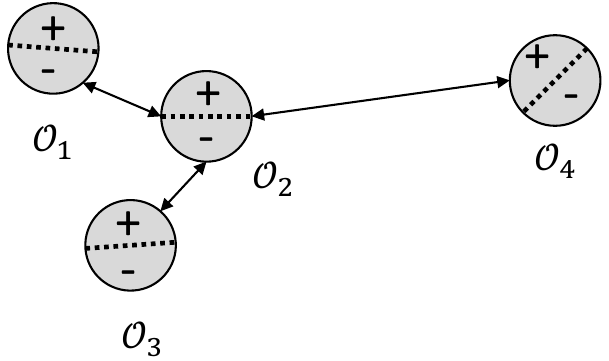} 
\end{center}
\vskip -0.55cm
 \caption{\small Among the four domains, domain $\OM_2$ is the closest to the other three domains. Assigning more labels to this representative domain $\OM_2$ could better generalize to other domains, since similar domains may have similar decision boundaries. }
\label{fig:intro}
\vskip -0.8cm
\end{figure}

Existing AL research mostly focuses on the single-domain setting, where all data come from the same domain, e.g., the same dataset. 
In practice, however, real-world tasks often require actively querying labels among multiple domains with the same input space. For example, to train an object recognition model that detects and classifies wildlife animals in different environments, where images from each environment constitute one domain, one needs to carefully decide how to spend the labeling budget among the different domains to achieve the highest average accuracy across all domains. 
In such cases, it is sub-optimal to directly perform single-domain AL separately for each domain. 
The reasons are two-fold. (1) \emph{Domain similarity}: 
Single-domain AL fails to effectively consider the similarities of different domains when considering cross-domain generalization and assigning labeling budget. In the wildlife detection example, it would be more cost-effective to spend more computing resources and more budget on representative environments, i.e., domains that are more similar to other domains, as the improvement on such domains could better generalize to other domains, {as shown in \figref{fig:intro}}.
(2) \emph{Distribution shift}: Single-domain AL fails to handle the distribution shift of data across different domains 
because it tends to learn domain-specific features instead of domain-invariant features, leading to poor cross-domain generalization and rendering the query strategies less effective. 

Therefore the key challenges for effective multi-domain AL are to estimate/incorporate domain similarity and to handle distribution shift. 
Our approach for addressing both challenges starts by constructing a \emph{surrogate domain} for each original domain. Specifically, a surrogate domain consists of a weighted collection of all labeled data points from all domains, where the weights reflect the similarity among different domains. For example, the weight between two similar domains is expected to be higher than that between two distant domains. 
To estimate and incorporate domain similarity, we develop an upper bound on the average error of all domains and then estimate the similarity weights among the domains 
by minimizing this bound. 
To handle distribution shift, we use the same upper bound as the loss function to learn an encoder that maps input data from different domains into an aligned feature space, thereby reducing the distribution shift. 

Specifically, our theoretical analysis shows that: (1) by jointly estimating the similarity weights and learning the encoder, one can minimize the upper bound for the average error among all domains; (2) the optimal strategy is achieved when each domain's labeling budget is proportional to its total similarity weight, i.e., the sum of  similarity weights from one domain to all $N$ domains. 
We summarize our contributions as follows:
\begin{compactitem}
\item We propose Composite Active Learning (CAL) as the first general deep AL method {to take into consideration both domain-level and instance-level information} for addressing the problem of multi-domain active learning. 
\item We analyze our method and provide theoretical guarantees that CAL with our budget assignment strategy achieves a better upper bound on the average error of all domains. 

\item We provide empirical results on both synthetic and real-world datasets with detailed ablation studies, showing that CAL significantly improves performance over the state of the art for multi-domain active learning.
\end{compactitem}





\section{Related Work}\label{sec:related}
\textbf{Active Learning.} 
There is rich literature on active learning \cite{al_survey, al_survey2, al_learn_alg, al_bayesian, al_dis_theory, al_agnostic, active_change}. Typically they apply a query strategy to find the most informative unlabeled samples to label, thereby achieving improved accuracy given a fixed labeling budget. Common query strategies include uncertainty-based strategies to choose data with high uncertainty \cite{margin, al_influence,  al_svm, al_cls, al_new, al_less, al_semi, al_adversarial}, density-based methods to choose representative samples \cite{al_corset, al_variational, al_longtail, al_align0, al_align1, al_align2}, and hybrid approaches to balance uncertainty and diversity of chosen samples \cite{badge, cluster-margin, al_online, al_learn, al_info}. A few early works are related to both active learning and multi-domain learning for specific applications such as text classification~\cite{mudal_text} and recommend system~\cite{mudal_recommendation}. However, they are limited to linear methods and need careful feature engineering; therefore they are not applicable to our general setting that often involves highly nonlinear data and deep learning (see {Sec. 3 in the Supplement} for detailed discussion and results). 
Different from these previous works, our method focuses on the general multi-domain active learning setting. We also note that our method does not assume specific \emph{instance-level} query strategies and is therefore \emph{compatible with (and orthogonal to) any previous AL methods (that is, our proposed framework can be used to extend any single-domain AL methods to the multi-domain setting)}, as shown in later sections. 

\textbf{(Active) Domain Adaptation.} 
Among prior work on domain adaptation (a different problem setting), most relevant to our work is domain adaptation methods that leverage domain relations, e.g., domain adaptation across continuously indexed domains~\cite{cida}, graph-relational domains~\cite{GRDA}, taxonomy-structured domains~\cite{TSDA}, incremental domains~\cite{UDIL}, and domains with unknown domain indices (to be inferred from data)~\cite{VDI}. 
A few recent works 
utilize AL to improve performance in a target domain~\cite{al4da_0,al4da_1,al4da_2,al4da_3}. 
Their key ideas are to first perform domain adaptation to match the distributions of the source and target domains, and then apply query strategies to select useful unlabeled samples from the target domain. 
Here we note several key differences between active domain adaptation and multi-domain AL. 
(1) Active domain adaptation distinguishes between source domains and target domains, while multi-domain AL does not. 
(2) Active domain adaptation assumes access to all labels in the source domains even from the beginning, while multi-domain AL starts with only unlabeled data in all domains (except that in AL's initial round, one would randomly sample a few data points to label).
(3) Active domain adaptation aims to improve performance (e.g., accuracy) only on the target domain, while multi-domain AL aims to improve the average performance of all domains. 
Such differences preclude its direct application to our multi-domain active learning settings; in~\secref{sec:experiments}, we show that even after careful adaptation to our setting, active domain adaptation often underperform even naive active learning baselines.

\section{Methodology}


In this section, we review the basics of single-domain AL, and formalize the \textbf{MU}lti-\textbf{D}omain \textbf{A}ctive \textbf{L}earning (\textbf{MUDAL}) setting. We then revisit our key ideas in~\secref{sec:intro} and describe our methods. 

\subsection{Preliminaries: Single-Domain AL}
Existing AL methods typically consider a single-domain setting and optimize performance by using a \emph{query strategy} to choose data points to label, within a given budget. 
Specifically, an AL model is 
trained in $R+1$ rounds with a total labeling budget of $M=m_0 + R\times m$. In this process, $m_0$ data points, randomly sampled from the unlabeled training set, are labeled in the initial round. In each of the following $R$ rounds during the query stage, the encoder extracts features from all unlabeled data. A query strategy then utilizes these features to select and label additional $m$ samples.

\begin{figure*}[t]
\begin{center}
\vskip -0.5cm
\includegraphics[width=0.85\linewidth]{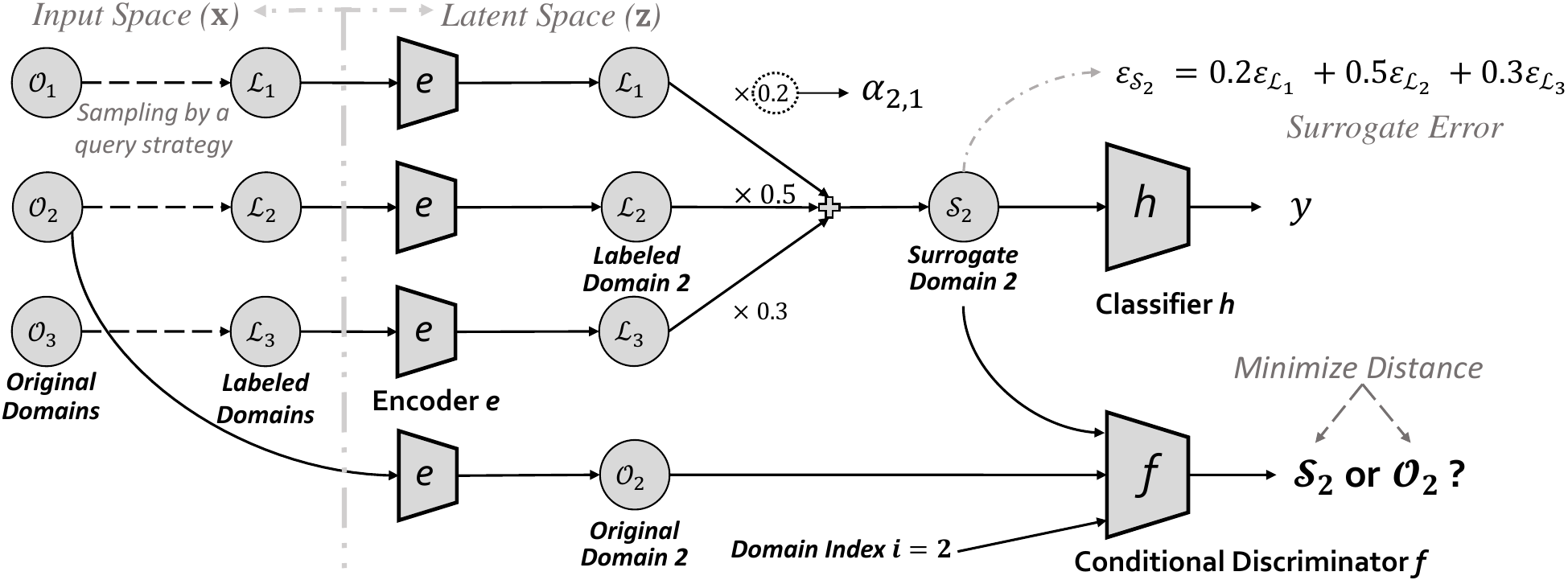} 
\end{center}
\vskip -0.4cm
 \caption{\textbf{ \small Overview of our proposed framework.} \textbf{Left:} $3$ labeled domains and $3$ original domains {in the input space}. \textbf{Right:} {In the latent space}, CAL constructs surrogate domain $\SM_2$ using $\alpha$-weighted labeled domains $\SM_2(Z)=\sum\nolimits_{j=1}^3 \alpha_{2,j}\LM_j(Z)$, and estimate similarity weights $\alpha_{2,j}$ by minimizing the distance between surrogate domain $\SM_2$ and its original domain $\OM_2$. {All encoders $e$ share the same parameters. The encoder $e$, domain similarity $\{\alpha_{2,j}|j=1,2,3\}$, and conditional discriminator $f$ play a min-max game to reduce the distance between $\SM_2$ and $\OM_2$ by joint similarity estimation and feature alignment. The classifier $h$ is trained on all surrogate domains $\SM_1$, $\SM_2$ and $\SM_3$. For clarity, we omit surrogate domains $\SM_1$ and $\SM_3$. } }
\label{fig:insight}
\vskip -0.6cm
\end{figure*}

\subsection{Multi-Domain AL}\label{sec:mudal}

\textbf{Notation.} 
With the input and labels denoted as $\x$ and $y$, respectively, we denote the $N$ \emph{original domains}' associated input data distributions as $\{\mathcal{O}_i(X)\}_{i=1}^N$ and the corresponding joint input-label distributions as $\{\mathcal{O}_i(X, Y)\}_{i=1}^N$; all domains share the same input space and label space. We assume a shared encoder $e$ and a shared classifier $h$ for all domains. We denote as $\z=e(\x)$ the feature extracted by the encoder and $\{\mathcal{O}_i(Z)\}_{i=1}^N$ the $N$ domains' feature distributions. Accordingly, we use $X$, $Y$, and $Z$ to denote random variables. We treat the already labeled data points as separate domains (more details in~\secref{sec:overview}) and therefore have $N$ additional \emph{labeled domains} with feature distributions $\{\mathcal{L}_j(Z)\}_{j=1}^N$.  
Similarly we have $N$ additional \emph{surrogate domains} $\{\mathcal{S}_i(Z)\}_{i=1}^N$, each of which is a mixture of the $N$ labeled domains. With slight notation overload, we use $\OM_i$, $\SM_i$, $\LM_j$ to denote original, surrogate, and labeled domains, respectively. 

\textbf{Problem Setting.} 
We assume a two-step procedure for MUDAL. 
Specifically, with $R+1$ training rounds, denote 
{$M = m_0 + R \times m$
as the total budget, where $m =\sum_j m_j^{(r)}$ ($m_j^{(r)}$ denotes the sub-budget for domain $j$ in round $r$).} In Round $0$, $\frac{m_0}{N}$ instances are randomly sampled from each domain as the initial training set. Each remaining round $r$ consists of two steps: (1) \emph{domain-level selection}, where a MUDAL algorithm decides on an additional sub-budget $m_j^{(r)}$ for {each} domain $j$, and  (2) \emph{instance-level selection}, where the algorithm applies a query strategy to choose $m_j^{(r)}$ data points in {each} domain $j$ to label. 
The goal is to estimate the optimal sub-budget $m_j^{(r)}$ such that the average classification error among all domains can be minimized. 
To facilitate analysis below, we further define $\beta_j^{(r)}$ as domain $j$'s proportion of the total budget until round $r$, i.e., $\beta_j^{(r)}=\frac{m_0/N+\sum_{k=1}^r m_j^{(k)}}{m_0+r\times m}$. 

\subsection{{Method Overview}}\label{sec:overview}

Below we provide an overview for a \emph{simplified} version of CAL 
{including domain-level and instance-level selection}, 
using~\figref{fig:insight} as a running example. 
\textbf{Constructing Surrogate Domains.} 
A labeled domain cannot fully represent its original domain; this is because labeled data are not I.I.D. samples from the original domain due to query strategies. 
Therefore in each round $r$, 
our approach starts by constructing a \emph{surrogate domain} for each original domain. Specifically, a surrogate domain consists of a weighted collection of all labeled domains (defined in~\secref{sec:mudal}), where the weights reflect the similarity among different domains. \figref{fig:insight}(left) shows an example with $N=3$ domains {in the input space}. Correspondingly we have $3$ labeled domains $\{\mathcal{L}_j\}_{j=1}^3$, where data are already {sampled by query strategies} and labeled in the previous rounds during AL, and $3$ original domains $\{\mathcal{O}_i\}_{i=1}^3$, which include both labeled and unlabeled data 
(i.e., the size of an original domain remains constant across rounds). 
For each original domain $\OM_i(Z)$ in the latent space induced by the encoder $e$, e.g., $\OM_2(Z)$ at the bottom of \figref{fig:insight}(right), 
we construct a surrogate domain $\SM_i(Z)$ as a mixture of all labeled domains, i.e., $\SM_i(Z)=\sum\nolimits_{j=1}^3 \alpha_{i,j}\LM_j(Z)$, where $\alpha_{i,j}\geq 0$ and $\sum_j \alpha_{i,j}=1$.  \figref{fig:insight}(right) shows an example of constructing surrogate domain $\SM_2$ {for original domain $\OM_2(Z)$} as $\SM_2(Z)=0.2\LM_1(Z)+0.5\LM_2(Z)+0.3\LM_3(Z)$.
Accordingly, {with classifier $h$ taking $\SM_2(Z)$ as input,} $\epsilon_{\SM_2}=0.2\epsilon_{\LM_1}+0.5\epsilon_{\LM_2}+0.3\epsilon_{\LM_3}$ is the surrogate error to approximate original domain $\OM_2$'s error.

\textbf{Domain Similarity (Importance) Estimation.} 
Each surrogate domain $\SM_i$ is a similarity-weighted sum of the $N$ labeled domains $\{\LM_j\}_{j=1}^N$. Given an encoder $e$, we can then estimate the similarity weight $\alpha_{i,j}$ between labeled domain $\LM_j$ and original domain $\OM_i$, by minimizing the distance between the original domain feature distribution $\OM_i(Z)$ and the surrogate domain feature distribution $\SM_i(Z)$, denoted as $d(\OM_i(Z),\SM_i(Z))$.  For example, in~\figref{fig:insight}(right) we estimate the similarity weights $\{\alpha_{2,j}\}_{j=1}^3$ by minimizing the distance $d(\OM_2(Z),\SM_2(Z))$, where $\SM_2(Z)=\sum\nolimits_{j=1}^3 \alpha_{2,j}\LM_j(Z)$. 

\textbf{Distribution-Shift Reduction with Feature Alignment.} 
To reduce distribution shift across domains, we propose to learn an encoder $e$ such that feature distributions from different domains can be aligned. Suppose the similarity weights $\{\alpha_{i,j}\}$ are given, we search for an encoder $e$ that minimizes the distance $d(\OM_i(Z),\SM_i(Z))=d(\OM_i(e(X)),\SM_i(e(X)))$. 
Note that this is different from adversarial domain adaptation~\cite{DANN} which directly aligns different $\OM_i(Z)$'s; our preliminary results show that such direct alignment does not improve, and sometimes even hurts performance.  



\textbf{Joint Similarity Estimation and Feature Alignment.} 
Finally, we perform joint optimization combining both similarity estimation (which learns $\alpha_{i,j}$) and feature alignment (which learns the encoder $e$). {As shown in \figref{fig:insight}, The encoder $e$, similarity weights, and conditional discriminator $f$ play a min-max game to reduce the distance between surrogate domain $\SM_2$ and its original domain $\OM_2$ in the latent (feature) space with joint optimization.} Further theoretical analysis shows that such joint optimization provides a tighter upper bound for the average classification error of all domains (more details in~\secref{sec:full} and~\secref{sec:theory}). 


\textbf{Domain-Level Selection as Sub-Budget Decision Using Similarity Weights.} 
With the estimated similarity $\alpha_{i,j}$, we then assign budget such that
representative domains, i.e., domains that are more similar (important) to other domains (larger $\sum_{i=1}^N \alpha_{i,j}$), will receive more labeling budget. 

\textbf{Instance-Level Selection.} With the assigned domain-level budget, we then apply an instance-level query strategy (e.g., uncertainty-based strategy~\cite{al_survey}) to select data points to label. 

\textbf{Remark: Domain-Level Selection Is Agnostic to Instance-Level Selection.}  
Domain-level selection is agnostic to, and therefore compatible with, any instance-level selection methods. 
We consider such independence as an advantage of our CAL framework; any existing instance-level selection method, such as Margin \cite{margin} and BADGE \cite{badge}, can be used as a sub-routine for CAL to work with our proposed domain-level selection algorithm. Therefore, CAL's key innovation is in (1) the first general two-level framework for multi-domain AL and (2) the first general domain-level selection method, rather than inventing new instance-level selection methods. 

\subsection{Composite Active Learning}\label{sec:full}

\textbf{Upper Bound for the Average Error of All Domains.} 
We are now ready to present our full method, dubbed Composite Active Learning (CAL), to jointly estimate similarity weights, perform feature alignment, and assign sub-budget across domains.
Formally, CAL tries to minimize an upper bound for the average error of all original domains $\frac{1}{N}\sum_i \epsilon_{\OM_i}(h\circ e)$.

Our bound consists of three terms: \textbf{(i)} an $\alpha_j$-weighted average of errors in all labeled domains $\sum\nolimits_j \alpha_{j} \epsilon_{\mathcal{L}_j}(h \circ e)$, 
where $\alpha_{j}= \frac{1}{N}\sum_i \alpha_{i, j}$ is the summary similarity score for labeled domain $\LM_j$, $h(\cdot)$ is the classifier, and $\epsilon_{\mathcal{L}_j}(h \circ e)$ is the expected classification error in labeled domain $\LM_j$ given $h(\cdot)$ and $e(\cdot)$; \textbf{(ii)} the average $\mathcal{H}\Delta \mathcal{H}$-distance~\cite{domain_theory} between each original domain $\OM_i$ and its surrogate domain $\SM_i$, denoted as $\frac{1}{N}\sum\nolimits_i \frac{1}{2} d_{\mathcal{H}\Delta \mathcal{H}}(\mathcal{O}_i(e (X)), \mathcal{S}_i(e(X)))$; \textbf{(iii)} an $h$-agnostic term $\frac{1}{N}\lambda_i$, where $\lambda_{i} = \min_{h_i} (\epsilon_{\mathcal{O}_i}(h_i\circ e)+ \epsilon_{\mathcal{S}_i}(h_i\circ e))$ and $h_i$ is the $i$-th domain-specific classifier; $\epsilon_{\mathcal{O}_i}(h_i\circ e)$ and $\epsilon_{\mathcal{S}_i}(h_i\circ e))$ are the expected classification error in original domain $\OM_i$ and surrogate domain $\SM_i$, respectively. We summarize the upper bound below (proofs in~\secref{sec:theory} and~{Sec. 2 of the Supplement}): 
\begin{align}
 \min_{h, e} \min_{\alp} 
\sum\nolimits_j \alpha_{j} \epsilon_{\mathcal{L}_j}(h \circ e) +  \nonumber\\ 
\frac{1}{N}\sum\nolimits_i \frac{1}{2} d_{\mathcal{H}\Delta \mathcal{H}}(\mathcal{O}_i(e (X)), \mathcal{S}_i(e(X))) 
+ \frac{1}{N} \sum\nolimits_i \lambda_{i},
\label{eq:cadol}
\end{align} 
where $i$ indexes original domains and surrogate domains while $j$ indexes labeled domains. $\alp = [\alpha_{i, j}]_{i=1, j=1}^{N, N}$ is a similarity matrix containing the similarity weights. Note that by the definition of surrogate domains, the first term $\sum\nolimits_j \alpha_{j} \epsilon_{\mathcal{L}_j}(h \circ e)=\frac{1}{N} \sum\nolimits_i \epsilon_{\mathcal{S}_i}(h \circ e)$. 
The surrogate $\mathcal{S}_i(e(X)) = \sum_j \alpha_{i, j} \mathcal{L}_j(e(X))$ is a weighted average of all labeled domains; therefore $d_{\mathcal{H}\Delta \mathcal{H}}(\mathcal{O}_i(e (X)), \mathcal{S}_i(e(X)))$ \emph{connects the labeled domain error, which can be computed, to the original domain error, which we want to minimize, through our constructed surrogate domains $\SM_i$}.

\textbf{Objective Function.} The three terms in \eqnref{eq:cadol} lead to a minimax game as our objective function. 

\textbf{The first term} $\sum\nolimits_j \alpha_{j} \epsilon_{\mathcal{L}_j}(h \circ e)$, equal to $\frac{1}{N}\sum\nolimits_i \epsilon_{\mathcal{S}_i}(h \circ e)$, in~\eqnref{eq:cadol} leads to the first term of our objective function: 
\begin{align}
V_h(h, e, \alp) & = \frac{1}{N} \sum\nolimits_{i=1}^N \mathbb{E}^{\mathcal{S}_i} [L_Y(h \circ e(\x), y)]\nonumber\\ 
&  =  \sum\nolimits_{j=1}^N  \alpha_{j} \mathbb{E}^{\mathcal{L}_j} [L_Y(h \circ e(\x), y)],
\end{align}
where {$\mathbb{E}^{\mathcal{S}_i}$ and $\mathbb{E}^{\mathcal{L}_j}$ denote the expectation over the data distribution of $(\x,y)$ in surrogate domain $\SM_i$ and labeled domain $\LM_j$ respectively} 
and $L_Y$ is the cross-entropy loss for multi-class classification. {This term indicates that classifier $h$ is shared by and trained on all surrogate domains.}

\textbf{The second term} $\frac{1}{N}\sum\nolimits_i \frac{1}{2} d_{\mathcal{H}\Delta \mathcal{H}}(\mathcal{O}_i(e (X)), \mathcal{S}_i(e(X)))$ in~\eqnref{eq:cadol} leads to the second term of our objective function: 
\begin{align}
V_d(f, e, \alp) =  \frac{1}{2N}\sum\nolimits_{i=1}^N  \{\mathbb{E}^{\mathcal{O}_i} [L_D(f(e(\x), i), 1)]
\nonumber\\ 
+ \sum\nolimits_{j=1}^N \alpha_{i,j} \mathbb{E}^{\mathcal{L}_j} [L_D(f(e(\x), i), 0)]\}, \label{eq:v_d}
\end{align}
where $\mathbb{E}^{\mathcal{O}_i}$ denotes expectation over the data distribution of $(\x,y)$ in original domain $\OM_i$. $L_D$ is the cross-entropy loss for binary classification. $f$ is a conditional feature discriminator that takes the feature $e(\x)$ and the domain index $i$ as input and classifies whether $e(\x)$ comes from original domain $\OM_i$ or surrogate domain $\SM_i=\sum_j\alpha_{i,j}\LM_j$. Note that $\min_f V_d(f, e, \alp)$ measures how well the discriminator can distinguish between original domain $\OM_i$ and surrogate domain $\SM_i=\sum_j\alpha_{i,j}\LM_j$, and therefore connects to the distance term $\frac{1}{N}\sum\nolimits_i \frac{1}{2} d_{\mathcal{H}\Delta \mathcal{H}}(\mathcal{O}_i(e (X)), \mathcal{S}_i(e(X)))$ in~\eqnref{eq:cadol}. 

\textbf{The third term} $\frac{1}{N} \sum\nolimits_i \lambda_{i}$ in~\eqnref{eq:cadol} leads to the third term of our objective function: 
\begingroup\makeatletter\def\f@size{8.4}\check@mathfonts
\def\maketag@@@#1{\hbox{\m@th\normalsize\normalfont#1}}%
\begin{align}
& V_{\lambda}(\{h_i\}_{i=1}^N, e, \alp) = \frac{1}{N}\sum\nolimits_{i=1}^N \mathbb{E}^{\mathcal{S}_i} [L_Y(h_i \circ e(\x), y)] \nonumber\\
& = \frac{1}{N}\sum\nolimits_{i=1}^N \sum\nolimits_{j=1}^N \alpha_{i, j} \mathbb{E}^{\mathcal{L}_j} [L_Y(h_i \circ e(\x), y)], \label{eq:v_lambda}
\end{align}
\endgroup
which is an upper bound of $\frac{1}{N} \sum_i \lambda_{i}$ in~\eqnref{eq:cadol} (more details in {Sec. 1.2 in the Supplement}). 
{Different from the classifier $h$ trained on all surrogate domains, $h_i$ is the domain-specific classifier which is trained only on surrogate domain $\SM_i$ (note that $h_i$ is omitted in \figref{fig:insight} for clarity).} 

\textbf{Putting them all together}, \eqnref{eq:cadol} corresponds to the following minimax game as the final objective function: 
\begin{align}
\min_{h, e, \{h_i\}_{i=1}^N} \min_{\alp} \max_{f} 
V_h(h, e, \alp) 
- \lambda_d V_d(f, e, \alp) \nonumber\\ 
+ V_{\lambda}(\{h_i\}_{i=1}^N, e, \alp),
\label{eq:objective}
\end{align}
{where $\lambda_d$ is a hyperparameter to balance loss terms (the performance of the classifier and feature alignment in particular), since too heavy penalization over the feature-level misalignment may harm the performance. In all experiments, $\lambda_d$ is set to $1$, which already achieves good performance.} 

\textbf{Assigning Sub-Budget for Each Domain.} 
After training converges in round $r$, we then assign budget such that domain $j$'s proportion of the total budget until round r,  $\beta_j^{(r)}\propto\sum_{i=1}^N \alpha_{i,j}$. In \secref{sec:theory}, we prove that this is the optimal sub-budget that minimizes the error bound.

\subsection{Enhancing CAL with Augmented Instance-Level Acquisition}
We propose a new instance-level strategy, dubbed \textbf{Gra}dient with \textbf{D}iscriminator \textbf{S}core (GraDS), 
which combines domain-level information with the gradient-based method, BADGE, to enhance our CAL. We choose to build GraDS on BADGE, since we found that combining CAL with BADGE is more effective than other strategies (more results in the Supplement). 
Our GraDS incorporates into BADGE the computed ``outlier score'' (i.e., our discriminator's output probability $f(Z, i)=\frac{\OM_i(Z)}{\OM_i(Z)+\SM_i(Z)}$), which evaluates how much a data point looks like an outlier to the surrogate domain and therefore needs to be labeled in the next round. Multiplying the BADGE gradient with the outlier score $f(Z, i)$ leads to a revised gradient, which is then used with k-means++ \cite{k-means++} to select samples for labeling. GraDS prioritizes data points with higher outlier score ($f(Z, i)$) and classifier uncertainty (BADGE gradient) in each round to 
reduce classification error in the original domain. See more details on GraDS in Sec. 1.5 of the Supplement.

\section{Theoretical Analysis}\label{sec:theory}
In this section, we provide theoretical analysis for CAL. We first provide an upper bound for the error of one original domain in~\lemref{lem:bound_one}, based on which we develop the upper bound for the average error over all original domains in~\thmref{thm:bound_all}. We then prove the optimality of CAL's sub-budget assignment strategy in~\thmref{thm:beta}. {\textbf{All proofs are in {Sec. 2 of the Supplement}.}} 



\lemref{lem:bound_one} below provides theoretical guarantees that connect the prediction errors of $\SM_i$ and $\OM_i$. 





\begin{lemma}[\textbf{Error Bound for One Domain}]\label{lem:bound_one}
Let $\mathcal{H}$ be a hypothesis space, and $h$, $h_i$ $\in \mathcal{H}$ : $\mathcal{Z} \xrightarrow{} [0, 1]$. $\mathcal{O}_i(Z)$ is the feature distribution of original domain $i$, and its surrogate domain $\mathcal{S}_i(Z)= \sum_j \alpha_{i, j} \mathcal{L}_j(Z)$ is a weighted average of $N$ labeled domains $\{\mathcal{L}_j(Z)\}_{j=1}^N$. With the surrogate error $\epsilon_{\mathcal{S}_i}(h)=\sum_j \alpha_{i, j} \epsilon_{\mathcal{L}_j}(h)$ and $\lambda_{i} = \min_{h_i} (\epsilon_{\mathcal{O}_i}(h_i)+ \epsilon_{\mathcal{S}_i}(h_i))$, we have:
\begingroup\makeatletter\def\f@size{9}\check@mathfonts
\def\maketag@@@#1{\hbox{\m@th\normalsize\normalfont#1}}%
\begin{align*}
\epsilon_{\mathcal{O}_i}(h) 
& \leq \epsilon_{\mathcal{S}_i}(h) 
+ \frac{1}{2}d_{\mathcal{H}\Delta \mathcal{H}}(\mathcal{O}_i(Z), \mathcal{S}_i(Z) ) 
+ \lambda_{i} \nonumber\\ 
& =\sum_{j} \alpha_{i, j} \epsilon_{\mathcal{L}_j}(h) 
+ \frac{1}{2}d_{\mathcal{H}\Delta \mathcal{H}}(\mathcal{O}_i(Z), \sum_{j} \alpha_{i, j} \mathcal{L}_j(Z))
+ \lambda_{i}
\end{align*}
\endgroup
\end{lemma}

Based on~\lemref{lem:bound_one}, \thmref{thm:bound_all} upper bounds the average error of all original domains. 


\begin{theorem}[\textbf{Error Bound for All Domains}]
\label{thm:bound_all}
Let $\mathcal{H}$ be a hypothesis space of VC dimension $d$ and $h, h_i \in \mathcal{H}: \mathcal{Z} \xrightarrow{} [0, 1]$ be any hypothesis in $\mathcal{H}$. If labeled domains contain $\MM$ data points in total, with $\beta_j \MM$ assigned to labeled domain $j$, then
for any $\delta \in (0 , 1)$, with probability at least $1-\delta$:
\begingroup\makeatletter\def\f@size{8.4}\check@mathfonts
\def\maketag@@@#1{\hbox{\m@th\normalsize\normalfont#1}}%
\begin{align}
 \frac{1}{N}\sum\nolimits_i\epsilon_{\mathcal{O}_i}(h) \leq
 \sum\nolimits_j \alpha_j \epsilon_{\mathcal{L}_j}(h)  + \frac{1}{2N}\sum\nolimits_i d_{\mathcal{H}\Delta \mathcal{H}}(\mathcal{O}_i(Z), \mathcal{S}_i(Z) ) \nonumber\\
 + \frac{1}{N} \sum\nolimits_i \lambda_{i} 
= \UM 
\leq \UM_E\label{eq:bound_cadol}
\end{align}
\endgroup
where $\UM_E$ is a further upper bound involving the empirical error $\sum_j \alpha_j \hat{\epsilon}_{\mathcal{L}_j}(h)$:
\begingroup\makeatletter\def\f@size{7.8}\check@mathfonts
\def\maketag@@@#1{\hbox{\m@th\normalsize\normalfont#1}}%
\begin{align}
\UM_E
= \sum_j \alpha_j \hat{\epsilon}_{\mathcal{L}_j}(h) +2\sqrt{(\sum_j \frac{\alpha_j^2}{\beta_j})(\frac{2d\log(2(\MM+1)+\log(\frac{4}{\delta}))}{\MM})} \nonumber\\
+ \frac{1}{2N}\sum_i d_{\mathcal{H}\Delta \mathcal{H}}(\mathcal{O}_i(Z), \mathcal{S}_i(Z)) + \frac{1}{N} \sum_i \lambda_{i},\label{eq:ue}
\end{align}
\endgroup
where $\alpha_j= \frac{1}{N}\sum_i \alpha_{i, j}$, $\mathcal{S}_i(Z) = \sum_j \alpha_{i, j} \mathcal{L}_j(Z)$, $ \sum_j \alpha_{i, j}=1$, and $\lambda_{i} = \min_{h_i} (\epsilon_{\mathcal{O}_i}(h_i)+ \epsilon_{\mathcal{S}_i}(h_i))$.
\end{theorem}


With the upper bound $\UM$ in \eqnref{eq:bound_cadol} above, one can tighten the upper bound using $\min_{h, e} \min_{\alp} ~\mathcal{U}$, leading to CAL's objective function in~\eqnref{eq:objective}. 

Note that in each round of multi-domain AL, domain $j$'s accumulated budget ratio is $\beta_j$.
Therefore one can search for the optimal $\beta_j$ that minimizes the error bound in \eqnref{eq:ue}. \thmref{thm:beta} below shows that the optimal sub-budget strategy is achieved when $\beta_j=\alpha_j=\frac{1}{N}\sum_i\alpha_{i,j}$. 


\begin{theorem}[\textbf{Optimal Budget Assignment}]\label{thm:beta}
Assuming $\alpha_j>0$ and $\beta_j>0$ for $j=1, 2, ..., N$, with $\sum_j \alpha_j=1$ and $\sum_j \beta_j=1$. The optimal upper bound for the average error of all domains, i.e., $\UM_E$ in~\eqnref{eq:ue}, is achieved when $\beta_j=\alpha_j=\frac{1}{N}\sum_i\alpha_{i,j}$ for $j=1, 2, ..., N$. 
\end{theorem}

Intuitively the optimal sub-budget assignment $\beta_j=\frac{1}{N}\sum_i\alpha_{i,j}$ implies that representative domains that are more similar (important) to other domains will receive a greater labeling budget. 



\section{Experiments}\label{sec:experiments}



\newcommand{\tabincell}[2]{\begin{tabular}{@{}#1@{}}#2\end{tabular}}
\begin{table*}[h]
\setlength{\tabcolsep}{2.5pt}
\caption{\small RotatingMNIST results ($\%$). ``Joint'' and ``Separate'' indicate joint and separate assignment, respectively (see details in \secref{sec:baselines}). We mark the best results with \textbf{bold face}.}
\label{tab:RotatingMNIST}
\scriptsize
\begin{center}
\vskip -0.5cm
\begin{tabular}{l|cc|cc|cc|cc|cc|cc|c}
\toprule[1.5pt]
\multirow{2}*{Query} & \multicolumn{2}{c|}{Random} & \multicolumn{2}{c|}{Margin} & \multicolumn{2}{c|}{BADGE} & \multicolumn{2}{c|}{Cluster-Margin} & \multicolumn{2}{c|}{Energy} & \multicolumn{2}{c|}{BvSB-DA} & \multirow{2}*{CAL (Ours)} \\ 
\cmidrule(r){2-13} 
 & Joint & Separate  & Joint & Separate  & Joint & Separate & Joint & Separate & Joint & Separate & Joint & Separate &   \\ \midrule
Round 0       &49.3&49.2&49.3&49.2&49.3&49.2&49.3&49.2&49.3&49.2&47.1&47.1&\textbf{51.7}\\
Round 1       &59.3&58.4&59.9&59.3&59.6&61.0&59.7&58.7&59.8&59.8&58.3&57.6&\textbf{71.8}\\
Round 2       &65.2&64.5&65.4&65.8&65.9&66.2&66.0&65.6&65.5&64.7&62.8&63.2&\textbf{81.0}\\
Round 3       &69.4&68.8&69.7&70.6&69.7&71.4&70.1&70.0&69.7&69.2&68.0&69.0&\textbf{85.7}\\
Round 4       &77.7&77.7&80.5&80.3&79.2&80.5&80.2&80.6&79.2&80.1&74.1&74.2&\textbf{87.8}\\
Round 5       &79.9&80.5&81.9&83.1&82.5&83.1&82.2&83.3&82.1&82.9&77.4&76.9&\textbf{87.9}\\ \midrule
Average       &66.8&66.5&67.8&68.1&67.7&68.6&67.9&67.9&67.6&67.6&64.6&64.6&\textbf{77.7}\\ 
\bottomrule[1.5pt]
\end{tabular}
\end{center}
\vskip -0.6cm
\end{table*}


\begin{table*}[h]
\vskip 0.1cm
\setlength{\tabcolsep}{2.5pt}
\caption{\small Office-Home results ($\%$). {``Joint'' and ``Separate'' indicate joint and separate assignment, respectively (see details in \secref{sec:baselines}).} We mark the best results with \textbf{bold face}.}
\label{tab:Office-Home}
\scriptsize
\begin{center}
\vskip -0.5cm
\begin{tabular}{l|cc|cc|cc|cc|cc|cc|cc}
\toprule[1.5pt]
\multirow{2}*{Query} & \multicolumn{2}{c|}{Random} & \multicolumn{2}{c|}{Margin} & \multicolumn{2}{c|}{BADGE} & \multicolumn{2}{c|}{Cluster-Margin} & \multicolumn{2}{c|}{Energy} & \multicolumn{2}{c|}{BvSB-DA} & \multirow{2}*{CAL (Ours)} \\ 
\cmidrule(r){2-13} 
 & Joint & Separate  & Joint & Separate  & Joint & Separate & Joint & Separate & Joint & Separate & Joint & Separate &   \\ \midrule
Round 0       &32.8&32.8&32.8&32.8&32.8&32.8&32.8&32.8&32.8&32.8&31.0&31.0&\textbf{34.5}\\
Round 1       &42.0&42.8&41.3&41.6&42.1&42.6&41.4&42.8&41.5&41.2&37.6&38.0&\textbf{46.3}\\
Round 2       &48.3&48.6&47.4&46.6&48.3&48.1&46.4&47.0&47.7&48.0&44.6&44.8&\textbf{52.8}\\
Round 3       &52.4&52.6&50.5&51.3&51.9&52.0&50.0&50.6&51.4&51.2&48.3&48.3&\textbf{57.4}\\
Round 4       &56.9&57.6&55.2&55.9&57.3&57.1&54.9&55.9&56.2&56.0&51.4&53.2&\textbf{59.6}\\
Round 5       &57.7&58.0&55.4&57.7&57.7&56.9&56.0&57.5&58.1&58.4&52.9&53.9&\textbf{60.9}\\ \midrule
Average       &48.4&48.7&47.1&47.6&48.3&48.2&46.9&47.8&48.0&47.9&44.3&44.9&\textbf{51.9}\\ 
\bottomrule[1.5pt]
\end{tabular}
\end{center}
\vskip -0.8cm
\end{table*}

\subsection{Baselines and Implementations}\label{sec:baselines}

We compared our CAL with six AL baselines, including \textbf{Random} \cite{al_survey}, \textbf{Margin} \cite{margin}, \textbf{BADGE} \cite{badge}, \textbf{Cluster-Margin}~\cite{cluster-margin}, \textbf{Energy} \cite{al4da_energy}, and \textbf{BvSB-DA} \cite{mudal_survey}. Among them, \textbf{Random} is a simple strategy that randomly selects data points to label. \textbf{Energy} is a state-of-the-art hybrid query strategy for active domain adaptation (see~\secref{sec:related} for differences between multi-domain AL and active domain adaptation); we adapt their query methods for our multi-domain AL setting. \textbf{BvSB-DA} combines domain adaptation (DA) and active learning. 
\textbf{Margin}, \textbf{BADGE}, and \textbf{Cluster-Margin} are state-of-the-art query strategies based on uncertainty and/or diversity (more details in the Supplement). 
Each baseline has two variants, \textbf{Joint}, which treats all data as a single domain and performs instance-level AL, and \textbf{Separate}, which assigns identical labeling budgets to each domain and performs instance-level AL in each domain.

\textbf{Implementation.} 
We run a model in $R=5$ rounds plus an initial round, with three different random seeds, and report the average results over three seeds (see {Sec. 1 of the Supplement} for more details). 

\begin{table*}[h]
\setlength{\tabcolsep}{2.5pt}
\caption{\small ImageCLEF results ($\%$). {``Joint'' and ``Separate'' indicate joint and separate assignment, respectively (see details in \secref{sec:baselines}).} We mark the best results with \textbf{bold face}.}
\label{tab:ImageCLEF}
\scriptsize
\begin{center}
\vskip -0.5cm
\begin{tabular}{l|cc|cc|cc|cc|cc|cc|cc}
\toprule[1.5pt]
\multirow{2}*{Query} & \multicolumn{2}{c|}{Random} & \multicolumn{2}{c|}{Margin} & \multicolumn{2}{c|}{BADGE} & \multicolumn{2}{c|}{Cluster-Margin} & \multicolumn{2}{c|}{Energy} & \multicolumn{2}{c|}{BvSB-DA} & \multirow{2}*{CAL (Ours)} \\ 
\cmidrule(r){2-13} 
 & Joint & Separate  & Joint & Separate  & Joint & Separate & Joint & Separate & Joint & Separate & Joint & Separate &   \\ \midrule
Round 0       &39.4&39.4&39.4&39.4&39.4&39.4&39.4&39.4&39.4&39.4&35.8&35.8&\textbf{41.7}\\
Round 1       &52.9&49.8&58.2&58.8&54.4&54.0&56.1&52.2&54.2&58.0&53.0&49.7&\textbf{61.2}\\
Round 2       &60.0&57.8&64.6&61.7&62.2&58.5&59.5&59.9&63.2&63.3&58.4&57.0&\textbf{69.5}\\
Round 3       &66.2&65.0&68.8&69.0&66.4&64.7&65.8&65.6&66.5&70.2&63.3&66.5&\textbf{75.3}\\
Round 4       &70.0&69.8&74.3&71.2&73.1&68.0&71.9&70.0&71.5&72.7&69.3&70.7&\textbf{74.9}\\
Round 5       &69.4&69.5&72.0&71.5&71.9&70.3&71.0&71.0&71.8&71.1&66.6&67.5&\textbf{77.1}\\ \midrule
Average       &59.7&58.6&62.9&62.0&61.2&59.2&60.6&59.7&61.1&62.5&57.7&57.9&\textbf{66.6}\\ 
\bottomrule[1.5pt]
\end{tabular}
\end{center}
\vskip -0.55cm
\end{table*}

\begin{table*}[h]
\setlength{\tabcolsep}{2.5pt}
\caption{\small Office-Caltech results ($\%$). {``Joint'' and ``Separate'' indicate joint and separate assignment, respectively (see details in \secref{sec:baselines}).} We mark the best results with \textbf{bold face}.}
\label{tab:Office-Caltech}
\scriptsize
\begin{center}
\vskip -0.5cm
\begin{tabular}{l|cc|cc|cc|cc|cc|cc|cc}
\toprule[1.5pt]
\multirow{2}*{Query} & \multicolumn{2}{c|}{Random} & \multicolumn{2}{c|}{Margin} & \multicolumn{2}{c|}{BADGE} & \multicolumn{2}{c|}{Cluster-Margin} & \multicolumn{2}{c|}{Energy} & \multicolumn{2}{c|}{BvSB-DA} & \multirow{2}*{CAL (Ours)} \\ 
\cmidrule(r){2-13} 
 & Joint & Separate  & Joint & Separate  & Joint & Separate & Joint & Separate & Joint & Separate & Joint & Separate &   \\ \midrule 
Round 0       &50.0&50.0&50.0&50.0&50.0&50.0&50.0&50.0&50.0&50.0&47.0&47.0&\textbf{53.0}\\
Round 1       &73.7&77.6&72.1&72.5&74.5&68.8&70.9&69.0&70.5&74.5&65.0&67.3&\textbf{80.7}\\
Round 2       &79.4&80.9&79.0&81.7&80.3&78.9&77.5&81.1&82.5&82.9&79.7&76.2&\textbf{89.3}\\
Round 3       &85.4&85.3&84.6&86.3&86.8&87.7&83.7&87.8&88.1&87.9&83.9&86.5&\textbf{91.6}\\
Round 4       &89.3&88.4&89.6&91.5&89.0&91.7&88.6&90.4&90.4&90.3&87.8&90.0&\textbf{93.2}\\
Round 5       &90.6&90.4&90.3&91.0&89.5&92.4&91.3&89.3&87.7&91.0&88.4&89.2&\textbf{93.0}\\ \midrule
Average       &78.1&78.8&77.6&78.8&78.3&78.3&77.0&77.9&78.2&79.4&75.3&76.1&\textbf{83.5}\\ 
\bottomrule[1.5pt]
\end{tabular}
\end{center}
\vskip -0.8cm
\end{table*}

\subsection{RotatingMNIST}

To show insight and effectiveness of our methods, we begin with a synthetic dataset, RotatingMNIST-D$6$ with $6$ domains, where original domain $i$ contains images with rotation angles in the range $[(i-1)\times {30^\circ},  i \times {30^\circ})$. 
The training and test sets of each domain contain $10000$ and $1666$ images, respectively. 
The total labeling budget per round $m$ is $150$, which is $0.25\%$ of the training set.



\textbf{Accuracy.} 
\tabref{tab:RotatingMNIST} shows the accuracy for different methods. One interesting observation is that the performance of joint assignment and separate assignment is very similar for each baseline method. 
Additionally, \tabref{tab:RotatingMNIST} reveals that directly aligning the original domains as BvSB-DA does can actually harm performance; 
in the DA setting, aligning source and target domains improves performance because source domains have plenty of labeled data and therefore provide valuable classification-relevant information for target domains; 
this is \emph{not} the case for our AL setting because most data points are unlabeled; 
{therefore, insufficient labeled data of the original domains in BvSB-DA lead to inaccurate estimation of domain classification accuracy, which introduces misleading classification-relevant information to each domain, leading to a decrease in overall classification accuracy.}
It is also worth noting that our full method CAL achieves the highest accuracy, surpassing BvSB-DA and the five instance-level methods (Random, Margin, BADGE, Cluster-Margin, and Energy) by up to $13.1\%$ and $11.2\%$, respectively, in terms of average accuracy across all rounds.

\begin{figure}
  \begin{center}
  \vskip 0.2cm
    \includegraphics[width=0.22\textwidth]{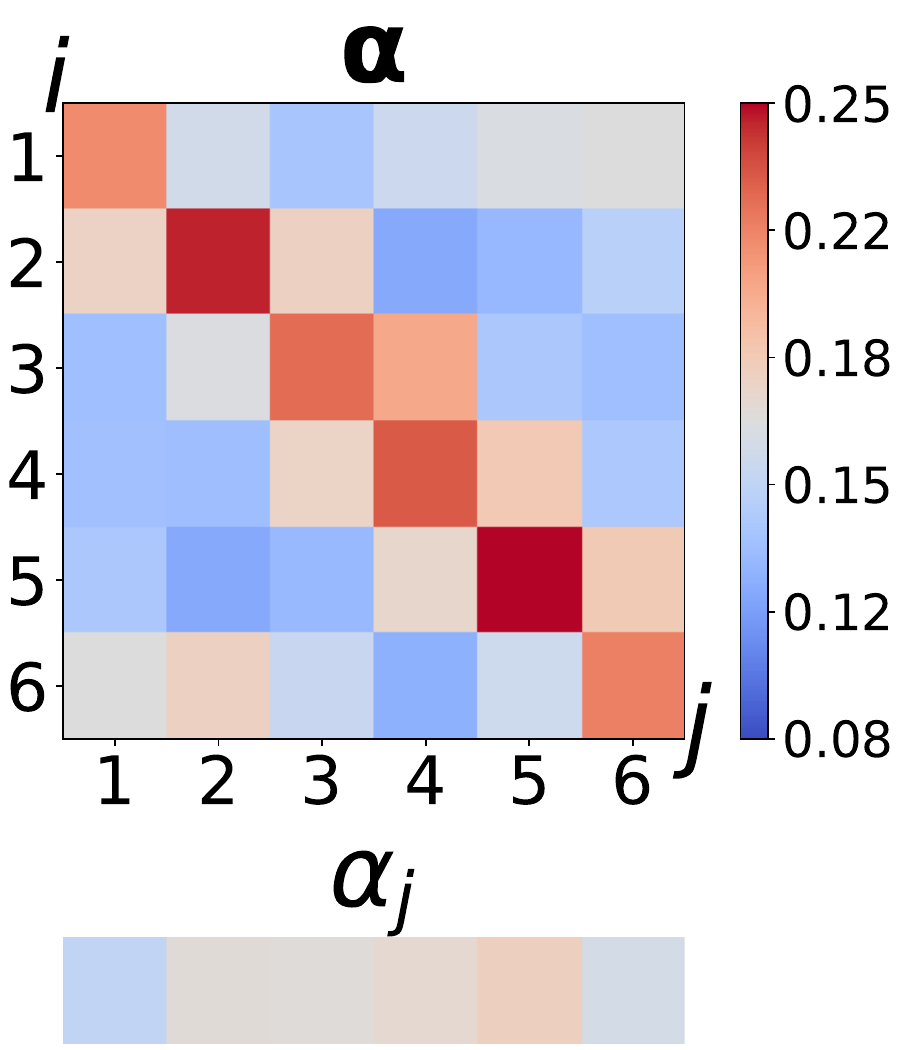}
  \end{center}
  \vskip -0.5cm
  \caption{\small Estimated similarity $\alpha_{i,j}$ and $\alpha_j$ for RotatingMNIST-D$6$.}\label{fig:heat_d6}
  \vskip -0.7cm
  \end{figure}

\textbf{Estimated Domain Similarity $\alpha_{i,j}$.} 
To gain more insights on CAL's domain selection step, we visualize the  estimated similarity weights $\alpha_{i,j}$ in \figref{fig:heat_d6}.  Specifically, \figref{fig:heat_d6}
shows CAL's estimated similarity matrix $\alp=[\alpha_{i,j}]_{i=1,j=1}^{6,6}$ on RotatingMNIST-D$6$ and the corresponding sub-budget for each domain $\beta_j=\alpha_j=\frac{1}{6}\sum_i \alpha_{i,j}$. We observe that (1) nearby domains have larger $\alpha_{i,j}$, which makes sense since they have similar rotation angles, (2) more budget is assigned to the middle domains, which is expected since labeling middle domains improves performance for domains on both sides, (3) $\alpha_{1,6}$ and $\alpha_{6,1}$ are larger, which is because digits $0$, $1$, and $8$ are identical after they are rotated by around $180^\circ$ (see {Sec. 3 in the Supplement} for more results on visualizing $\alpha_{i,j}$, {including visualization on more datasets and the evolution of the similarity matrix from round 0 to round 5}). 

\subsection{Real-World Datasets}

{We use three real-world datasets: Office-Home (65 classes) \cite{office-home}, ImageCLEF (12 classes) \cite{imageclef}, and Office-Caltech (10 classes) \cite{office-caltech}. Each dataset consists of four domains (for more details, refer to Section 1.4 of the Supplement). We split each dataset into training and test sets. In each round, we allocate a labeling budget of 200, 20, and 20, respectively for the three datasets. This budget represents approximately $1\%$ of the training set for each dataset. 
Tables \ref{tab:Office-Home}$\sim$\ref{tab:Office-Caltech} present the accuracy (\%) for different methods. Notably, our CAL demonstrates significant improvements compared to the baselines. Similar to the findings for RotatingMNIST, it is expected that BvSB-DA performs worse than the instance-level baselines. In contrast, our CAL consistently 
outperforms the instance-level baselines (Random, Margin, BADGE, Cluster-Margin, and Energy) across all five AL rounds.

\subsection{Ablation Studies}\label{sec:ablation}

\begin{table}
\setlength{\tabcolsep}{3.5pt}
\caption{\small Ablation study ($\%$) for CAL. $2^{nd}$ and $3^{rd}$ denote the second and the third terms of~\eqnref{eq:objective}, respectively. {CAL+BADGE denotes the combination of domain-level CAL and BADGE. We mark the best results with \textbf{bold face}.} } 
\vskip -0.5cm
\label{table:ablation}
\scriptsize
\begin{center}
\begin{tabular}{l|c|c|c|c|c|c}
\toprule[1.5pt]
\multirow{2}*{Method}  &CAL & CAL+       &w/o &w/o & w/o $2^{nd}$   &  \multirow{2}*{BADGE} \\ 
     &  (Ours)   & BADGE   &$3^{rd}$             & $2^{nd}$  & \& $3^{rd}$   &   \\ \midrule
Round 0       &\textbf{41.7}&41.7&43.4&36.7&39.6&39.4\\
Round 1       &\textbf{61.2}&58.3&55.5&58.5&55.3&54.0\\
Round 2       &\textbf{69.5}&67.8&63.5&63.3&59.7&58.5\\
Round 3       &\textbf{75.3}&71.2&70.7&68.5&69.8&64.7\\
Round 4       &\textbf{74.9}&71.2&72.1&71.0&71.2&68.0\\
Round 5       &\textbf{77.1}&74.9&76.7&73.3&71.2&70.3\\ \midrule
Average       &\textbf{66.6}&64.2&63.7&61.9&61.1&59.2\\
\bottomrule[1.5pt]
\end{tabular}
\end{center}
\vskip -0.8cm
\end{table}
\textbf{CAL's Components.} To investigate the impact of different components in CAL, we conduct an ablation study using the ImageCLEF dataset. \tabref{table:ablation} shows the results. The full CAL's accuracy drops by $2.4\%$ if our instance-level stragegy GraDS is replaced with BADGE, demonstrating the effectiveness of our GraDS. 
Moreover, removing the third and second terms leads to performance drops of $2.9\%$ and $4.7\%$ respectively. This highlights the effectiveness of the second term in reducing distribution shift across domains and the benefits of upper bounding $\lambda_i$ in the third term, rather than ignoring it. Furthermore, removing both the third and second terms leads to a larger performance drop of $5.1\%$. The results in \tabref{table:ablation} indicate that all CAL's components contribute to its performance improvement over the state of the art.

{\textbf{Adapting Active Domain Adaptation Methods for MUDAL.} In \secref{sec:related}, we provide a thorough comparison between multi-domain active learning (MUDAL) and domain adaptation (DA). We conduct experiments using several state-of-the-art (active) DA methods to highlight the distinctions between MUDAL and (active) DA. Consistent with the findings for BvSB-DA, Table 4 in the Supplement reveals that directly aligning the original domains can actually lead to a decline in performance. For more detailed information, please refer to Section 3 of the Supplement.} 

\textbf{Please refer to Sec. 3 in the Supplement for more visualization results, baselines, and ablation studies.}





\section{Conclusion {and Future Work}}\label{sec:conclusion}

We identify the problem of multi-domain active learning, propose the first general AL method {that integrates domain-level and instance-level information}, and provide both detailed theoretical analysis and empirical results. Our work demonstrates the effectiveness of our proposed CAL (and its simplified variants) on multi-domain data and shows its potential for significant real-world applications. 

{Our CAL method can potentially be applied to 
Natural Language Processing (NLP). There are studies that have utilized active learning in \emph{single-domain} NLP to improve classification and in-context learning \cite{nlp_bert, nlp_in_context}. With our CAL, these methodologies can potentially be adapted to multi-domain settings, even in imbalanced scenarios~\cite{MDLT}, with promising performance; this is thanks to our CAL's compatibility with instance-level active learning methods.}

\textbf{Acknowledgements:} Wang and Gao would like to acknowledge funding support (CCF-2118953, CCF-2208663, DMS-2311064, DMS-2220271, IIS-2127918).



\bibliography{aaai24}

\begin{thebibliography}{53}
\providecommand{\natexlab}[1]{#1}

\bibitem[{Anthony et~al.(1999)Anthony, Bartlett, Bartlett
  et~al.}]{neural_theory}
Anthony, M.; Bartlett, P.~L.; Bartlett, P.~L.; et~al. 1999.
\newblock \emph{Neural network learning: Theoretical foundations}, volume~9.
\newblock cambridge university press Cambridge.

\bibitem[{Arthur and Vassilvitskii(2007)}]{k-means++}
Arthur, D.; and Vassilvitskii, S. 2007.
\newblock {k-means++}: the advantages of careful seeding.
\newblock In \emph{Proceedings of the eighteenth annual ACM-SIAM symposium on
  Discrete algorithms}, 1027--1035.

\bibitem[{Ash et~al.(2020)Ash, Zhang, Krishnamurthy, Langford, and
  Agarwal}]{badge}
Ash, J.~T.; Zhang, C.; Krishnamurthy, A.; Langford, J.; and Agarwal, A. 2020.
\newblock Deep batch active learning by diverse, uncertain gradient lower
  bounds.
\newblock In \emph{International Conference on Learning Representations}.

\bibitem[{Bachman, Sordoni, and Trischler(2017)}]{al_learn_alg}
Bachman, P.; Sordoni, A.; and Trischler, A. 2017.
\newblock Learning algorithms for active learning.
\newblock In \emph{international conference on machine learning}, 301--310.
  PMLR.

\bibitem[{Balcan, Beygelzimer, and Langford(2009)}]{al_agnostic}
Balcan, M.-F.; Beygelzimer, A.; and Langford, J. 2009.
\newblock Agnostic active learning.
\newblock \emph{Journal of Computer and System Sciences}, 75(1): 78--89.

\bibitem[{Baram, Yaniv, and Luz(2004)}]{al_online}
Baram, Y.; Yaniv, R.~E.; and Luz, K. 2004.
\newblock Online choice of active learning algorithms.
\newblock \emph{Journal of Machine Learning Research}, 5(Mar): 255--291.

\bibitem[{Ben-David et~al.(2010)Ben-David, Blitzer, Crammer, Kulesza, Pereira,
  and Vaughan}]{domain_theory}
Ben-David, S.; Blitzer, J.; Crammer, K.; Kulesza, A.; Pereira, F.; and Vaughan,
  J.~W. 2010.
\newblock A theory of learning from different domains.
\newblock \emph{Machine learning}, 79(1): 151--175.

\bibitem[{Ben-David et~al.(2006)Ben-David, Blitzer, Crammer, and
  Pereira}]{domain_theory0}
Ben-David, S.; Blitzer, J.; Crammer, K.; and Pereira, F. 2006.
\newblock Analysis of representations for domain adaptation.
\newblock \emph{Advances in neural information processing systems}, 19.

\bibitem[{Berlind and Urner(2015)}]{active_change}
Berlind, C.; and Urner, R. 2015.
\newblock Active nearest neighbors in changing environments.
\newblock In \emph{International conference on machine learning}, 1870--1879.
  PMLR.

\bibitem[{Citovsky et~al.(2021)Citovsky, DeSalvo, Gentile, Karydas,
  Rajagopalan, Rostamizadeh, and Kumar}]{cluster-margin}
Citovsky, G.; DeSalvo, G.; Gentile, C.; Karydas, L.; Rajagopalan, A.;
  Rostamizadeh, A.; and Kumar, S. 2021.
\newblock Batch active learning at scale.
\newblock \emph{Advances in Neural Information Processing Systems}, 34.

\bibitem[{Dor et~al.(2020)Dor, Halfon, Gera, Shnarch, Dankin, Choshen,
  Danilevsky, Aharonov, Katz, and Slonim}]{nlp_bert}
Dor, L.~E.; Halfon, A.; Gera, A.; Shnarch, E.; Dankin, L.; Choshen, L.;
  Danilevsky, M.; Aharonov, R.; Katz, Y.; and Slonim, N. 2020.
\newblock Active learning for BERT: an empirical study.
\newblock In \emph{Proceedings of the 2020 Conference on Empirical Methods in
  Natural Language Processing (EMNLP)}, 7949--7962.

\bibitem[{Du, Zhong, and Shao(2019)}]{al_align1}
Du, X.; Zhong, D.; and Shao, H. 2019.
\newblock Building an active palmprint recognition system.
\newblock In \emph{2019 IEEE International Conference on Image Processing
  (ICIP)}, 1685--1689. IEEE.

\bibitem[{Ducoffe and Precioso(2018)}]{al_adversarial}
Ducoffe, M.; and Precioso, F. 2018.
\newblock Adversarial active learning for deep networks: a margin based
  approach.
\newblock \emph{arXiv preprint arXiv:1802.09841}.

\bibitem[{Fernando et~al.(2014)Fernando, Habrard, Sebban, and
  Tuytelaars}]{office-caltech}
Fernando, B.; Habrard, A.; Sebban, M.; and Tuytelaars, T. 2014.
\newblock Subspace alignment for domain adaptation.
\newblock \emph{arXiv preprint arXiv:1409.5241}.

\bibitem[{Fu et~al.(2021{\natexlab{a}})Fu, Cao, Wang, and Long}]{al4da_3}
Fu, B.; Cao, Z.; Wang, J.; and Long, M. 2021{\natexlab{a}}.
\newblock Transferable query selection for active domain adaptation.
\newblock In \emph{Proceedings of the IEEE/CVF Conference on Computer Vision
  and Pattern Recognition}, 7272--7281.

\bibitem[{Fu et~al.(2021{\natexlab{b}})Fu, Yuan, Wan, Xu, and Ye}]{al_align2}
Fu, M.; Yuan, T.; Wan, F.; Xu, S.; and Ye, Q. 2021{\natexlab{b}}.
\newblock Agreement-discrepancy-selection: active learning with progressive
  distribution alignment.
\newblock In \emph{Proceedings of the AAAI Conference on Artificial
  Intelligence}, volume~35, 7466--7473.

\bibitem[{Gal, Islam, and Ghahramani(2017)}]{al_bayesian}
Gal, Y.; Islam, R.; and Ghahramani, Z. 2017.
\newblock Deep bayesian active learning with image data.
\newblock In \emph{International Conference on Machine Learning}, 1183--1192.
  PMLR.

\bibitem[{Ganin et~al.(2016)Ganin, Ustinova, Ajakan, Germain, Larochelle,
  Laviolette, Marchand, and Lempitsky}]{DANN}
Ganin, Y.; Ustinova, E.; Ajakan, H.; Germain, P.; Larochelle, H.; Laviolette,
  F.; Marchand, M.; and Lempitsky, V. 2016.
\newblock Domain-adversarial training of neural networks.
\newblock \emph{JMLR}, 17(1): 2096--2030.

\bibitem[{Geifman and El-Yaniv(2017)}]{al_longtail}
Geifman, Y.; and El-Yaniv, R. 2017.
\newblock Deep active learning over the long tail.
\newblock \emph{arXiv e-prints}, arXiv--1711.

\bibitem[{Gissin and Shalev-Shwartz(2019)}]{al_align0}
Gissin, D.; and Shalev-Shwartz, S. 2019.
\newblock Discriminative active learning.
\newblock \emph{arXiv preprint arXiv:1907.06347}.

\bibitem[{Hanneke et~al.(2014)}]{al_dis_theory}
Hanneke, S.; et~al. 2014.
\newblock Theory of disagreement-based active learning.
\newblock \emph{Foundations and Trends{\textregistered} in Machine Learning},
  7(2-3): 131--309.

\bibitem[{He et~al.(2016)He, Zhang, Ren, and Sun}]{resnet}
He, K.; Zhang, X.; Ren, S.; and Sun, J. 2016.
\newblock Deep residual learning for image recognition.
\newblock In \emph{Proceedings of the IEEE conference on computer vision and
  pattern recognition}, 770--778.

\bibitem[{He et~al.(2022)He, Liu, He, and Tang}]{mudal_survey}
He, R.; Liu, S.; He, S.; and Tang, K. 2022.
\newblock Multi-domain active learning: literature review and comparative
  study.
\newblock \emph{IEEE Transactions on Emerging Topics in Computational
  Intelligence}, 7(3): 791--804.

\bibitem[{Hsu and Lin(2015)}]{al_learn}
Hsu, W.-N.; and Lin, H.-T. 2015.
\newblock Active learning by learning.
\newblock In \emph{Twenty-Ninth AAAI conference on artificial intelligence}.

\bibitem[{Huang, Jin, and Zhou(2010)}]{al_info}
Huang, S.-J.; Jin, R.; and Zhou, Z.-H. 2010.
\newblock Active learning by querying informative and representative examples.
\newblock \emph{Advances in neural information processing systems}, 23.

\bibitem[{Joshi, Porikli, and Papanikolopoulos(2009)}]{al_cls}
Joshi, A.~J.; Porikli, F.; and Papanikolopoulos, N. 2009.
\newblock Multi-class active learning for image classification.
\newblock In \emph{2009 ieee conference on computer vision and pattern
  recognition}, 2372--2379. IEEE.

\bibitem[{Kingma and Ba(2015)}]{adam}
Kingma, D.~P.; and Ba, J. 2015.
\newblock Adam: A Method for Stochastic Optimization.
\newblock In \emph{ICLR (Poster)}.

\bibitem[{Kremer, Steenstrup~Pedersen, and Igel(2014)}]{al_svm}
Kremer, J.; Steenstrup~Pedersen, K.; and Igel, C. 2014.
\newblock Active learning with support vector machines.
\newblock \emph{Wiley Interdisciplinary Reviews: Data Mining and Knowledge
  Discovery}, 4(4): 313--326.

\bibitem[{Li et~al.(2012)Li, Jin, Pan, and Sun}]{mudal_text}
Li, L.; Jin, X.; Pan, S.~J.; and Sun, J.-T. 2012.
\newblock Multi-domain active learning for text classification.
\newblock In \emph{Proceedings of the 18th ACM SIGKDD international conference
  on Knowledge discovery and data mining}, 1086--1094.

\bibitem[{Liu et~al.(2023)Liu, Xu, He, Hao, Lee, and Wang}]{TSDA}
Liu, T.; Xu, Z.; He, H.; Hao, G.; Lee, G.-H.; and Wang, H. 2023.
\newblock Taxonomy-Structured Domain Adaptation.
\newblock In \emph{ICML}.

\bibitem[{Liu et~al.(2021)Liu, Ding, Zhong, Li, Dai, and He}]{al_influence}
Liu, Z.; Ding, H.; Zhong, H.; Li, W.; Dai, J.; and He, C. 2021.
\newblock Influence selection for active learning.
\newblock In \emph{Proceedings of the IEEE/CVF International Conference on
  Computer Vision}, 9274--9283.

\bibitem[{Ma, Gao, and Xu(2021)}]{al4da_2}
Ma, X.; Gao, J.; and Xu, C. 2021.
\newblock Active universal domain adaptation.
\newblock In \emph{Proceedings of the IEEE/CVF International Conference on
  Computer Vision}, 8968--8977.

\bibitem[{Margatina et~al.(2023)Margatina, Schick, Aletras, and
  Dwivedi-Yu}]{nlp_in_context}
Margatina, K.; Schick, T.; Aletras, N.; and Dwivedi-Yu, J. 2023.
\newblock Active learning principles for in-context learning with large
  language models.
\newblock \emph{arXiv preprint arXiv:2305.14264}.

\bibitem[{{National Bureau of Statistics}(2014)}]{imageclef}
{National Bureau of Statistics}. 2014.
\newblock ImageCLEF dataset.
\newblock \url{https://www.imageclef.org/2014/adaptation/}.

\bibitem[{Prabhu et~al.(2021)Prabhu, Chandrasekaran, Saenko, and
  Hoffman}]{clue}
Prabhu, V.; Chandrasekaran, A.; Saenko, K.; and Hoffman, J. 2021.
\newblock Active domain adaptation via clustering uncertainty-weighted
  embeddings.
\newblock In \emph{Proceedings of the IEEE/CVF International Conference on
  Computer Vision}, 8505--8514.

\bibitem[{Ren et~al.(2021)Ren, Xiao, Chang, Huang, Li, Gupta, Chen, and
  Wang}]{al_survey}
Ren, P.; Xiao, Y.; Chang, X.; Huang, P.-Y.; Li, Z.; Gupta, B.~B.; Chen, X.; and
  Wang, X. 2021.
\newblock A survey of deep active learning.
\newblock \emph{ACM Computing Surveys (CSUR)}, 54(9): 1--40.

\bibitem[{Roth and Small(2006)}]{margin}
Roth, D.; and Small, K. 2006.
\newblock Margin-based active learning for structured output spaces.
\newblock In \emph{European Conference on Machine Learning}, 413--424.
  Springer.

\bibitem[{Saha et~al.(2011)Saha, Rai, Daum{\'e}, Venkatasubramanian, and
  DuVall}]{al4da_0}
Saha, A.; Rai, P.; Daum{\'e}, H.; Venkatasubramanian, S.; and DuVall, S.~L.
  2011.
\newblock Active supervised domain adaptation.
\newblock In \emph{Joint European Conference on Machine Learning and Knowledge
  Discovery in Databases}, 97--112. Springer.

\bibitem[{Schohn and Cohn(2000)}]{al_less}
Schohn, G.; and Cohn, D. 2000.
\newblock Less is more: Active learning with support vector machines.
\newblock In \emph{ICML}, volume~2, 6.

\bibitem[{Sener and Savarese(2017)}]{al_corset}
Sener, O.; and Savarese, S. 2017.
\newblock Active learning for convolutional neural networks: A core-set
  approach.
\newblock \emph{arXiv preprint arXiv:1708.00489}.

\bibitem[{Settles(2009)}]{al_survey2}
Settles, B. 2009.
\newblock Active learning literature survey.
\newblock TR1648, University of Wisconsin-Madison Department of Computer
  Sciences.

\bibitem[{Shi and Wang(2023)}]{UDIL}
Shi, H.; and Wang, H. 2023.
\newblock A Unified Approach to Domain Incremental Learning with Memory: Theory
  and Algorithm.
\newblock In \emph{NeurIPS}.

\bibitem[{Sinha, Ebrahimi, and Darrell(2019)}]{al_variational}
Sinha, S.; Ebrahimi, S.; and Darrell, T. 2019.
\newblock Variational adversarial active learning.
\newblock In \emph{Proceedings of the IEEE/CVF International Conference on
  Computer Vision}, 5972--5981.

\bibitem[{Su et~al.(2020)Su, Tsai, Sohn, Liu, Maji, and Chandraker}]{al4da_1}
Su, J.-C.; Tsai, Y.-H.; Sohn, K.; Liu, B.; Maji, S.; and Chandraker, M. 2020.
\newblock Active adversarial domain adaptation.
\newblock In \emph{Proceedings of the IEEE/CVF Winter Conference on
  Applications of Computer Vision}, 739--748.

\bibitem[{Tur, Hakkani-T{\"u}r, and Schapire(2005)}]{al_semi}
Tur, G.; Hakkani-T{\"u}r, D.; and Schapire, R.~E. 2005.
\newblock Combining active and semi-supervised learning for spoken language
  understanding.
\newblock \emph{Speech Communication}, 45(2): 171--186.

\bibitem[{Venkateswara et~al.(2017)Venkateswara, Eusebio, Chakraborty, and
  Panchanathan}]{office-home}
Venkateswara, H.; Eusebio, J.; Chakraborty, S.; and Panchanathan, S. 2017.
\newblock Deep hashing network for unsupervised domain adaptation.
\newblock In \emph{Proceedings of the IEEE Conference on Computer Vision and
  Pattern Recognition}, 5018--5027.

\bibitem[{Wang and Shang(2014)}]{al_new}
Wang, D.; and Shang, Y. 2014.
\newblock A new active labeling method for deep learning.
\newblock In \emph{2014 International joint conference on neural networks
  (IJCNN)}, 112--119. IEEE.

\bibitem[{Wang, He, and Katabi(2020)}]{cida}
Wang, H.; He, H.; and Katabi, D. 2020.
\newblock Continuously Indexed Domain Adaptation.
\newblock In \emph{International Conference on Machine Learning}, 9898--9907.
  PMLR.

\bibitem[{Xie et~al.(2021)Xie, Yuan, Li, Liu, Cheng, and Wang}]{al4da_energy}
Xie, B.; Yuan, L.; Li, S.; Liu, C.~H.; Cheng, X.; and Wang, G. 2021.
\newblock Active learning for domain adaptation: An energy-based approach.
\newblock \emph{arXiv preprint arXiv:2112.01406}.

\bibitem[{Xu et~al.(2023)Xu, Hao, He, and Wang}]{VDI}
Xu, Z.; Hao, G.; He, H.; and Wang, H. 2023.
\newblock Domain Indexing Variational Bayes: Interpretable Domain Index for
  Domain Adaptation.
\newblock In \emph{ICLR}.

\bibitem[{Xu et~al.(2022)Xu, Lee, Wang, Wang et~al.}]{GRDA}
Xu, Z.; Lee, G.-H.; Wang, Y.; Wang, H.; et~al. 2022.
\newblock Graph-Relational Domain Adaptation.
\newblock In \emph{ICLR}.

\bibitem[{Yang, Wang, and Katabi(2022)}]{MDLT}
Yang, Y.; Wang, H.; and Katabi, D. 2022.
\newblock On Multi-Domain Long-Tailed Recognition, Generalization and Beyond.
\newblock In \emph{ECCV}.

\bibitem[{Zhang et~al.(2016)Zhang, Jin, Li, Ding, and
  Yang}]{mudal_recommendation}
Zhang, Z.; Jin, X.; Li, L.; Ding, G.; and Yang, Q. 2016.
\newblock Multi-domain active learning for recommendation.
\newblock In \emph{Thirtieth AAAI Conference on Artificial Intelligence}.

\end{thebibliography}
\appendix
\onecolumn









\section{Implementation Details}\label{sec:implement}

\subsection{Upper Bound and Objective Function}
To provide context, we recall that CAL uses the following upper bound:
\begin{align}
\min_{h, e} \min_{\alp} \sum_j \alpha_{j} \epsilon_{\LM_j}(h \circ e) + \frac{1}{N}\sum_i \frac{1}{2} d_{\mathcal{H}\Delta \mathcal{H}}(\mathcal{O}_i(e (X)), \mathcal{S}_i(e(X))) + \frac{1}{N} \sum_i \lambda_{i},
\label{app_eq:cadol}
\end{align} 

\eqnref{app_eq:cadol} then leads to the following minimax game as our objective function: 
\begin{align}
\min_{h, e, \{h_i\}_{i=1}^N} \min_{\alp} \max_{f} ~~ \TM = \min_{h, e, \{h_i\}_{i=1}^N} \min_{\alp} \max_{f} ~~
V_h(h, e, \alp) 
-\lambda_d V_d(f, e, \alp) 
+ V_{\lambda}(\{h_i\}_{i=1}^N, e, \alp),
\label{app_eq:objective}
\end{align}
where each term is listed below.
\begin{align}
& V_h(h, e, \alp) = \sum\nolimits_{j=1}^N  \alpha_{j} \mathbb{E}^{\LM_j} [L_Y(h \circ e(\x), y)],\label{app_eq:v_h}
\\
& V_d(f, e, \alp) = \frac{1}{2N}\sum\nolimits_{i=1}^N  \{\mathbb{E}^{\mathcal{O}_i} [L_D(f(e(\x), i), 1)]+\sum\nolimits_{j=1}^N \alpha_{i,j} \mathbb{E}^{\LM_j} [L_D(f(e(\x), i), 0)]\},
\label{app_eq:v_d}\\
&V_{\lambda}(\{h_i\}_{i=1}^N, e, \alp) = \frac{1}{N}\sum\nolimits_{i=1}^N \sum\nolimits_{j=1}^N \alpha_{i, j} \mathbb{E}^{\LM_j} [L_Y(h_i \circ e(\x), y)],\label{app_eq:v_lambda}
\end{align}

\begin{algorithm}[h]
\caption{Learning CAL}
\begin{algorithmic}[1]\label{alg:cadol}
\STATE \textbf{Require:} 
\\ $\theta\triangleq\{e, h, \{h_i\}_{i=1}^N, \alp, f  \}$: model parameters,  $D_{\OM}$: unlabeled dataset, $D_{\LM}$: labeled dataset, 
\\ $N$: number of domains, $R+1$: number of training rounds, $m_0$: number of labeled samples in the initial round, $m$: number of additional labeled samples in other rounds, 
\\ $T$: number of epochs,  $D_{\OM}^{mini}$: minibatch of size $B$ drawn from $D_{\OM}$, $D_{\LM}^{mini}$: minibatch of size $B$ drawn from $D_{\LM}$, $Q$: a query strategy.
\STATE Initialize labeled dataset $D_{\LM}$:  $D_{\LM} \gets \{\}$.
\STATE \emph{$\%$ \textbf{Initial Round}}
\STATE Labeled dataset $D_{\LM} \gets m_0$ domain-balanced examples randomly drawn from $D_{\OM}$.
\FOR{each round $r=1:R+1$}
\STATE \emph{$\%$ \textbf{Model Training}}
\STATE Initialize $\theta$.
\FOR{each epoch $t=1:T$}
\FOR{each minibatch in $D_{\OM}$ and $ D_{\LM}$}
\STATE Update the discriminator $f \gets f + \rho \nabla_{f}\TM(D_{\OM}^{mini}, D_{\LM}^{mini} ;\theta)$ with objective function $\TM$ ( \eqnref{app_eq:objective}).
\STATE Update the similarity weights $\alp \gets \alp - \rho \nabla_{\alp}\TM(D_{\OM}^{mini}, D_{\LM}^{mini} ;\theta)$ with objective function $\TM$ ( \eqnref{app_eq:objective}).
\STATE Update the encoder and classifiers $(e, h, \{h_i\}_{i=1}^N) \gets (e, h, \{h_i\}_{i=1}^N) - \rho \nabla_{(e, h, \{h_i\}_{i=1}^N)}\TM(D_{\OM}^{mini}, D_{\LM}^{mini} ;\theta)$ with the objective function $\TM$ ( \eqnref{app_eq:objective}).
\ENDFOR
\ENDFOR
\IF{$r<R+1$}
\FOR{each labeled domain $j=1:N$}
\STATE \emph{$\%$ \textbf{Domain-Level Selection}}
\STATE Calculate sub-budget for domain $j$: $m_j^{(r)} = (\alpha_j^{(r)}-\alpha_j^{(r-1)} )\times m$.
\STATE \emph{$\%$ \textbf{Instance-Level Selection}}
\STATE Labeled dataset $D_{\LM} \gets m_j^{(r)}$ examples drawn from data of domain $j$ in $ D_{\OM} \enspace \backslash \enspace D_{\LM}  $ by query strategy $Q$.
\ENDFOR
\ENDIF
\ENDFOR
\end{algorithmic}
\end{algorithm}

\subsection{Practical Implementation for Objective Functions}\label{sec:practical}
\textbf{Algorithm for CAL.} Training procedure of CAL is summarized in~\algref{alg:cadol}.

\textbf{Upper Bounding $\sum_i \lambda_{i}$ in~\eqnref{app_eq:cadol}.}\label{sec:3rd_term} There is no exact estimation for $\sum_i \lambda_{i}$ in~\eqnref{app_eq:cadol}, since labels are not available for most data in original domains. Therefore in practice, we approximate it using its upper bound:

\begingroup\makeatletter\def\f@size{8}\check@mathfonts
\def\maketag@@@#1{\hbox{\m@th\large\normalfont#1}}%
\begin{align*}
\sum_i \lambda_{i} = \sum_i \min_{h_i} (\epsilon_{\mathcal{O}_i}(h_i)+ \epsilon_{\mathcal{S}_i}(h_i))
\leq \sum_i (\epsilon_{\mathcal{O}_i}(h_i)+ \epsilon_{\mathcal{S}_i}(h_i))
\leq N \epsilon_{\mathcal{O}_T}(h_T)+ \sum_i \epsilon_{\mathcal{S}_i}(h_i)
\leq N + \sum_i \epsilon_{\mathcal{S}_i}(h_i)
\end{align*}
\endgroup

where $\epsilon_{\mathcal{O}_i}(h_i)$ and $\epsilon_{\mathcal{S}_i}(h_i)$ is the shorthand of $\epsilon_{\mathcal{O}_i}(h_i \circ e)$ and $\epsilon_{\mathcal{S}_i}(h_i \circ e)$, respectively. $T= \argmax_i \epsilon_{\mathcal{O}_i}(h_i)$ and $N \epsilon_{\mathcal{O}_T}(h_T)$ is not greater than $N$. We then have the final upper bound $\sum_i \lambda_{i} \leq \sum_i \epsilon_{\mathcal{S}_i}(h_i) + N$, leading to~\eqnref{app_eq:v_lambda} in the objective function. Therefore, to optimize upper bound of the third term, we apply $N$ additional domain-specific classifiers. 



\textbf{Optimization of $\alp$.} 
{We use $0/1$ error to compute CAL's objective function $\TM$ and then optimize $\alp$ in~\eqnref{app_eq:objective}; this is to better match the theoretical upper bound~\eqnref{app_eq:cadol}. Our preliminary results show that using cross entropy to compute $\TM$ also works well.} In order to empirically estimate $\HM \Delta \HM$ distance between an origin domain and its surrogate domain, we use the following equation \cite{domain_theory}.
\begin{align}
\hat{d}_{\HM \Delta \HM}(\OM_i, \SM_i) = 2\left(1 - \min_{f \in \HM \Delta \HM} \left[\frac{1}{N_{\OM_i}} \sum_{\z: f(\z, i) = 0}\!\! \I[\z\in \OM_i] + \frac{1}{N_{\SM_i}} \sum_{\z: f(\z, i) = 1}\!\! \I[\z \in \SM_i] \right] \right).
\label{app_eq:h_distance}
\end{align}
where $N_{\OM_i}$and $N_{\SM_i}$ denotes the number of samples in original domain $i$ and its surrogate domain $i$, respectively, and $f$ is the optimal conditional discriminator taking as input a feature vector $\z$ and its domain index $i$ to distinguish between the two domains. Specifically, $f(\z, i)=1$ means that the discriminator predicts that the feature vector $\z$ comes from original domain $i$. 
{For convenient implementation, the equation can be also rewritten using labeled domains $\{\LM_j\}_{j=1}^N$ and similarity weights $\{\alpha_{i, j}\}_{j=1}^N)$ as below:}
\begin{align}
\hat{d}_{\HM \Delta \HM}(\OM_i, \{\LM_j\}_{j=1}^N, \{\alpha_{i, j}\}_{j=1}^N) = 2\left(1 - \min_{f \in \HM \Delta \HM} \left[\frac{1}{N_{\OM_i}} \sum_{\z: f(\z, i) = 0}\!\! \I[\z\in \OM_i] +  \sum_{j}  \frac{\alpha_{i, j}}{N_{\LM_j}} \sum_{\z: f(\z, i) = 1}\!\!  \I[\z \in \LM_j] \right] \right),
\label{app_eq:h_distance2}
\end{align}
{where $N_{\LM_j}$ denotes the number of samples in labeled domain $j$. The last term about surroagate domain of \eqnref{app_eq:h_distance} is equal to a linear combination about labeled domains in \eqnref{app_eq:h_distance2}.}




\textbf{Initialization of $\alp$.} We initialize all elements in $\alp$, i.e., $[\alpha_{i, j}]_{i=1,j=1}^{N,N}$, to $\frac{1}{N}$ before training, assuming no prior knowledge on the similarity among different domains.

\subsection{The Introduction of Baselines} \label{sec:baseline}

{We compared our CAL with six AL baselines, including \textbf{Random} \cite{al_survey}, \textbf{Margin} \cite{margin}, \textbf{BADGE} \cite{badge}, \textbf{Cluster-Margin}~\cite{cluster-margin}, \textbf{Energy} \cite{al4da_energy}, and \textbf{BvSB-DA} \cite{mudal_survey}. Among them, \textbf{Random} is a simple strategy that randomly selects data points to label. \textbf{Energy} is a state-of-the-art hybrid query strategy for active domain adaptation (see the Related Work section in the main paper for differences between multi-domain AL and active domain adaptation); we adapt their query methods for our multi-domain AL setting. \textbf{BvSB-DA} combines domain adaptation (DA) and active learning. 
\textbf{Margin}, \textbf{BADGE}, and \textbf{Cluster-Margin} are state-of-the-art query strategies based on uncertainty and/or diversity. }



\begin{enumerate}
\item \textbf{Random} \cite{al_survey}: The basic baseline of randomly selecting samples in the query stage.

\item \textbf{Margin} \cite{margin}: An uncertainty-based method that selects samples with the smallest margins, where `margin' is defined as the difference between the largest confidence and the second-largest confidence of a sample.

\item \textbf{BADGE} \cite{badge}: A query strategy that utilizes the \emph{K-Means++} to capture both uncertainty and diversity by sampling diverse gradient embeddings of the last layer of the neural network.

\item \textbf{Cluster-Margin}~\cite{cluster-margin}: A recently proposed query algorithm to improve efficiency in the large-batch setting; it diversifies batches of samples with the smallest margins by leveraging hierarchical clustering. 

\item \textbf{Energy} \cite{al4da_energy}: A recently proposed hybrid query algorithm for active domain adaptation to select samples with high uncertainty as well as samples with high free energy that are unique to the target domain. 
{\item \textbf{BvSB-DA} \cite{mudal_survey}: A method combining alignment-based domain adaptation algorithms and the Margin strategy. \cite{mudal_survey} shows that the performance of BvSB-DA is comparable to  single-domain methods. However, our experiments indicate that BvSB-DA underperforms single-domain methods. Please see the Experiments section in the main paper for more details.}
\end{enumerate}

{Each baseline has two variants with separate assignment and joint assignment. \textbf{Separate Assignment:} With a total budget of $m$ labels on $N$ domains per round, our baselines first uniformly assign a sub-budget of $m/N$ to each domain, and perform single-domain AL using one of the five query strategies (which are designed for single-domain AL) above ({$m_0=m$ by default}). \textbf{Joint  Assignment:} Treating all domains as one whole domain and using the same strategies for \textbf{S}ingle \textbf{D}omain. \emph{Therefore, unlike separate assignment, where a sub-budget of $m/N$ is separately assigned to each domain, baselines with joint assignment select $m$ samples from the `combined' single domain using a query strategy in each training round.}}


\subsection{More Details on the Datasets}\label{sec:dataset}
\begin{figure*}
	\centering
	\subfigure[RotatingMNIST]{
		\begin{minipage}[b]{0.25\linewidth}
			\centering
			\includegraphics[width=0.65in]{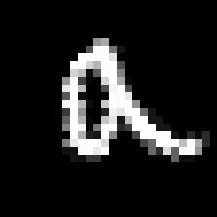}
			\includegraphics[width=0.65in]{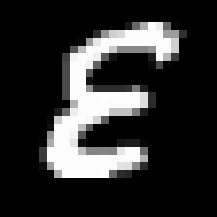}
			
		\end{minipage}%
	}%
		\subfigure[Office-Home]{
		\begin{minipage}[b]{0.25\linewidth}
			\centering
			\includegraphics[width=0.65in]{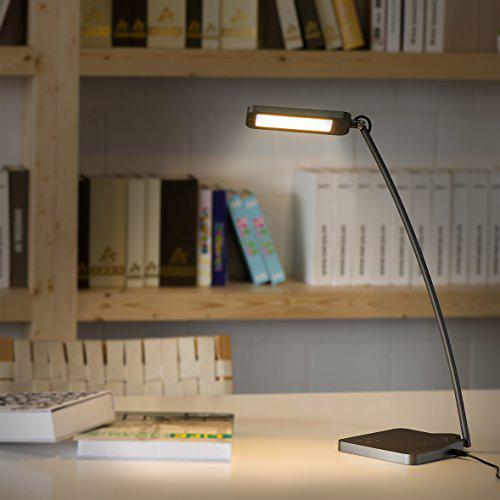}
			\includegraphics[width=0.65in]{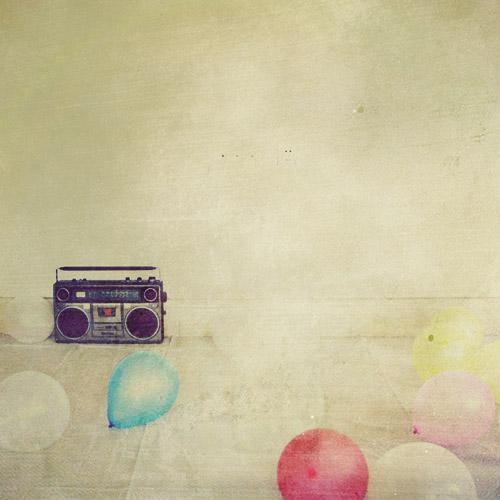}
			
		\end{minipage}%
	}%
	\subfigure[ImageCLEF]{
		\begin{minipage}[b]{0.25\linewidth}
			\centering
			\includegraphics[width=0.65in]{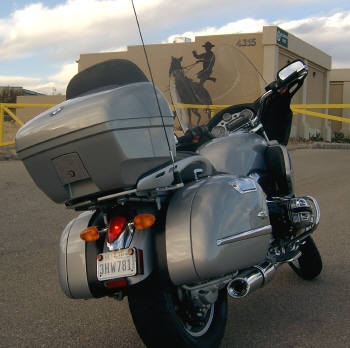}
			\includegraphics[width=0.65in]{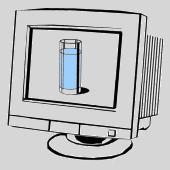}
			
		\end{minipage}%
	}%
	\subfigure[Office-Caltech]{
		\begin{minipage}[b]{0.25\linewidth}
			\centering
			\includegraphics[width=0.65in]{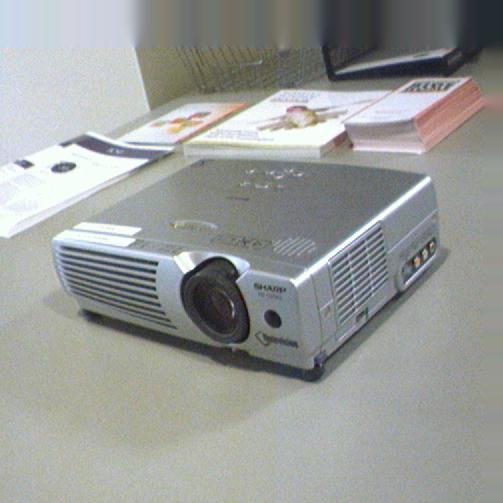}
			\includegraphics[width=0.65in]{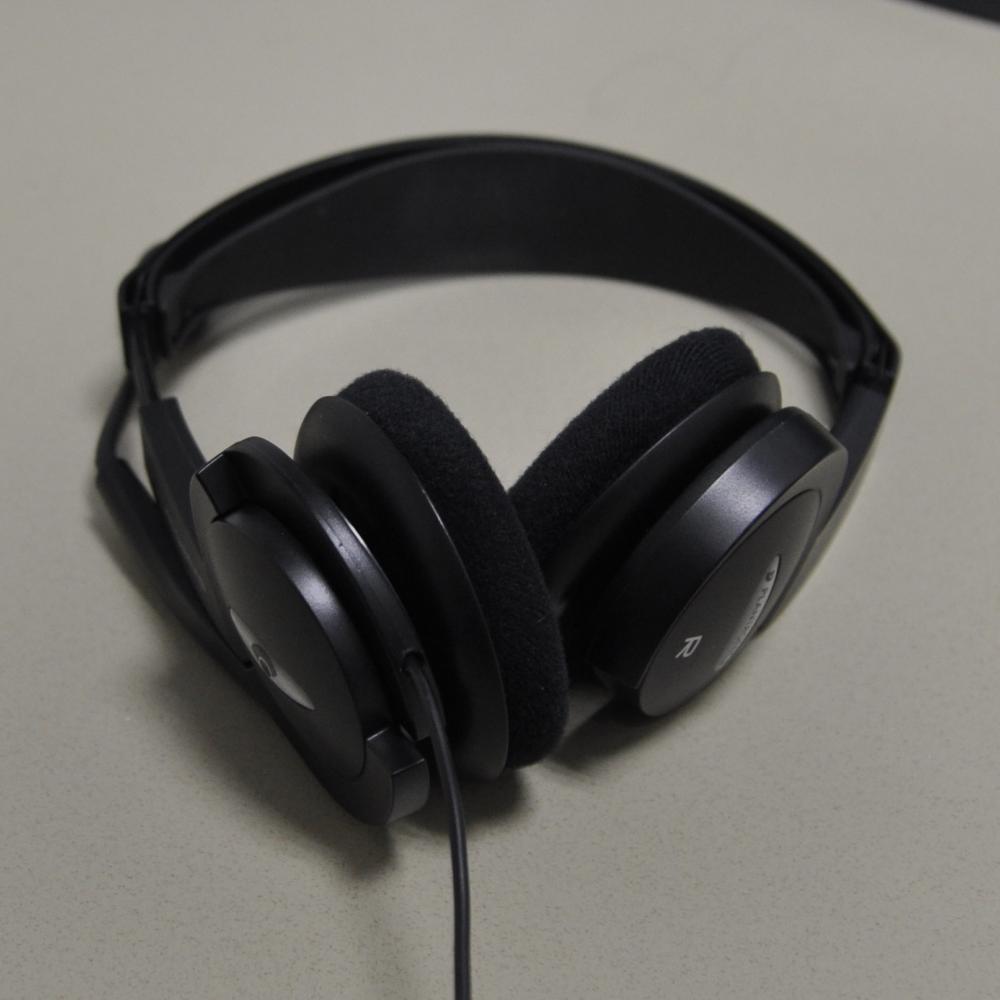}
		\end{minipage}%
	}%
	\centering
	\caption{Samples in Datasets}
	\label{fig:samples}
\end{figure*}

We use several synthetic datasets derived from RotatingMNIST \cite{cida} and three real-world datasets with multi-domain data, including Office-Home \cite{office-home}, ImageCLEF \cite{imageclef}, and Office-Caltech \cite{office-caltech}. The real-world datasets are popular in the domain adaptation and domain generalization community. Notably, Office-Home and Office-Caltech, despite their similar names, are irrelevant and come from different papers.
\begin{enumerate}
    \item \textbf{RotatingMNIST} \cite{cida} are designed by rotating images from MNIST in different rotation angle ranges. For example, to get a RotatingMNIST with $N$ domains, i.e., RotatingMNIST-D$N$, we designate images that rotate in $[(i-1)\times \frac{180^\circ}{N}),  i \times \frac{180^\circ}{N})$ as the original domain $i$. The training set and test set of each domain have $\frac{60,000}{N}$ and $\frac{10,000}{N}$ images, respectively. For example, the $3$-th original domain contains images rotating by $[60^\circ, 90^\circ)$ in RotatingMNIST-D$6$.
    \item \textbf{Office-Home \cite{office-home}} contains $15,588$ images that come from 4 domains, including Artistic images, Clip Art, Product images, and Real-World images. Each domain has $65$ object categories. 
    \item \textbf{ImageCLEF \cite{imageclef}} consists of $4$ domains with $12$ classes and total $600$ images for each domain. Images of each domain are collected from a different dataset.
    \item \textbf{Office-Caltech \cite{office-caltech}} contains $4$ domains, i.e., Amazons, Webcam, DSLR, and Caltech. It is constructed by keeping the 10 common classes shared by Office-31 and Caltech-256, with $2,533$ images in total.
\end{enumerate}
We split each real-world dataset into {training sets and test sets, where $80\%$ of the images in each domain as the training sets}. Some samples from these datasets are shown in~\figref{fig:samples}.

\subsection{Enhancing CAL with Augmented Instance-Level Acquisition}

{We propose a new instance-level strategy, dubbed \textbf{Gra}dient with \textbf{D}iscriminator \textbf{S}core (GraDS), 
which combines domain-level information with the gradient-based method, BADGE, to enhance our CAL. We choose to build GraDS on BADGE, since we found that combining CAL with BADGE is more effective than other strategies.}

{
Our decision to incorporate exclusive information from CAL (to be discussed in detail later) into the BADGE strategy is based on empirical evidence demonstrating that CAL+BADGE surpasses CAL combined with alternative strategies across the majority of datasets. BADGE consists of two primary steps: (1) calculating the gradient of the predicted label $\hat{y}$ with respect to the parameters $\theta$ in the classifier $h$'s final output layer, i.e., $\frac{\partial h_{\hat{y}}(Z)}{\partial \theta}$, and (2) employing k-means++ for gradient clustering to facilitate sample selection. A larger gradient magnitude indicates a data point with higher uncertainty, thereby offering an uncertainty estimation for a given data point.}


{Nonetheless, BADGE might not precisely estimate uncertainty, particularly for data points outside the training domain. In our scenario, this refers to a surrogate domain containing only labeled data. Such outlier data points are expected to exhibit greater uncertainty for the classifier $h$. To tackle this issue, we propose incorporating such "outlier score" into BADGE, resulting in our enhanced method, GraDS. Our conditional discriminator $f$ can evaluate if a data point $Z$ from the original domain $\OM_i$ is an outlier of the surrogate domain $\SM_i(Z)$ in the latent space. This is because $f(Z, i)=\frac{\OM_i(Z)}{\OM_i(Z)+\SM_i(Z)}=1.0-\frac{\SM_i(Z)}{\OM_i(Z)+\SM_i(Z)}$, where the closer $f$ is to $1$, the more uncertainty the outlier data point contributes to the classifier. As a result, $\frac{\OM_i(Z)}{\OM_i(Z)+\SM_i(Z)}$ can also serve as a sample uncertainty indicator.}


{With GraDS, we assign weights to the gradient using the product of outlier uncertainty and classifier uncertainty, i.e., $\frac{\OM_i(Z)}{\OM_i(Z)+\SM_i(Z)}\frac{\partial h_{\hat{y}}(Z)}{\partial \theta}$. This encourages GraDS to prioritize data points with higher outlier and classifier uncertainty, thereby minimizing both types of uncertainty in the subsequent training round. This process further narrows the domain distance between the original and surrogate domains and reduces classification errors in the original domain.}



{Additionally, we incorporate a temperature parameter $T$ into classifier $h$ during predicted label calculations, as follows: $\hat{y}=\argmax_{y} \frac{\exp{\frac{l_y(Z)}{T}}}{\sum_Y \exp{\frac{l_Y(Z)}{T}}}$, where $\hat{y}$ represents the predicted label and $l_y(Z)$ denotes the logit of class $Y=y$. The temperature parameter $T$ adjusts the smoothness of predicted confidence levels across all classes. When $T<1$, the gradient magnitude for a more uncertain point becomes relatively larger than that of a less uncertain point. Our empirical research indicates that $T<1$ yields superior results.}

\subsection{Other Implementation Details}\label{sec:other_implement}

\textbf{Training Process.} We run a model in $R=5$ rounds plus an initial round, with three different random seeds, and assign the same quantity of label budget in each round, i.e., $m_0=m$. We report the average results over three seeds. In each round, the number of epochs is the same for the same dataset.

\textbf{Model Architectures and Hyper-parameters.} We adopt the same neural network architecture in all methods for the same dataset. Specifically, CAL, CAL-$\alp$, CAL-FA, and the five baselines have the same encoder and classifier. For CAL and its simplified variants, there is an additional discriminator. 
{\tabref{tab:net_structure} summarizes the network architecture we use in CAL on RotatingMNIST, where the domain-specific classifiers $\{h_i\}_{i=1}^N$ share parameters with the common classifier $h$ except for the last fully-connected layer.} 
Generally, we set $\lambda_d = 1.0$ for CAL and its variants. We train all models with the Adam optimizer \cite{adam} with a learning rate of $10^{-4}$. We run all experiments on NVIDIA GEFORCE RTX 3090 GPUs. For the real-world datasets, we use Resnet50 \cite{resnet} as the backbone network, {and the architectures of the encoder, the classifiers, and the conditional discriminator are similar to those for RotatingMNIST.}

\begin{table}[t!]
\centering
\caption{Network Architecture for RotatingMNIST Experiments. C: channel size; K: kernel size, S: stride size; P: padding size; Flatten: {the function} to flatten spatial feature maps to feature vectors.}
\begin{tabular}{l|c}
\toprule[1.5pt]
Layer & Encoder $e$ \\
\midrule
1& Conv(C256, K3, S2, P1), BatchNorm2d, ReLU \\
2& Conv(C256, K3, S2, P1), BatchNorm2d, ReLU \\
3& Conv(C256, K3, S2, P1), BatchNorm2d, ReLU \\
4& Conv(C100, K4, S1, P0), ReLU \\
\midrule
Layer & Classifier $h$ \\
\midrule
1& Conv(C256, K1, S1, P0), BatchNorm2d, ReLU \\
2& Conv(C256, K1, S1, P0), BatchNorm2d, ReLU, Flatten \\
3& Linear(N10) \\
\midrule
Layer & Conditional Discriminator $f$ \\
\midrule
1& Conv(C256, K1, S1, P0), BatchNorm2d, LeakyReLU \\
2& Conv(C256, K1, S1, P0), BatchNorm2d, LeakyReLU \\
3& Conv(C256, K1, S1, P0), BatchNorm2d, LeakyReLU, Flatten \\
4& Linear(N1), Sigmoid \\
\bottomrule[1.5pt]
\end{tabular}
\label{tab:net_structure}
\end{table}

\section{Theoretical Analysis}\label{sec:app_theory}
\subsection{Theories of CAL}
In this section, we provide theoretical analysis for CAL. We first provide an upper bound for the error of one original domain in~\lemref{app_lem:bound_one}, based on which we develop the upper bound for the average error over all original domains in~\thmref{app_thm:bound_all}. We then prove the optimality of CAL's sub-budget assignment strategy in~\thmref{app_thm:beta}. 



In the main paper, we propose to use a surrogate domain $\SM_i$ to imitate an original domain $\OM_i$, thereby approximating the model's prediction error on $\OM_i$. 
\lemref{app_lem:bound_one} below provides theoretical justification and guarantees that the error of an original domain $\OM_i$ is upper bounded by the sum of (1) the error of its surrogate domain $\SM_i$ (consisting of a weighted average of labeled domains $\{\LM_j\}_{j=1}^N$), (2) the distance between $\OM_i$ and $\SM_i$, and (3) $\lambda_i$ which is constant for any $h$ and fixed $\alp$. 


\begin{lemma}[\textbf{Error Bound for One Domain}]\label{app_lem:bound_one}
Let $\mathcal{H}$ be a hypothesis space, and $h, h_i \in \mathcal{H}$. $\mathcal{O}_i(Z)$ is the feature distribution of original domain $i$, and its surrogate domain $\mathcal{S}_i(Z)= \sum_j \alpha_{i, j} \LM_j(Z)$ is a weighted average of $N$ labeled domains $\{\LM_j(Z)\}_{j=1}^N$. With the surrogate error $\epsilon_{\mathcal{S}_i}(h)=\sum_j \alpha_{i, j} \epsilon_{\LM_j}(h)$ and $\lambda_{i} = \min_{h_i} (\epsilon_{\mathcal{O}_i}(h_i)+ \epsilon_{\mathcal{S}_i}(h_i))$, we have:
\begingroup\makeatletter\def\f@size{9}\check@mathfonts
\def\maketag@@@#1{\hbox{\m@th\large\normalfont#1}}%
\begin{align*}
\epsilon_{\mathcal{O}_i}(h) 
\leq \epsilon_{\mathcal{S}_i}(h) 
+ \frac{1}{2}d_{\mathcal{H}\Delta \mathcal{H}}(\mathcal{O}_i(Z), \mathcal{S}_i(Z) ) 
+ \lambda_{i}
=\sum_{j} \alpha_{i, j} \epsilon_{\LM_j}(h) 
+ \frac{1}{2}d_{\mathcal{H}\Delta \mathcal{H}}(\mathcal{O}_i(Z), \sum_{j} \alpha_{i, j} \LM_j(Z))
+ \lambda_{i}.
\end{align*}
\endgroup
\end{lemma}

\begin{proof}
For a hypothesis $h: \mathcal{Z} \xrightarrow{} [0, 1]$, its probability of disagreeing with another hypothesis $h'\in \HM$, in terms of the distribution $p(Z)$, is defined as:
$\epsilon_p(h, h')=\mathbb{E}_{z\sim p(Z)}[|h(z)-h'(z)|]$. For any two distributions $p_a(Z)$ and $p_b(Z)$, by the definition of $\mathcal{H}\Delta \mathcal{H}$ distance, we have:

\begin{align}
d_{\mathcal{H}\Delta \mathcal{H}}(p_a(Z), p_b(Z)) & = 2 \sup_{h, h' \in \mathcal{H}} |\Pr_{z\sim p_a(Z)}[h(z) \neq h'(z)]-\Pr_{z\sim p_b(Z)}[h(z) \neq h'(z)]|
\\
& = 2 \sup_{h, h' \in \mathcal{H}} |\epsilon_{p_a}(h, h') - \epsilon_{p_b}(h, h')|
\\
& \geq 2|\epsilon_{p_a}(h, h') - \epsilon_{p_b}(h, h')|.
\end{align}

With the triangle inequality for the classification error \cite{domain_theory0}, i.e., $\epsilon(f_1, f_2) \leq \epsilon(f_1, f_3) + \epsilon(f_2, f_3)$ for any labeling functions $f_1$, $f_2$, and $f_3$, we have:

\begin{equation}
\begin{aligned}
\epsilon_{p_a}(h) & \leq \epsilon_{p_a}(h^*) + \epsilon_{p_a}(h, h^*)
\\ & \leq \epsilon_{p_a}(h^*) + \epsilon_{p_b}(h, h^*) + |\epsilon_{p_a}(h, h^*) - \epsilon_{p_b}(h, h^*)|
\\ & \leq \epsilon_{p_a}(h^*) + \epsilon_{p_b}(h, h^*) + \frac{1}{2} d_{\mathcal{H}\Delta \mathcal{H}}(p_a(Z), p_b(Z))
\\ & \leq \epsilon_{p_a}(h^*) + \epsilon_{p_b}(h) + \epsilon_{p_b}(h^*) + \frac{1}{2} d_{\mathcal{H}\Delta \mathcal{H}}(p_a(Z), p_b(Z))
\\ & \leq \epsilon_{p_b}(h) + \frac{1}{2} d_{\mathcal{H}\Delta \mathcal{H}}(p_a(Z), p_b(Z)) + min_{h''} (\epsilon_{p_a}(h'')+ \epsilon_{p_b}(h''))
\end{aligned}
\end{equation}
where $h^* = argmin_{h''}(\epsilon_{p_a}(h'')+ \epsilon_{p_b}(h''))$.

$\mathcal{S}_i(Z)$ is a mixture distribution, which can be treated as a single domain. We can therefore see that the upper bound holds by replacing $p_a(Z)$ and $p_b(Z)$ with $\mathcal{O}_i(Z)$ and $\mathcal{S}_i(Z)$.
\end{proof}

Before proving our main theoretical results~\eqnref{app_eq:cadol}, we introduce another lemma here. Lemma~\ref{theory:lemma_alpha_error} shows that the true $\alpha$-weighed error is bounded by the empirical $\alpha$-weighed error and a 
term that is related to the weight $\{\alpha_j\}_{j=1}^N$ of each error and the accumulated budget ratio $\{\beta_j\}_{j=1}^N$ of each domain. 

\begin{lemma}

Let $\mathcal{H}$ be a hypothesis space of VC dimension d. Assume $\sum_j \alpha_j =1$ and $\sum_j \beta_j =1$. The total number of data points in $N$ labeled domains $\{\LM_j(Z)\}_j^N$ is $\MM$ and $\LM_j(Z)$ contains $\beta_j \MM$ data points. For any $\delta \in (0 , 1)$, with probability at least $1-\delta$:
\begin{equation}
\begin{aligned}
\sum\nolimits_j \alpha_j \epsilon_{\LM_j}(h)
\leq 
\sum\nolimits_j \alpha_j \hat{\epsilon}_{\LM_j}(h) +2\sqrt{\big(\sum\nolimits_j \frac{\alpha_j^2}{\beta_j}\big)
\big(\frac{2d\log(2(\MM+1)+\log(\frac{4}{\delta}))}{\MM}}\big) 
\end{aligned}
\end{equation}
\label{theory:lemma_alpha_error}
\end{lemma}
\begin{proof}
For each $j \in \{1, 2, ... , N\}$, let $L_j$ be labeled samples of size $\beta_j \MM$ sampling from $\LM_j(Z)$ with the labeling function $f_j$. For fixed weights $\{\alpha_j\}_{j=1}^N$, let $\hat{\epsilon}_{\alpha}(h)=\sum_j \alpha_{j} \hat{\epsilon}_{\LM_j}(h)$ be the empirical $\alpha$-weighted error and let  $\epsilon_{\alpha}(h)$ be the true $\alpha$-weighted error.
For each labeled domain $j$, let $R_{j, 1}, ..., R_{j, \beta_j \MM}$ be random variables that take the values $(\alpha_j/\beta_j)|h(z)-f_j(z)|$. Thus, $R_{j, 1}, ..., R_{j, \beta_j \MM} \in [0, (\alpha_j/\beta_j)]$:

\begin{equation}
\begin{aligned}
\hat{\epsilon}_{\alpha}(h)=\sum_j \alpha_{j} \hat{\epsilon}_{\LM_j}(h) 
= \sum_j \alpha_j \frac{1}{\beta_j \MM} \sum_{z \in L_j} |h(z)-f_j(z)|
= \frac{1}{\MM}\sum_{j=1}^N \sum_{k=1}^{\beta_j \MM} R_{j, k}.
\end{aligned}
\end{equation}
$\mathbb{E}[\hat{\epsilon}_{\alpha}(h)]=\epsilon_{\alpha}(h)$ holds by linearity of expectations, and therefore by Hoeffding's inequality, for every $h \in \mathcal{H}$, we have
\begin{equation}
\begin{aligned}
Pr[|\hat{\epsilon}_{\alpha}(h)-\epsilon_{\alpha}(h)| \geq \epsilon] \leq 2\exp{\frac{-2\MM^2 \epsilon^2}{\sum_j \sum_{k=1}^{\beta_j \MM} range^2(R_{j, k})}} = 2\exp{\frac{-2\MM \epsilon^2}{\sum_j \frac{\alpha_j^2}{\beta_j}}}
\end{aligned}
\end{equation}
Finally, we can conclude the proof by applying the uniform convergence of empirical and true errors as well as the Vapnik-Chervonenkis theorem \cite{neural_theory}.
\end{proof}


Based on the two lemmas above, we provide a theorem that upper bounds the average error of original domains. In the upper bound, the first term is the average error of all surrogate domains, and the second term is the average domain gap between each origin domain and its surrogate domain. Minimizing the bound therefore leads to a lower average error of all original domains.

\begin{theorem}[\textbf{Error Bound for All Domains}]
\label{app_thm:bound_all}
Let $\mathcal{H}$ be a hypothesis space of VC dimension $d$ and let $h, h_i \in \mathcal{H}: \mathcal{Z} \xrightarrow{} [0, 1]$ be any hypothesis in $\mathcal{H}$. If labeled domains contain $\MM$ data points in total, with $\beta_j \MM$ assigned to the $i$-th labeled domain, then
for any $\delta \in (0 , 1)$, with probability at least $1-\delta$:
\begin{align}
\frac{1}{N}\sum\nolimits_i\epsilon_{\mathcal{O}_i}(h) 
& \leq \sum\nolimits_j \alpha_j \epsilon_{\LM_j}(h) + \frac{1}{2N}\sum\nolimits_i d_{\mathcal{H}\Delta \mathcal{H}}(\mathcal{O}_i(Z), \mathcal{S}_i(Z) ) + \frac{1}{N} \sum\nolimits_i \lambda_{i} 
= \UM 
\leq \UM_E\label{app_eq:bound_cadol}
\end{align}
where $\UM_E$ is further upper bound involving the empirical error $\sum_j \alpha_j \hat{\epsilon}_{\LM_j}(h)$:
\begingroup\makeatletter\def\f@size{7.8}\check@mathfonts
\def\maketag@@@#1{\hbox{\m@th\large\normalfont#1}}%
\begin{align}
\UM_E
= \sum_j \alpha_j \hat{\epsilon}_{\LM_j}(h) +2\sqrt{(\sum_j \frac{\alpha_j^2}{\beta_j})(\frac{2d\log(2(\MM+1)+\log(\frac{4}{\delta}))}{\MM})} 
+ \frac{1}{2N}\sum_i d_{\mathcal{H}\Delta \mathcal{H}}(\mathcal{O}_i(Z), \mathcal{S}_i(Z)) + \frac{1}{N} \sum_i \lambda_{i},\label{app_eq:ue}
\end{align}
\endgroup
where $\alpha_j= \frac{1}{N}\sum_i \alpha_{i, j}$, $\mathcal{S}_i(Z) = \sum_j \alpha_{i, j} \LM_j(Z)$, $ \sum_j \alpha_{i, j}=1$, and $\lambda_{i} = \min_{h_i} (\epsilon_{\mathcal{O}_i}(h_i)+ \epsilon_{\mathcal{S}_i}(h_i))$.
\end{theorem}

\begin{proof}
The first inequality in~\eqnref{app_eq:bound_cadol} follows after applying \lemref{app_lem:bound_one} on $N$ original domains and summing up the resulting $N$ inequalities. The second inequality in~\eqnref{app_eq:bound_cadol} follows from Lemma \ref{theory:lemma_alpha_error}. 
\end{proof}

With the upper bound $\UM$ in \eqnref{app_eq:bound_cadol} above, one can tighten the upper bound using $\min_{h, e} \min_{\alp} ~\mathcal{U}$, leading to CAL's objective function in~\eqnref{app_eq:objective}. 

Note that in each round $r$ of multi-domain AL, domain $j$ will be assigned an accumulated budget ratio of $\beta_j$ (the true budget for domain $j$ in round $r$ is therefore the difference between the accumulated budgets in two consecutive rounds). Therefore one can search for the optimal $\beta_j$ that minimizes the error bound in \eqnref{app_eq:ue}. \thmref{app_thm:beta} below shows that the optimal sub-budget strategy is achieved when $\beta_j=\alpha_j=\frac{1}{N}\sum_i\alpha_{i,j}$.

\begin{theorem}[\textbf{Optimal Budget Assignment}]\label{app_thm:beta}
Assuming $\alpha_j>0$ and $\beta_j>0$ for $j=1, 2, ..., N$, with $\sum_j \alpha_j=1$ and $\sum_j \beta_j=1$. The optimal upper bound for the average error of all domains, i.e., $\UM_E$ in~\eqnref{app_eq:ue}, is achieved when $\alpha_j=\beta_j$ for $j=1, 2, ..., N$. 
\end{theorem}
\begin{proof}
Note that optimization of $U_E$ in \eqnref{app_eq:ue} w.r.t. $\beta_j$ involves only the second term of \eqnref{app_eq:ue}:
\begin{align}
2\sqrt{\big(\sum\nolimits_j \frac{\alpha_j^2}{\beta_j}\big)
\big(\frac{2d\log(2(\MM+1)+\log(\frac{4}{\delta}))}{\MM}\big)}.
\end{align}

Given that both $\alpha_j$ and $\beta_j$ are positive and that $\sum_j \alpha_j=1$ and $\sum_j \beta_j=1$, by applying Jensen's inequality twice, we have
\begin{align}
\log \sum\nolimits_j \alpha_j \frac{\alpha_j}{\beta_j} \geq  \sum\nolimits_j \alpha_j \log  \frac{\alpha_j}{\beta_j} =  - \sum\nolimits_j \alpha_j \log  \frac{\beta_j}{\alpha_j} \geq  - \log \sum\nolimits_j \alpha_j   \frac{\beta_j}{\alpha_j}=0.\label{app_eq:opt_alpha_beta}
\end{align}
Since $f(x)=\log x$ is monotonically increasing, \eqnref{app_eq:opt_alpha_beta} implies that $\sum_i \alpha_j \frac{\alpha_j}{\beta_j}\geq 1$, where the equality happens if and only if $\frac{\alpha_j}{\beta_j}=1$ for any $j$, concluding the proof. 
\end{proof}
Intuitively the optimal sub-budget assignment $\beta_j=\frac{1}{N}\sum_i\alpha_{i,j}$ implies that representative domains that are more similar (important) to other domains will obtain more labeling budget. 

\subsection{Two Simplified Variants of CAL and  {a Single-Domain Baseline}}

To gain more insights, we now describe two simplified variants of CAL, namely CAL-$\alpha$, which only estimates the similarity weights $\alp$, and CAL-FA, which only performs feature alignment. 

\textbf{CAL-$\alp$.} 
Our first variant CAL-$\alpha$ corresponds to a simplified and looser upper bound compared to~\eqnref{app_eq:cadol}. Specifically, we have
\begingroup\makeatletter\def\f@size{7}\check@mathfonts
\def\maketag@@@#1{\hbox{\m@th\normalsize\normalfont#1}}%
\begin{align}
& \min_{h, e_h} \min_{\alp} ~ \frac{1}{N} \sum\nolimits_j (\sum\nolimits_i \alpha_{i, j}) \epsilon_{\mathcal{L}_j}(h\circ e_h) + \frac{1}{N}\sum\nolimits_i \frac{1}{2} d_{\mathcal{H}\Delta\mathcal{H}}(\mathcal{O}_i(e_f(X)), \sum\nolimits_j \alpha_{i, j} \mathcal{L}_j(e_f(X))) + \frac{1}{N} \sum\nolimits_i \lambda_{i},\label{app_eq:cadol_alpha}
\end{align}
\endgroup
where $e_h$ and $e_f$ are two encoders with shared parameters. 
Different from CAL whose second term is reduced by optimizing both $\alp$ and the encoder, CAL-$\alp$ minimizes the second term only by optimizing $\alp$, without optimizing the feature alignment encoder $e_f$. 
Accordingly~\eqnref{app_eq:cadol_alpha} becomes: 
\begin{align}
\min_{h, e_h, \{h_i\}_{i=1}^N} \min_{\alp} \max_{f} 
V_h(h, e_h, \alp) - 
\lambda_d V_d(f, e_f, \alp) + V_\lambda(\{h_i\}_{i=1}^N, e_h, \alp),
\label{app_eq:objective_alpha}
\end{align}
\textbf{CAL-FA.} 
Similarly, our second variant CAL-FA corresponds to another simplified and looser upper bound compared to~\eqnref{app_eq:cadol}. Specifically, we have
\begingroup\makeatletter\def\f@size{7}\check@mathfonts
\def\maketag@@@#1{\hbox{\m@th\normalsize\normalfont#1}}%
\begin{align}
 \min_{h, e_h, e_f} ~~ \frac{1}{N} \sum\nolimits_j (\sum\nolimits_i \alpha_{i, j}) \epsilon_{\mathcal{L}_j}(h\circ e_h) + 
 \frac{1}{N}\sum\nolimits_i \frac{1}{2} d_{\mathcal{H}\Delta \mathcal{H}}(\mathcal{O}_i(e_f(X)), \sum\nolimits_j \alpha_{i, j} \mathcal{L}_j(e_f(X))) + \frac{1}{N} \sum\nolimits_i \lambda_{i},\label{app_eq:cadol_fa}
\end{align}
\endgroup
where the only difference between~\eqnref{app_eq:cadol} and \eqnref{app_eq:cadol_fa} is that \eqnref{app_eq:cadol_fa} does \emph{not} optimize $\alp$. Instead \eqnref{app_eq:cadol_fa} focuses on searching for the feature {alignment} encoder $e_f$ that reduces the distribution shift across domains. Accordingly, we have the objective function for \eqnref{app_eq:cadol_fa} as follows:
\begin{align}
\min_{h, e_h, e_f} \max_{f} 
V_h(h, e_h, \alp) 
-\lambda_d V_d(f, e_f, \alp),
\label{app_eq:objective_fa}
\end{align}
where there is no third term compared to~\eqnref{app_eq:objective}, since $\frac{1}{N} \sum_i \lambda_{i}$ is a constant w.r.t. $h$. 

{A Single-Domain Baseline.} 
A natural baseline is to train the encoder $e_h$ and the classifier $h$ using empirical risk minimization (ERM), evenly assign sub-budget to different domains, i.e., $\alpha_j=\beta_j^{(r)}=\frac{1}{N}$, and apply single-domain AL for each domain. This is equivalent to using the upper bound below: 
\begingroup\makeatletter\def\f@size{6}\check@mathfonts
\def\maketag@@@#1{\hbox{\m@th\normalsize\normalfont#1}}%
\begin{align}
 \min_{h, e_h} ~~ \frac{1}{N} \sum\nolimits_j (\sum\nolimits_i \alpha_{i, j}) \epsilon_{\mathcal{L}_j}(h\circ e_h) + \frac{1}{N}\sum\nolimits_i \frac{1}{2} d_{\mathcal{H}\Delta \mathcal{H}}(\mathcal{O}_i(e_f(X)), \sum\nolimits_j \alpha_{i, j} \mathcal{L}_j(e_f(X))) + \frac{1}{N} \sum\nolimits_i \lambda_{i},\label{app_eq:vanilla}
\end{align}
\endgroup
which is equivalent to $\min_{h, e_h}~\frac{1}{N} \sum_j (\sum_i \alpha_{i, j}) \epsilon_{\mathcal{L}_j}(h\circ e_h)$. 

\textbf{Comparison of Different Upper Bounds.} 
Rewriting the upper bound in~\eqnref{app_eq:cadol} as $\min_{h,e}\min_{\alp}~\UM$, we can then compare the four upper bounds in~\eqnref{app_eq:cadol}, \eqnref{app_eq:cadol_alpha}, \eqnref{app_eq:cadol_fa}, and \eqnref{app_eq:vanilla} as:
\begingroup\makeatletter\def\f@size{8.4}\check@mathfonts
\def\maketag@@@#1{\hbox{\m@th\normalsize\normalfont#1}}%
\begin{align}
\min_{h, e} \min_{\alp} ~\mathcal{U} 
\leq \min_{h, e_h} \min_{\alp} ~\mathcal{U}   \leq \min_{h, e_h}~ \mathcal{U} , ~~~~~~~~
 \min_{h, e} \min_{\alp}~ \mathcal{U} 
\leq \min_{h, e_h, e_f}~ \mathcal{U}   \leq \min_{h, e_h}~ \mathcal{U},
\label{app_eq:bound_order}
\end{align}
\endgroup
meaning that~\eqnref{app_eq:cadol} is the best (tightest) upper bound among the four while \eqnref{app_eq:vanilla} is the worst (loosest) one. Both simplified variants using \eqnref{app_eq:cadol_alpha} and \eqnref{app_eq:cadol_fa} are better than the single-domain baseline using \eqnref{app_eq:vanilla} but worse than our full method using~\eqnref{app_eq:cadol}.



	

\begin{figure*}
	\centering
	\subfigure[D$30$]{
		\begin{minipage}[t]{0.3\linewidth}
			\centering
		\includegraphics[scale=0.25,valign=b]{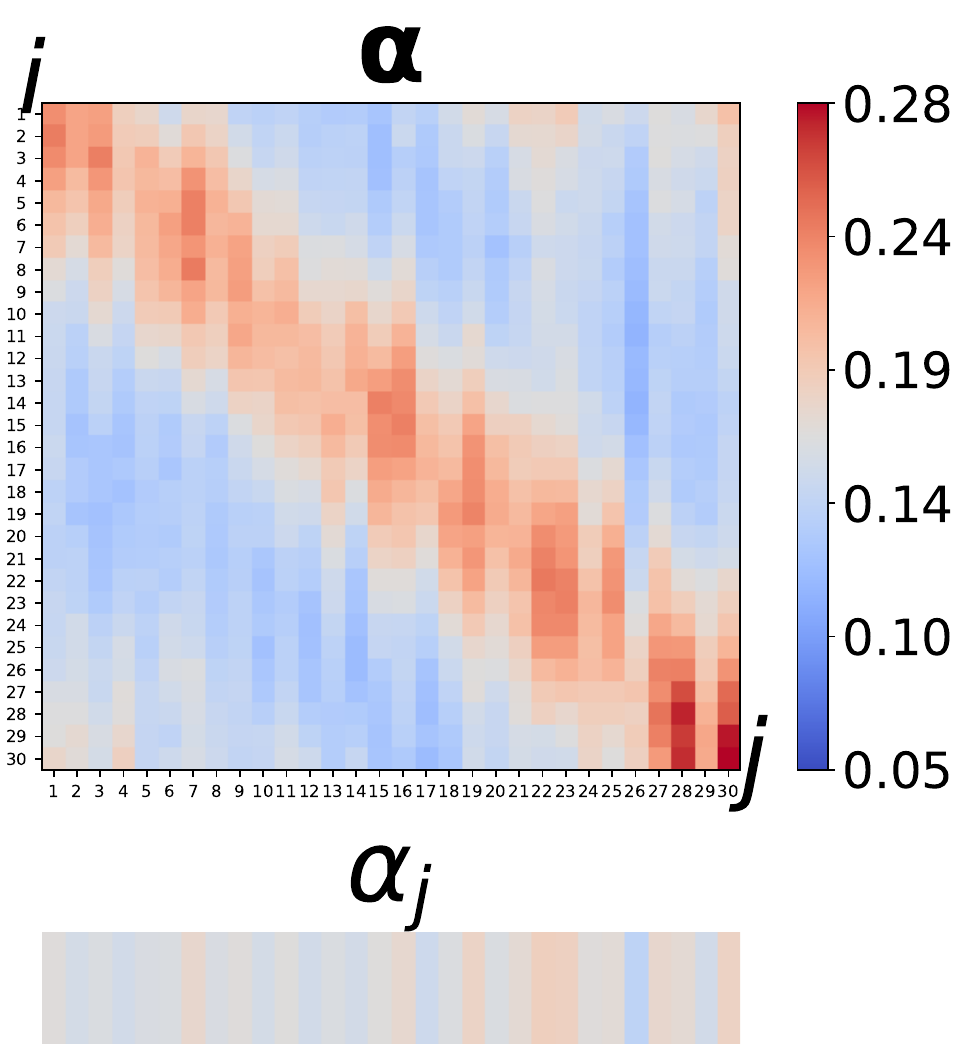}
		\end{minipage}%
	}%
	\subfigure[D$6$-$90^\circ$]{
		\begin{minipage}[t]{0.3\linewidth}
			\centering
			\includegraphics[scale=0.25,valign=b]{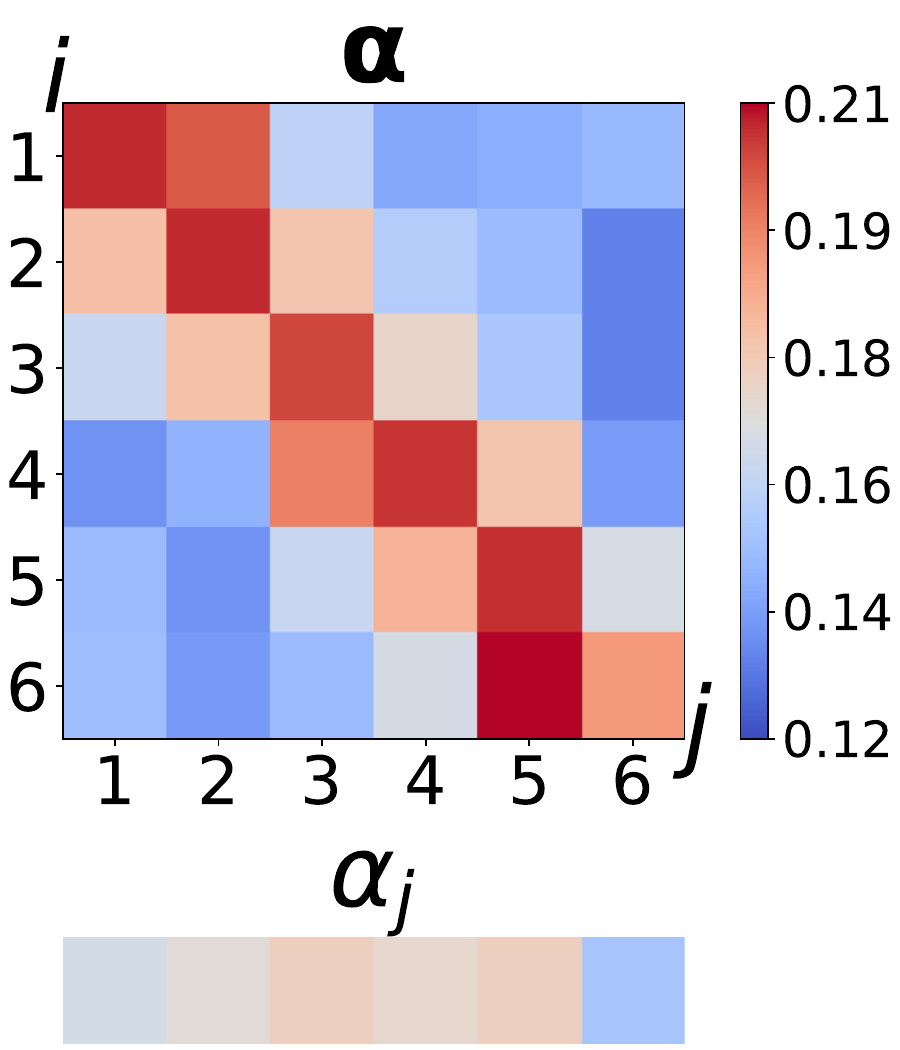}
		\end{minipage}%
	}
	
	\subfigure[D$1\&5$]{
		\begin{minipage}[t]{0.3\linewidth}
			\centering
			\includegraphics[scale=0.25,valign=b]{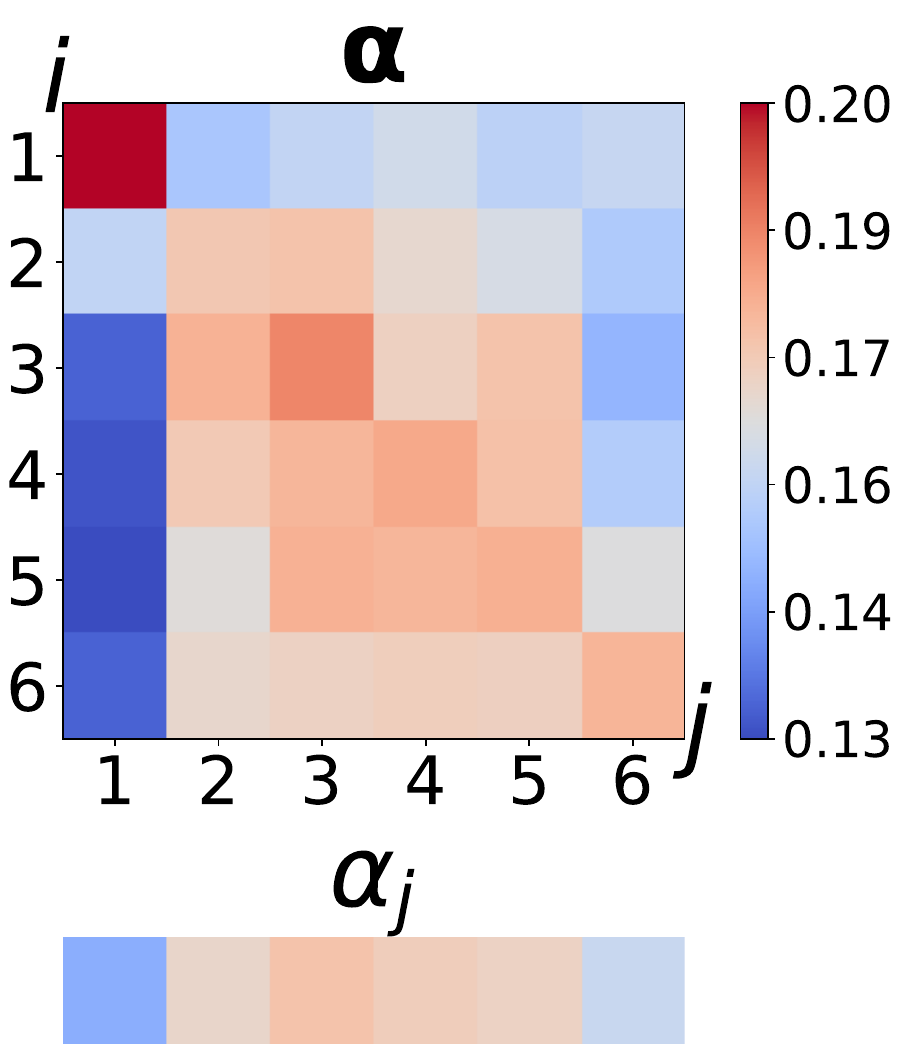}
		\end{minipage}%
	}%
		\subfigure[D$3\&3$]{
		\begin{minipage}[t]{0.3\linewidth}
			\centering
			\includegraphics[scale=0.25,valign=b]{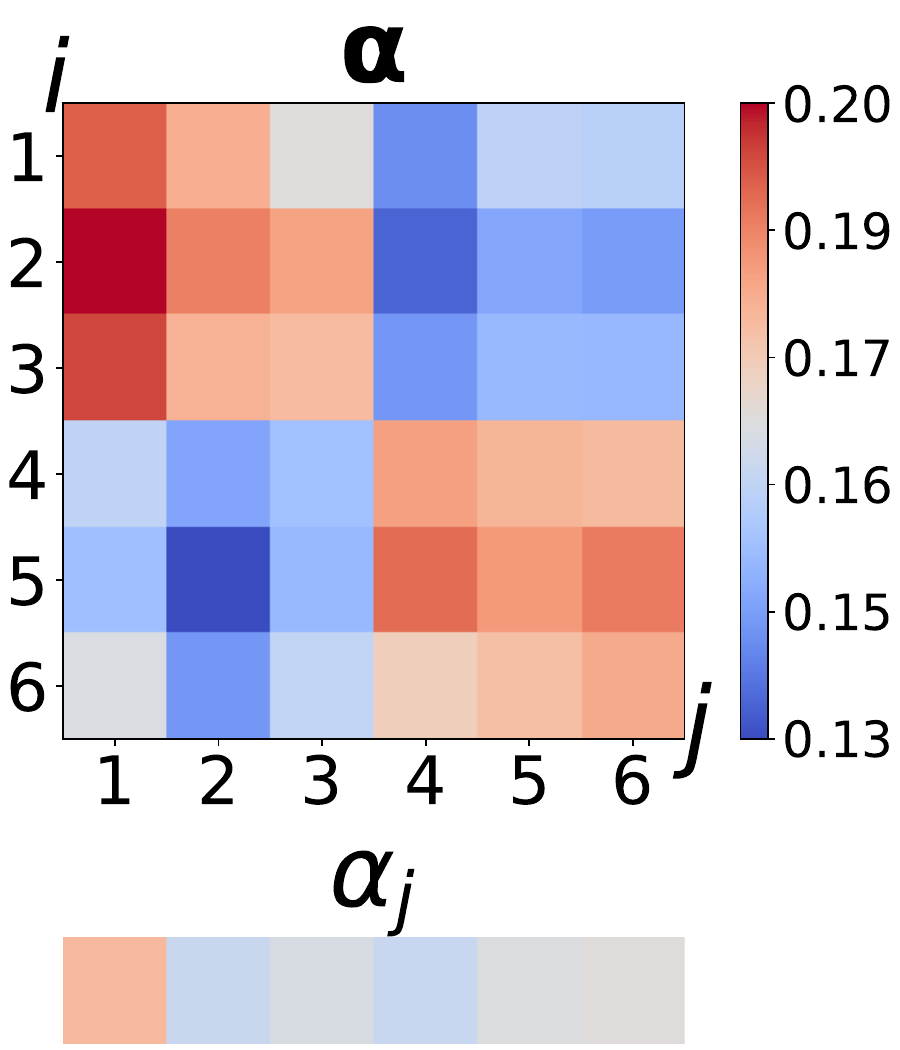}
		\end{minipage}%
	}%
	\centering
	\caption{Visualization of Estimated $\alp$ for RotatingMNIST} 

	\label{fig:heatmap}
	\vskip -0.5cm
\end{figure*}


\section{Additional Results}\label{sec:add_res}

\begin{table}[t]
\setlength{\tabcolsep}{2pt}
\caption{Ablation study ($\%$) for {CAL+Random} using RotatingMNIST-D$6$. {OA means optimal assignment.} $2^{nd}$ and $3^{rd}$ denote the second and the third terms of~\eqnref{app_eq:objective}, respectively.}
\label{app_table:ablation}
\footnotesize
\begin{center}
\begin{tabular}{l|ccccc}
\toprule[1.5pt]
Method         &CAL&w/o OA &w/o $3^{rd}$ &w/o $2^{nd}$ & {Random} \\ \midrule
Round 0       &52.1&52.1&49.7&49.6&49.2\\
Round 1       &70.2&67.6&68.6&58.8&58.4\\
Round 2       &80.3&80.0&79.2&66.4&64.5\\
Round 3       &85.0&84.9&82.2&73.4&68.8\\
Round 4       &87.6&87.4&85.8&77.1&77.7\\
Round 5       &88.7&88.5&87.6&79.6&80.5\\ \midrule
Average &77.3&76.8&75.5&67.5&66.5\\ 
\bottomrule[1.5pt]
\end{tabular}
\end{center}
\vskip -0.5cm
\end{table}
\begin{table}[h]
\vskip -0.3cm
\setlength{\tabcolsep}{2pt}
\scriptsize
\caption{Accuracy differences ($\%$) among {Random (Rand.)}, CAL-$\alpha$ (C.-$\alpha$), CAL-FA (C.-FA), and CAL with an increasing number of domains.
}
\label{table:domain_num}
\begin{center}
\vskip -0.0 cm
\begin{tabular}{c|ccc}
\toprule[1.5pt]
No. &CAL $-$ {Rand.} &CAL $-$ C.-FA&CAL $-$ C.-$\alpha$\\ \midrule
2       &9.8&4.2&8.2\\
6       &10.8&4.5&9.6\\
50       &11.6&5.2&9.9\\
150       &12.3&4.9&11.3\\
\bottomrule[1.5pt]
\end{tabular}
\end{center}
\vskip -0.6cm
\end{table}

\begin{table}[t]
\setlength{\tabcolsep}{1.0pt}
\vskip -0.3cm
\caption{Results ($\%$) for active domain adaptation methods {, Random, and CAL with Random strategy (CAL-R.)} on ImageCLEF. 
}
\vskip -0.2cm
\label{tab:ada_baseline}
\scriptsize
\begin{center}
\begin{tabular}{l|ccccccc}
\toprule[1.5pt]
Method  & DANN \cite{DANN} & CLUE \cite{clue} & EADA \cite{al4da_energy} &  {Random} \cite{al_survey} &  {CAL-R.}    \\ \midrule
Round 0 &28.1&31.1&31.1&\underline{39.4}&\textbf{41.7}\\
Round 1 &40.4&44.3&44.2&\underline{49.8}&\textbf{52.5}\\
Round 2 &44.2&50.4&54.2&\underline{57.8}&\textbf{62.6}\\
Round 3 &52.5&57.8&54.4&\underline{65.0}&\textbf{71.3}\\
Round 4 &60.3&64.5&66.3&\underline{69.8}&\textbf{71.0}\\
Round 5 &61.2&62.0&63.1&\underline{69.5}&\textbf{77.3}\\ \midrule
Average &47.8&51.7&52.2&\underline{58.6}&\textbf{62.7}\\
\bottomrule[1.5pt]
\end{tabular}
\end{center}
\end{table}

\begin{table}[t]
\setlength{\tabcolsep}{2pt}
\caption{Ablation study results ($\%$) for CAL-FA and `Alignment in Each Domain' using RotatingMNIST-D$6$ and the Random strategy.}
\label{table:align_each_domain}
\scriptsize
\begin{center}
\begin{tabular}{l|ccc}
\toprule[1.5pt]
Method         &CAL-FA&Alignment in each domain&{Random}\\ \midrule
Round 0       &49.2&49.1&49.2\\
Round 1       &64.2&57.9&58.4\\
Round 2       &74.2&64.2&64.5\\
Round 3       &79.9&68.6&68.8\\
Round 4       &83.4&77.8&77.7\\
Round 5       &86.0&83.7&80.5\\ \midrule
Average &72.8&66.9&66.5\\
\bottomrule[1.5pt]
\end{tabular}
\end{center}
\vskip -0.7cm
\end{table}



\begin{table}[t]
\vskip -0.2cm
\caption{{Results ($\%$) for methods based on deep learning with SVM-Text on RotatingMNIST-D$6$. CAL and its variants are based on Random strategy.} We mark the best results with \textbf{bold face} and the second-best results with \underline{underline}.}
\label{tab:SVM-Text}
\scriptsize
\begin{center}
\begin{tabular}{l|ccccc}
\toprule[1.5pt]
Method  & SVM-Text \cite{mudal_text} & Random  & CAL-$\alpha$  & CAL-FA  & CAL   \\ \midrule
Round 0 &45.7&49.2&\underline{49.6}&49.2&\textbf{52.1} \\
Round 1 &55.4&58.4&58.7&\underline{64.2}&\textbf{70.2} \\
Round 2 &59.9&64.5&66.3&\underline{74.2}&\textbf{80.3} \\
Round 3 &63.9&68.8&73.3&\underline{79.9}&\textbf{85.0} \\
Round 4 &65.7&77.7&77.6&\underline{83.4}&\textbf{87.6} \\ 
Round 5 &66.5&80.5&80.8&\underline{86.0}&\textbf{88.7}  \\ \midrule
Average &59.5&66.5&67.7&\underline{72.8}&\textbf{77.3}  \\
\bottomrule[1.5pt]
\end{tabular}
\end{center}
\vskip -0.7cm
\end{table}

\textbf{Visualization of Estimated Domain Similarity $\alpha_{i,j}$.}\label{sec:vis}
To gain more insights on CAL, we visualize the estimated similarity weights $\alpha_{i,j}$ of RotatingMNIST-D$6$ in Fig. 2 of the main paper. We also perform more visualization of $\alpha_{i,j}$ in the last round on RotatingMNIST-D$30$, RotatingMNIST-D$6$-$90^\circ$, RotatingMNIST-D$1\&5$, and RotatingMNIST-D$3\&3$, {with the Random strategy}. 

\begin{enumerate}

\item \textbf{RotatingMNIST-D$30$.} Similar to RotatingMNIST-D$6$, \figref{fig:heatmap}(a) shows that domains with close indices are more similar (in terms of $\alpha_{i,j}$) to each other. Notably, the $\alpha_j$ has multiple modes; this is because a single mode is unlikely to cover as many as $30$ domains and therefore the model naturally learns to use multiple modes. 


\item \textbf{RotatingMNIST-D$6$-$90^\circ$.}  Its total rotating range is $90^\circ$ and the dataset contains images rotating by $[0, 15)$, $[15, 30)$, $[30, 45)$, $[45, 60)$, $[60, 75)$, $[75, 90)$ degrees. As shown in \figref{fig:heatmap}(b), original domains and labeled domains with similar indices are more similar and thus the diagonal values in $\alp$ are higher, indicating that our CAL can effectively estimate relationships among domains. In addition, we expect that labeled domains with middle indices should obtain more labeling budget, since a labeled sample of a domain may benefit nearby domains both on its left side and on its right side. As expected, \figref{fig:heatmap}(b) shows that the average similarity of middle domains is larger; this demonstrates that our method can identify more important domains to the overall performance. These results are consistent with our insight. Different from RotatingMNIST-D$6$, $\alpha_{1,6}$ and $\alpha_{6,1}$ in RotatingMNIST-$6$-$90^\circ$ are smaller; this is because distant domains are more different in RotatingMNIST-D$6$-$90^\circ${, while digits $0$, $1$, and $8$ are identical after they are rotated by around $180^\circ$ in RotatingMNIST-D$6$}.


\item \textbf{RotatingMNIST-D$1\&5$ and RotatingMNIST-D$3\&3$.} To show more insightful cases, we visualize the estimated $\alpha_{i,j}$ on RotatingMNIST-D$1\&5$ and RotatingMNIST-D$3\&3$, which contains images rotated by $[0, 2)$, $[86, 88)$, $[88, 90)$, $[90, 92)$, $[92, 94)$, $[94, 96)$ degrees and by $[0, 2)$, $[2, 4)$, $[4, 6)$, $[90, 92)$, $[92, 94)$, $[94, 96)$ degrees, respectively. The results are in \figref{fig:heatmap}(c-d). As expected, we can observe that domains $2\sim6$ are more similar to each other in RotatingMNIST-D$1\&5$. Similarly in RotatingMNIST-D$3\&3$, domains $1\sim3$ are similar to each other and so are domains $4\sim6$. 

{\item \textbf{Evolution of the Similarity Matrix.}
\figref{fig:similiarity} shows the similarity matrix of each round on CAL in RotatingMNIST. We can see that: (1) In all rounds, nearby domains have larger $\alpha_{i, j}$, which makes sense since they have similar rotation angles. (2) As the training round increases, nearby domains' similarities are enhanced, i.e., the color around the diagonal is deeper, since CAL is more confident in the similarities.}  
\end{enumerate}

\begin{figure*}[h]
\vspace{0.4cm}
\begin{center}
    \begin{subfigure}[Round 0]{
           \centering
           \includegraphics[width=0.2\textwidth]{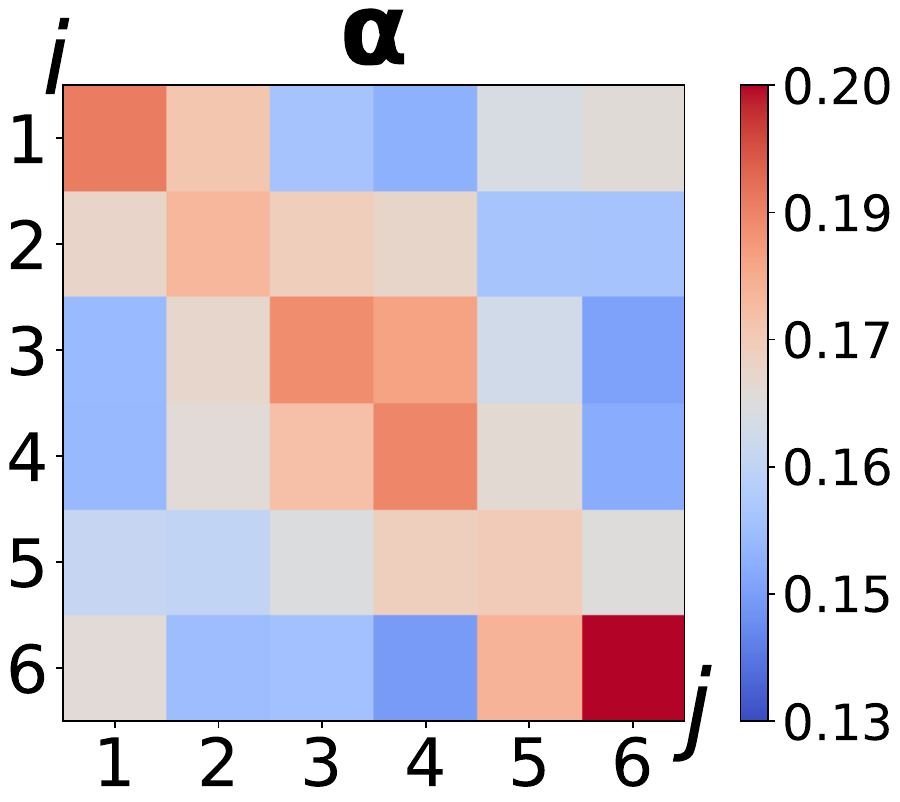}
           }
    \end{subfigure}
    \begin{subfigure}[Round 1]{
            \centering
            \includegraphics[width=0.2\textwidth]{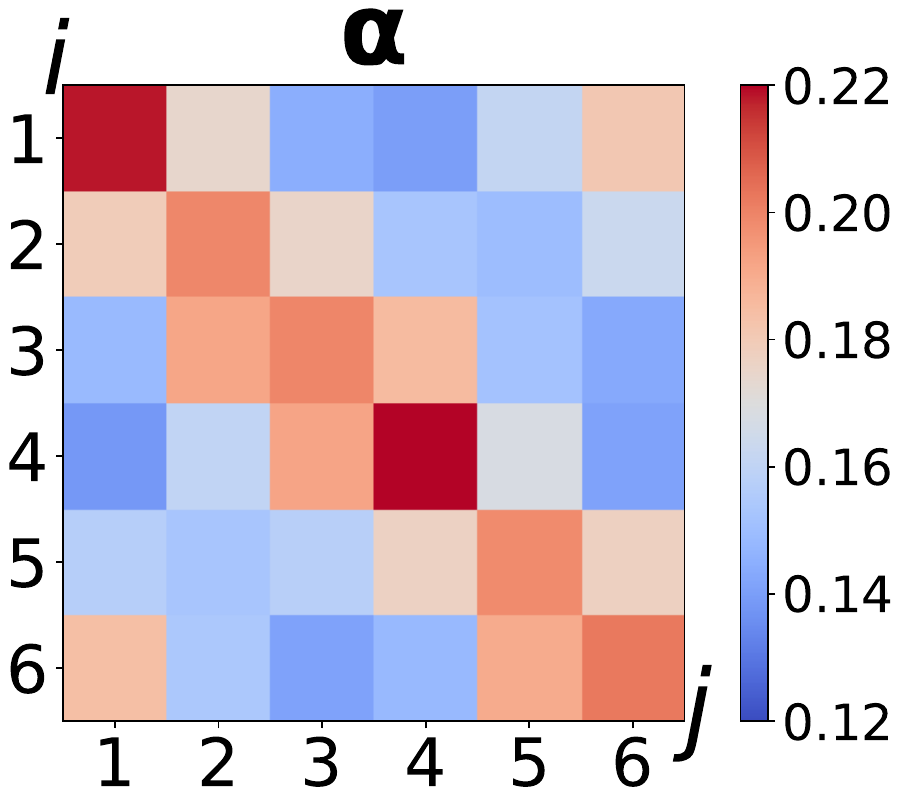}
            }
    \end{subfigure}
    \begin{subfigure}[Round 3]{
            \centering
            \includegraphics[width=0.2\textwidth]{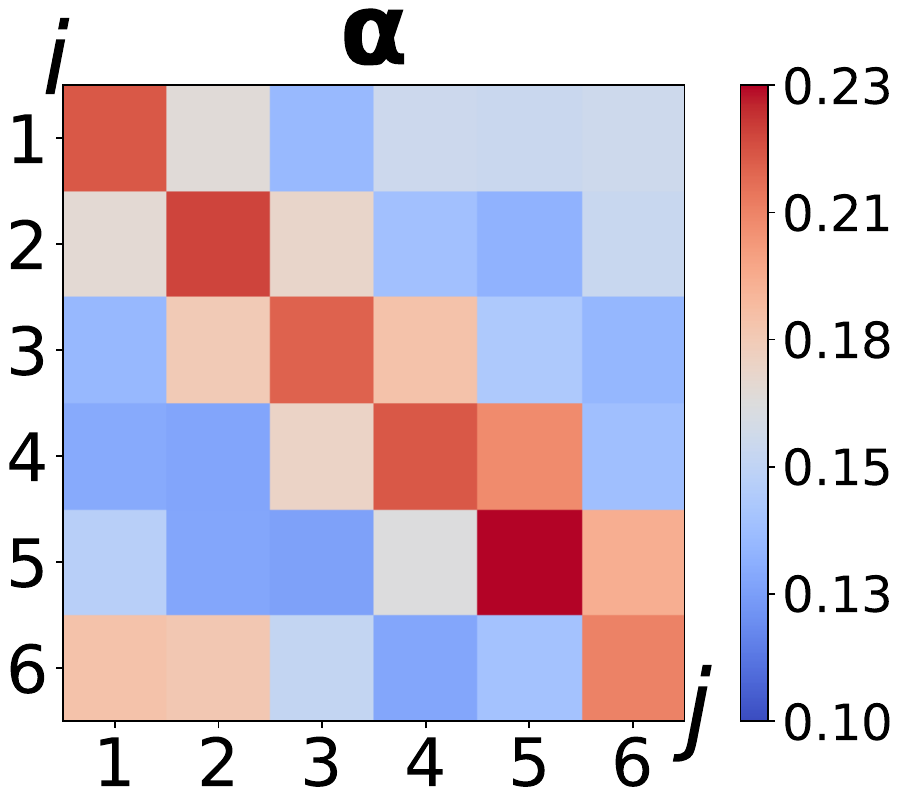}
            }
    \end{subfigure}
    \begin{subfigure}[Round 5]{
            \centering
            \includegraphics[width=0.2\textwidth]{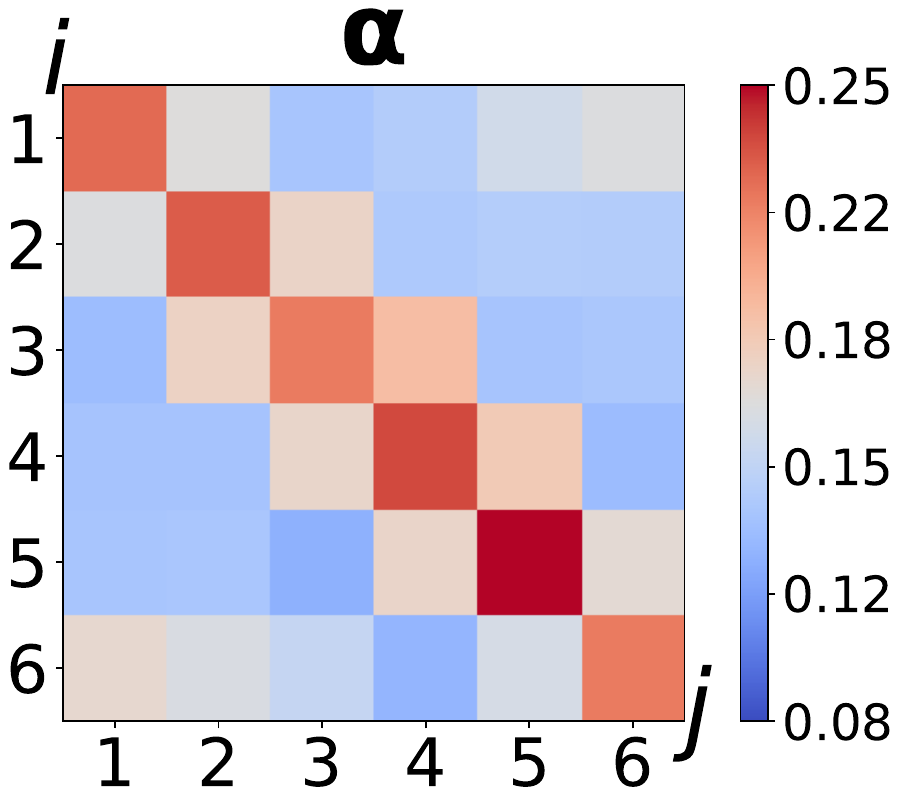}
            }
    \end{subfigure}
\end{center}
\vspace{-0.4cm}
    \caption{Visualization of Estimated $\alp$ for RotatingMNIST in Round 0, 1, 3, and 5 respectively. }
    \label{fig:similiarity}
\end{figure*}

{\textbf{Domain-Level CAL: Agnostic to Existing Query Methods.}
In Section 3 of the main paper, we explain why our Domain-Level CAL approach is agnostic to existing Query Methods. To validate this claim, we conducted additional experiments where we combined our domain-level selection methods (domain-level CAL, CAL-$\alpha$, and CAL-FA) with existing strategies. Specifically, we considered five strategies: Random, Energy, Margin, BADGE, and Cluster-Margin, and evaluated their performance on RotatingMNIST-D6, Office-Home, ImageCLEF, and Office-Caltech datasets. Table \ref{tab:rotatingmnist_mean} to \ref{tab:office-caltech_std} (in Page 13-15) present the main results, including the mean and standard deviation (represented as error bars). In these tables, ``Separate + a strategy'' (e.g., Margin) indicates using the separate assignment with the respective strategy, while ``Joint + a strategy'' implies using the joint assignment with the strategy.
The results demonstrate that the combination of our domain-level CAL methods, including its variants (CAL-$\alpha$ and CAL-FA), with each instance-level query method consistently outperforms using instance-level queries alone. This highlights the effectiveness and superiority of our domain-level CAL approach over the traditional instance-level query methods.
By incorporating domain-level selection methods with existing strategies, we achieve more robust and improved performance across the evaluated datasets. This confirms the generality of our domain-level CAL approach, as it seamlessly integrates with various existing query methods to achieve superior results.}

\textbf{Ablation Studies.} We perform an ablation study on different terms of CAL's objective function~\eqnref{app_eq:objective} and optimal assignment, using RotatingMNIST-D$6$ and the Random query strategy. \tabref{app_table:ablation} shows that CAL's performance drops by $9\%$ and $2\%$ after removing the second and the third terms, respectively; this verifies the effectiveness of the second term on reducing distribution shift across domains and the benefit of upper bounding $\lambda_i$ in the third term as opposed to ignoring it. Moreover, without our optimal assignment, i.e., using uniform label assignment instead of optimal assignment, our CAL reduces by $0.5\%$, showing the advantage of the optimal assignment with theoretical guarantees.

\textbf{The Number of Domains.}  
CAL estimates and leverages the similarity among domains to decide on the labeling sub-budget; therefore we expect CAL to bring more improvement as the number of domains increases. 
To verify this, we perform experiments on RotatingMNIST-D$2$, D$6$, 
D$50$, and D$150$ {using CAL+Random}. 
Here RotatingMNIST-D$N$ contains $N$ domains, where original domain $i$ contains images with rotation angles in the range $[(i-1)\times \frac{180^\circ}{N}),  i \times \frac{180^\circ}{N})$. The training and test sets of each domain contain $\frac{60,000}{N}$ and $\frac{10,000}{N}$ images, respectively. 
We use the \emph{Random} query strategy for {our CAL variants}.
Specifically, we first calculate average accuracy over all rounds of our CAL variants and {Random} on each dataset, and then obtain the accuracy differences between CAL and {Random (CAL$-$Rand.)}, CAL and CAL-FA (CAL$-$C.-FA), CAL and CAL-$\alpha$ (CAL$-$C.-$\alpha$), shown in \tabref{table:domain_num}. As expected, the improvement brought by our full CAL increases with the number of domains, verifying our hypothesis and CAL's effectiveness. 

\textbf{On the 'Optimal' Number of Domains}. We would like to first clarify that in our multi-domain active learning setting, the number of domains is \emph{predefined and fixed} (known a priori), which is often the case in practice. For example, each of our real-world datasets Office-Home, ImageCELF, and Office-Caltech contains 4 (predefined) domains. The RotatingMNIST is only meant as an additional 'semi-synthetic' dataset, where we divide the dataset into a certain number of domains (e.g., 2, 6, 50, 150). {\emph{The purpose of the RotatingMNIST experiments in \tabref{table:domain_num}
is to investigate whether our proposed CAL can bring more improvement as the number of domains increases (the answer is affirmative).}} Note that in the case of RotatingMNIST, there is no 'optimal' number of domains because we split the domains according to rotation angles, which are \emph{continuous} values. Investigating how to optimally split a single domain into multiple domains is certainly interesting future work, but out of the scope of this paper.

\textbf{Adapting Active Domain Adaptation Methods for MUDAL.} In Sec. 2 of the main paper, we compare multi-domain active learning (MUDAL) with domain adaptation (DA) in detail. We also run a classical DA method, DANN, with Random strategy and two state-of-the-art active DA methods, CLUE \cite{clue} and EADA \cite{al4da_energy}, to show the difference between MUDAL and (active) DA. {\emph{The key ideas of CLUE \cite{clue} and EADA \cite{al4da_energy} are to first perform domain adaptation to match the distributions of the source and target domains, and then apply query strategies to select useful unlabeled samples from the target domain.} Since the MUDAL setting does not distinguish between source and target domains and only has original domains,} we use these three methods to {directly} align all original domains {using} their proposed strategies on ImageCLEF. \tabref{tab:ada_baseline} shows that directly aligning
original domains can even hurt performance. 
In DA settings, source domains have abundant labeled data and provide fully-supervised classification-relevant information for target domains. 
Our AL settings are different because most data points are \emph{unlabeled}; {therefore, when DANN, CLUE, or EADA directly aligns original domains in multi-domain AL, insufficient labeled data of the original domains lead to inaccurate estimation of domain classification accuracy, which introduces misleading classification-relevant information to each domain, leading to a decrease in overall classification accuracy.}


\textbf{A Special Case of CAL-FA as an Additional Ablation Study.} We find that alignment {between labeled data and all data} of each original domain is a special case of our CAL-FA variant, when the similarity matrix $\alp$ is the identity matrix. Compared to this special case, our CAL-FA could utilize more labeled data to align each original domain; this is because 
CAL-FA aligns all labeled data of all domains with each original domain, while the special case only aligns labeled data of an original domain with that one original domain.
Results in \tabref{table:align_each_domain} verify our hypothesis and demonstrate the effectiveness of our CAL-FA{ using RotatingMNIST-D$6$ and the Random strategy}. 


\begin{table}[hbt]
\setlength{\tabcolsep}{2pt}
\caption{{Results ($\%$) for the heuristic method, Random strategy, and CAL on ImageCLEF.}
We mark the best results with \textbf{bold face} and the second-best results with \underline{underline}.}
\label{tab:heuristic}
\scriptsize
\begin{center}
\begin{tabular}{l|ccccc}
\toprule[1.5pt]
Method  & Heuristic & Random  &  {CAL+Random}  \\ \midrule
Round 0 &37.2 &\underline{39.4}&\textbf{41.7}\\
Round 1 &46.0 &\underline{49.8}&\textbf{52.5}\\
Round 2 &51.9 &\underline{57.8}&\textbf{62.6}\\
Round 3 &58.3 &\underline{65.0}&\textbf{71.3}\\
Round 4 &63.5 &\underline{69.8}&\textbf{71.0}\\
Round 5 &66.4 &\underline{69.5}&\textbf{77.3}\\ \midrule
Average &53.9 &\underline{58.6}&\textbf{62.7}\\
\bottomrule[1.5pt]
\end{tabular}
\end{center}
\vskip -0.5cm
\end{table}

{\textbf{A Heuristic Baseline: Allocating More Label Budget to Domains with Lower Performance.}
One heuristic approach for allocating the label budget is to prioritize domains with lower performance. In this baseline method, we use the domain performance from each training round as an estimator for the budget allocation. Additionally, we set aside $20\%$ of the labeled data as a validation set. However, the results presented in Table \ref{tab:heuristic} indicate that assigning more budget to domains with poor performance yields worse outcomes compared to single domain active learning (AL) methods. There are a couple of reasons for this. Firstly, the scarcity of labeled data means that adding a few more data points to poorly performing domains is unlikely to significantly improve their performance. Secondly, allocating budget for validating domain performance further reduces the available label budget for training. Overall, the combination of limited labeled data and the model's ease in handling these data make it difficult to substantially improve the performance of domains with poor initial performance through the allocation of additional data.}


\begin{table}[h]
\vskip -0.1cm
\setlength{\tabcolsep}{2pt}
\caption{{Results ($\%$) for the original version and the advanced version of CAL-FA (C.-FA) and CAL, using ImageCLEF and Energy strategy. We mark the best results with \textbf{bold face}.}}
\label{tab:cadol_plus}
\scriptsize
\begin{center}
\begin{tabular}{l|cc|cc}
\toprule[1.5pt]
Query & \multicolumn{4}{c}{Energy} \\ \midrule
Method         & C.-FA  & C.-FA++ &  CAL   & CAL++ \\ \midrule
Round 0 &\textbf{41.2} &39.3 &41.7 &\textbf{48.2}  \\
Round 1 &57.0 &\textbf{62.2} &55.0 &\textbf{63.5}  \\
Round 2 &\textbf{66.0} &65.1 &65.7 &\textbf{74.4}  \\
Round 3 &71.9 &\textbf{73.9} &\textbf{76.4}&74.9   \\
Round 4 &71.5 &\textbf{71.9} &72.7 &\textbf{73.6}  \\
Round 5 &74.0 &\textbf{74.4} &77.2 &\textbf{78.3}  \\ \midrule
Average &63.6 &\textbf{64.5}$_{+0.9}$ &64.8 &\textbf{68.8}$_{+4.0}$  \\
\bottomrule[1.5pt]
\end{tabular}
\end{center}
\vskip -0.6cm
\end{table}

\textbf{A Baseline Directly Related to MUDAL.}\label{sec:direct} As mentioned in Sec. 2 of the main paper, there are a few early works related to both active learning and multi-domain learning for specific applications such as text classification and recommend system.
However, they use linear methods and careful feature engineering and thus are not applicable to highly nonlinear data and deep learning architecture of our general setting. Even so, we try to adapt their methods with the best efforts and incorporate a comparison between our CAL and such linear methods. Specifically, we adapt \cite{mudal_text} to our setting. 
 It is worth noting that \cite{mudal_text} focuses on \emph{binary} text classification and we try our best to modify and apply its linear method to \emph{multi-class} classification. We call this adapted baseline SVM-Text. 
 \tabref{tab:SVM-Text} shows the results on RotatingMNIST-D$6$. We can see that SVM-Text \cite{mudal_text} with its own strategy {and heavy feature engineering} even underperforms the Random baseline (with a learned encoder) based on deep learning. 


{\textbf{An Advanced Version of CAL: CAL++.} We further develop an advanced version of CAL, i.e., CAL++, using two approaches \emph{that aim to further improve our method with better implementation. Note that the advanced version is still fully consistent with our theoretical analysis}. (1) As the \algref{alg:cadol} shows, CAL updates the conditional discriminator $f$ once before updating similarity matrix $\alp$ and then updates the encoder $e$. Compared with CAL, CAL++ updates the conditional discriminator $f$ once more before updating the encoder $e$. CAL is expected to underperform CAL++ since, after updating the similarity matrix, the discriminator does not remain optimal, therefore failing to provide the most accurate estimation of $\HM \Delta \HM$ distance for the encoder. To address this issue, we update the discriminator once more before updating the encoder. (2) CAL uses normalized continuous domain indices as input to the conditional discriminator, while the advanced version CAL++ uses one-hot domain indices. This is because one-hot domain indices can potentially make the conditional discriminator more expressive, 
as they are more different from each other compared to normalized continuous domain indices. \tabref{tab:cadol_plus} shows the effectiveness of our two implementation improvements. Using ImageCLEF and the Energy strategy, CAL-FA++ outperforms CAL-FA by $0.9\%$, and CAL++ outperforms CAL by $2.6\%$. In other words, CAODL++ improves upon Energy (whose results are in~\tabref{tab:imageclef_mean}) by $6.3\%$.}

\begin{table}[hbt]
\setlength{\tabcolsep}{2pt}
\caption{Results ($\%$) for the {Random strategy} and CAL, using ImageCLEF's domain "b". 
We mark the best results with \textbf{bold face}.}
\label{tab:single-al}
\scriptsize
\begin{center}
\begin{tabular}{l|ccccc}
\toprule[1.5pt]
Method  & Random  &  {CAL+Random}  \\ \midrule
Round 0 &16.4 &\textbf{17.8}\\
Round 1 &19.4 &\textbf{22.5}\\
Round 2 &20.6 &\textbf{23.3}\\
Round 3 &25.0 &\textbf{28.9}\\
Round 4 &26.1 &\textbf{28.9}\\
Round 5 &28.8 &\textbf{30.0}\\ \midrule
Average &22.7 &\textbf{25.2}\\
\bottomrule[1.5pt]
\end{tabular}
\end{center}
\vskip -0.5cm
\end{table}

{\textbf{{CAL for} Single-Domain Active Learning.}
Our method also provides valuable insight for the single-domain setting as well, though it is not the focus of our paper. We conducted an additional experiment on single-domain active learning. Specifically, we compare our CAL and the baseline on one domain of the ImageCLEF dataset; we use {Random and CAL+Random}. We then run the experiment for three different random seeds and report the average results over the three seeds in the \tabref{tab:single-al}. Our CAL improves upon the baseline by $2.5\%$ even in the single-domain active learning setting since CAL can reduce the domain gap between the labeled domain and the original domain by feature alignment. 
}

\textbf{{Significance of Performance Improvements.}} 
Our method achieves significant improvements over the state-of-the-art active learning methods. For example, the RotatingMNIST-D6 dataset's performance is improved by up to $10.8\%$, compared to single-domain baselines in terms of average
accuracy over all rounds. For the ImageCLEF datasets, the performance is improved by up to $4.2\%$.
Besides, according to results with different random seeds, the p-values on the performance improvements of our method CAL upon baselines range from $2.43\times 10^{-6}$ to $8.24\times 10^{-4}$, which are well below the threshold of 0.05 and indicate the significance of our performance gains.





\begin{table*}[h]
\setlength{\tabcolsep}{1.5pt}
\caption{RotatingMNIST-D$6$ results ($\%$) for ``Joint + a strategy'', ``Separate + a strategy'' (Sep.), CAL-$\alpha$ (C.-$\alpha$), CAL-FA (C.-FA), and CAL. We mark the best results with \textbf{bold face} and the second-best results with \underline{underline}.}
\label{tab:rotatingmnist_mean}
\scriptsize
\begin{center}
\begin{tabular}{l|ccccc|ccccc|ccccc|ccccc|ccccc}
\toprule[1.5pt]
Method         & Joint  & Sep.  & C.-$\alpha$  & C.-FA  & CAL & Joint  & Sep.  & C.-$\alpha$  & C.-FA  & CAL & Joint  & Sep.  & C.-$\alpha$  & C.-FA  & CAL & Joint  & Sep.  & C.-$\alpha$  & C.-FA  & CAL  & Joint  & Sep.  & C.-$\alpha$  & C.-FA  & CAL  \\ \midrule
Query & \multicolumn{5}{c|}{+Random} & \multicolumn{5}{c|}{+Margin} & \multicolumn{5}{c|}{+BADGE} & \multicolumn{5}{c|}{+Cluster-Margin} & \multicolumn{5}{c}{+Energy} \\ \midrule
Round 0       &49.3&49.2&\underline{49.6}&49.2&\textbf{52.1}&49.3&49.2&\underline{49.6}&49.2&\textbf{52.1}&49.3&49.2&\underline{49.6}&49.2&\textbf{52.1}&49.3&49.2&\underline{49.6}&49.2&\textbf{52.1}&49.3&49.2&49.6&\underline{49.9}&\textbf{52.1}\\
Round 1       &59.3&58.4&58.7&\underline{64.2}&\textbf{70.2}&59.9&59.3&\underline{60.8}&60.3&\textbf{62.0}&59.6&61.0&59.3&\underline{62.6}&\textbf{71.2}&59.7&58.7&59.9&\underline{60.9}&\textbf{65.2}&59.8&59.8&\underline{60.5}&58.9&\textbf{62.2}\\
Round 2       &65.2&64.5&66.3&\underline{74.2}&\textbf{80.3}&65.4&65.8&67.1&\underline{73.9}&\textbf{76.0}&65.9&66.2&67.1&\underline{74.1}&\textbf{79.5}&66.0&65.6&68.0&\underline{71.8}&\textbf{74.9}&65.5&64.7&68.4&\underline{73.0}&\textbf{76.1}\\
Round 3       &69.4&68.8&73.3&\underline{79.9}&\textbf{85.0}&69.7&70.6&75.6&\underline{80.0}&\textbf{82.7}&69.7&71.4&73.9&\underline{80.0}&\textbf{85.4}&70.1&70.0&\underline{77.0}&74.7&\textbf{79.2}&69.7&69.2&74.6&\underline{79.2}&\textbf{81.9}\\
Round 4       &77.7&77.7&77.6&\underline{83.4}&\textbf{87.6}&80.5&80.3&80.8&\underline{83.5}&\textbf{84.3}&79.2&80.5&79.6&\underline{81.5}&\textbf{87.1}&80.2&80.6&\underline{81.1}&79.6&\textbf{81.6}&79.2&80.1&78.6&\underline{82.1}&\textbf{84.7}\\
Round 5       &79.9&80.5&80.8&\underline{86.0}&\textbf{88.7}&81.9&83.1&82.8&\underline{85.4}&\textbf{86.8}&82.5&83.1&82.4&\underline{84.1}&\textbf{88.3}&82.2&\textbf{83.3}&82.8&81.3&\underline{83.1}&82.1&82.9&81.4&\underline{83.8}&\textbf{85.8}\\ \midrule
Average &66.8&66.5&67.7&\underline{72.8}&\textbf{77.3}&67.8&68.1&69.5&\underline{72.1}&\textbf{74.0}&67.7&68.6&68.7&\underline{71.9}&\textbf{77.3}&67.9&67.9&\underline{69.7}&69.6&\textbf{72.7}&67.6&67.6&68.9&\underline{71.1}&\textbf{73.8}\\ \midrule
\bottomrule[1.5pt]
\end{tabular}
\end{center}
\vskip -0.1cm
\end{table*}

\begin{table*}[h]
\setlength{\tabcolsep}{1.5pt}
\caption{Standard deviation of RotatingMNIST-D$6$ results ($\%$) for ``Joint + a strategy'', ``Separate + a strategy'' (Sep.), CAL-$\alpha$ (C.-$\alpha$), CAL-FA (C.-FA), and CAL. We train these models using different seeds and report the standard deviation of the accuracy. }
\label{tab:rotatingmnist_std}
\scriptsize
\begin{center}
\begin{tabular}{l|ccccc|ccccc|ccccc|ccccc|ccccc}
\toprule[1.5pt]
Method         & Joint  & Sep.  & C.-$\alpha$  & C.-FA  & CAL & Joint  & Sep.  & C.-$\alpha$  & C.-FA  & CAL & Joint  & Sep.  & C.-$\alpha$  & C.-FA  & CAL & Joint  & Sep.  & C.-$\alpha$  & C.-FA  & CAL  & Joint  & Sep.  & C.-$\alpha$  & C.-FA  & CAL  \\ \midrule
Query & \multicolumn{5}{c|}{+Random} & \multicolumn{5}{c|}{+Margin} & \multicolumn{5}{c|}{+BADGE} & \multicolumn{5}{c|}{+Cluster-Margin} & \multicolumn{5}{c}{+Energy} \\ \midrule
Round 0       &1.67&1.31&1.24&1.53&1.38&1.67&1.31&1.24&1.53&1.38&1.67&1.31&1.24&1.53&1.77&1.67&1.31&1.24&1.53&1.38&1.67&1.31&1.24&1.09&1.38\\ 
Round 1       &0.41&0.55&0.23&2.7&1.04&0.23&0.98&0.83&1.95&1.15&0.46&0.22&0.32&3.06&1.55&0.58&0.66&1.58&0.7&2.03&0.07&0.7&0.64&0.55&0.68\\ 
Round 2       &0.89&0.64&0.52&0.66&1.19&1.31&0.83&0.46&0.56&0.95&0.12&0.63&1.08&1.12&1.56&0.75&1.49&0.81&2.35&1.61&0.75&0.58&2.15&1.56&1.85\\ 
Round 3       &0.43&0.82&0.1&1.68&0.8&0.34&1.56&1.4&1.44&0.39&0.41&1.08&2.57&1.11&0.53&0.63&0.83&0.18&1.02&1.02&0.26&1.86&1.18&2.07&1.84\\ 
Round 4       &0.19&1.18&0.94&0.67&0.65&0.79&0.21&0.11&0.68&1.73&0.06&1.0&0.85&0.93&0.75&1.34&0.49&0.35&0.31&0.62&0.65&0.75&0.7&0.98&1.74\\ 
Round 5       &0.52&0.67&0.51&0.51&0.35&0.51&0.76&0.55&0.66&1.25&0.63&0.19&0.8&0.49&0.63&0.83&0.2&0.8&0.07&0.59&0.75&0.1&0.67&1.19&1.04\\  \midrule
Average &0.4&0.44&0.52&0.39&0.13&0.74&0.25&0.14&0.08&0.79&0.25&0.23&0.85&1.07&0.53&0.43&0.51&0.33&0.63&0.74&0.62&0.47&0.67&0.9&1.14\\ 
\bottomrule[1.5pt]
\end{tabular}
\end{center}
\vskip -0.1cm
\end{table*}

\begin{table*}[h]
\setlength{\tabcolsep}{1.5pt}
\caption{Office-Home results ($\%$) for ``Joint + a strategy'', ``Separate + a strategy'' (Sep.), CAL-$\alpha$ (C.-$\alpha$), CAL-FA (C.-FA), and CAL. We mark the best results with \textbf{bold face} and the second-best results with \underline{underline}.}
\label{tab:office-home_mean}
\scriptsize
\begin{center}
\begin{tabular}{l|ccccc|ccccc|ccccc|ccccc|ccccc}
\toprule[1.5pt]
Method         & Joint  & Sep.  & C.-$\alpha$  & C.-FA  & CAL & Joint  & Sep.  & C.-$\alpha$  & C.-FA  & CAL & Joint  & Sep.  & C.-$\alpha$  & C.-FA  & CAL & Joint  & Sep.  & C.-$\alpha$  & C.-FA  & CAL  & Joint  & Sep.  & C.-$\alpha$  & C.-FA  & CAL  \\ \midrule
Query & \multicolumn{5}{c|}{+Random} & \multicolumn{5}{c|}{+Margin} & \multicolumn{5}{c|}{+BADGE} & \multicolumn{5}{c|}{+Cluster-Margin} & \multicolumn{5}{c}{+Energy} \\ \midrule
Round 0       &32.8&32.8&\underline{33.5}&31.6&\textbf{34.5}&32.8&32.8&\underline{33.5}&31.6&\textbf{34.5}&32.8&32.8&\underline{33.5}&31.6&\textbf{34.5}&32.8&32.8&\underline{33.5}&31.6&\textbf{34.5}&32.8&32.8&\underline{33.5}&33.3&\textbf{35.0}\\
Round 1       &42.0&42.8&\underline{44.3}&43.5&\textbf{45.2}&41.3&41.6&42.4&\underline{43.5}&\textbf{45.0}&42.1&42.6&\underline{45.7}&44.6&\textbf{47.0}&41.4&42.8&\underline{43.0}&42.5&\textbf{45.2}&41.5&41.2&43.1&\textbf{43.5}&\underline{43.3}\\
Round 2       &48.3&48.6&\underline{50.9}&50.0&\textbf{52.7}&47.4&46.6&\underline{49.1}&48.6&\textbf{50.4}&48.3&48.1&50.5&\underline{51.3}&\textbf{52.2}&46.4&47.0&48.9&\underline{49.4}&\textbf{51.9}&47.7&48.0&49.5&\underline{49.7}&\textbf{52.1}\\
Round 3       &52.4&52.6&\underline{54.3}&54.2&\textbf{56.1}&50.5&51.3&\underline{54.1}&53.5&\textbf{55.0}&51.9&52.0&54.3&\underline{54.6}&\textbf{57.2}&50.0&50.6&\underline{53.4}&51.6&\textbf{54.2}&51.4&51.2&53.0&\underline{55.0}&\textbf{56.0}\\
Round 4       &56.9&\underline{57.6}&57.3&57.2&\textbf{59.2}&55.2&55.9&57.0&\underline{57.0}&\textbf{57.6}&57.3&57.1&\underline{58.7}&57.0&\textbf{59.6}&54.9&55.9&\textbf{57.6}&56.0&\underline{57.5}&56.2&56.0&56.7&\underline{58.3}&\textbf{58.9}\\
Round 5       &57.7&58.0&\underline{59.2}&58.8&\textbf{60.5}&55.4&57.7&\underline{58.6}&57.2&\textbf{59.2}&57.7&56.9&\underline{59.4}&58.8&\textbf{60.3}&56.0&57.5&\textbf{59.1}&56.6&\underline{58.9}&58.1&58.4&58.7&\underline{59.4}&\textbf{60.5}\\ \midrule
Average &48.4&48.7&\underline{49.9}&49.2&\textbf{51.4}&47.1&47.6&\underline{49.1}&48.6&\textbf{50.3}&48.3&48.2&\underline{50.3}&49.6&\textbf{51.8}&46.9&47.8&\underline{49.3}&47.9&\textbf{50.4}&48.0&47.9&49.1&\underline{49.9}&\textbf{51.0}\\ \midrule
\bottomrule[1.5pt]
\end{tabular}
\end{center}
\vskip -0.1cm
\end{table*}

\begin{table}[h]
\setlength{\tabcolsep}{1.5pt}
\caption{Standard deviation of Office-Home results ($\%$) for ``Joint + a strategy'', ``Separate + a strategy'' (Sep.), CAL-$\alpha$ (C.-$\alpha$), CAL-FA (C.-FA), and CAL. We train these models using different seeds and report the standard deviation of the accuracy. }
\label{tab:office-home_std}
\scriptsize
\begin{center}
\begin{tabular}{l|ccccc|ccccc|ccccc|ccccc|ccccc}
\toprule[1.5pt]
Method         & Joint  & Sep.  & C.-$\alpha$  & C.-FA  & CAL & Joint  & Sep.  & C.-$\alpha$  & C.-FA  & CAL & Joint  & Sep.  & C.-$\alpha$  & C.-FA  & CAL & Joint  & Sep.  & C.-$\alpha$  & C.-FA  & CAL  & Joint  & Sep.  & C.-$\alpha$  & C.-FA  & CAL  \\ \midrule
Query & \multicolumn{5}{c|}{+Random} & \multicolumn{5}{c|}{+Margin} & \multicolumn{5}{c|}{+BADGE} & \multicolumn{5}{c|}{+Cluster-Margin} & \multicolumn{5}{c}{+Energy} \\ \midrule
Round 0       &1.04&1.04&1.25&2.25&1.28&1.04&1.04&1.25&2.25&1.28&1.04&1.04&1.25&2.25&1.28&1.04&1.04&1.25&2.25&1.28&1.04&1.04&1.25&1.31&1.88\\ 
Round 1       &0.23&0.82&1.56&1.66&1.01&1.29&1.18&0.65&0.54&0.83&0.82&0.24&0.71&0.4&0.61&0.57&0.61&0.37&0.48&1.46&1.73&1.37&1.26&2.02&1.16\\ 
Round 2       &0.89&0.52&0.58&1.77&1.85&0.86&1.63&0.96&0.44&0.7&0.28&0.83&0.58&0.94&1.34&0.98&0.29&0.73&1.19&0.91&0.61&0.94&1.56&0.82&0.62\\ 
Round 3       &0.61&0.42&0.67&1.21&0.76&0.37&0.57&0.42&0.76&1.04&0.13&0.56&0.78&1.01&0.64&0.71&0.76&1.55&0.06&0.85&1.23&0.6&1.36&0.5&0.91\\ 
Round 4       &0.39&0.29&0.54&0.52&1.28&0.67&0.43&0.94&1.32&1.3&0.23&0.21&0.77&1.45&0.39&0.59&1.14&0.32&0.69&2.06&1.1&1.23&0.49&0.91&1.57\\ 
Round 5       &0.12&0.66&0.2&0.84&0.06&0.83&2.06&1.15&0.53&0.75&0.43&0.85&0.46&0.37&0.1&0.78&1.78&0.33&0.65&0.4&2.01&1.35&1.12&0.58&0.57\\  \midrule
Average &0.47&0.27&0.6&0.36&0.96&0.59&0.95&0.54&0.16&0.62&0.3&0.3&0.58&1.01&0.18&0.57&0.78&0.5&0.58&0.97&0.65&0.73&0.91&0.44&0.79\\ 
\bottomrule[1.5pt]
\end{tabular}
\end{center}
\vskip -0.1cm
\end{table}

\begin{table*}[h]
\setlength{\tabcolsep}{1.5pt}
\caption{ImageCLEF results ($\%$) for ``Joint + a strategy'', ``Separate + a strategy'' (Sep.), CAL-$\alpha$ (C.-$\alpha$), CAL-FA (C.-FA), and CAL. We mark the best results with \textbf{bold face} and the second-best results with \underline{underline}.}
\label{tab:imageclef_mean}
\scriptsize
\begin{center}
\begin{tabular}{l|ccccc|ccccc|ccccc|ccccc|ccccc}
\toprule[1.5pt]
Method         & Joint  & Sep.  & C.-$\alpha$  & C.-FA  & CAL & Joint  & Sep.  & C.-$\alpha$  & C.-FA  & CAL & Joint  & Sep.  & C.-$\alpha$  & C.-FA  & CAL & Joint  & Sep.  & C.-$\alpha$  & C.-FA  & CAL  & Joint  & Sep.  & C.-$\alpha$  & C.-FA  & CAL  \\ \midrule
Query & \multicolumn{5}{c|}{+Random} & \multicolumn{5}{c|}{+Margin} & \multicolumn{5}{c|}{+BADGE} & \multicolumn{5}{c|}{+Cluster-Margin} & \multicolumn{5}{c}{+Energy} \\ \midrule
Round 0       &39.4&39.4&36.8&\underline{41.2}&\textbf{41.7}&39.4&39.4&37.1&\underline{41.2}&\textbf{41.7}&39.4&39.4&36.8&\underline{41.2}&\textbf{41.7}&39.4&39.4&37.1&\underline{41.2}&\textbf{41.7}&39.4&39.4&37.2&\underline{41.2}&\textbf{41.7}\\
Round 1       &\textbf{52.9}&49.8&49.9&51.5&\underline{52.5}&58.2&\underline{58.8}&58.7&57.2&\textbf{60.0}&54.4&54.0&\underline{56.3}&54.4&\textbf{58.3}&56.1&52.2&\textbf{57.9}&53.4&\underline{56.5}&54.2&\underline{58.0}&\textbf{60.3}&57.0&55.0\\
Round 2       &60.0&57.8&58.5&\underline{60.3}&\textbf{62.6}&\underline{64.6}&61.7&63.5&63.4&\textbf{68.3}&\underline{62.2}&58.5&59.0&61.6&\textbf{67.8}&59.5&59.9&61.0&\underline{66.3}&\textbf{67.8}&63.2&63.3&64.3&\textbf{66.0}&\underline{65.7}\\
Round 3       &66.2&65.0&65.9&\underline{68.7}&\textbf{71.3}&68.8&69.0&70.6&\underline{73.2}&\textbf{73.8}&66.4&64.7&65.9&\textbf{72.2}&\underline{71.2}&65.8&65.6&68.0&\underline{70.4}&\textbf{71.9}&66.5&70.2&69.0&\underline{71.9}&\textbf{76.4}\\
Round 4       &70.0&69.8&69.8&\textbf{71.3}&\underline{71.0}&\textbf{74.3}&71.2&71.9&\underline{73.3}&72.4&\textbf{73.1}&68.0&69.6&\underline{72.2}&71.2&71.9&70.0&\underline{72.6}&\textbf{73.3}&72.5&71.5&\underline{72.7}&71.9&71.5&\textbf{72.7}\\
Round 5       &69.4&69.5&71.5&\underline{74.2}&\textbf{77.3}&72.0&71.5&71.0&\underline{75.8}&\textbf{77.8}&71.9&70.3&71.0&\textbf{75.8}&\underline{74.9}&71.0&71.0&72.5&\underline{73.8}&\textbf{78.8}&71.8&71.1&73.0&\underline{74.0}&\textbf{77.2}\\ \midrule
Average &59.7&58.6&58.8&\underline{61.2}&\textbf{62.7}&62.9&62.0&62.1&\underline{64.0}&\textbf{65.7}&61.2&59.2&59.8&\underline{62.9}&\textbf{64.2}&60.6&59.7&61.5&\underline{63.1}&\textbf{64.8}&61.1&62.5&62.6&\underline{63.6}& \textbf{64.8}\\ \midrule
\bottomrule[1.5pt]
\end{tabular}
\end{center}
\end{table*}

\begin{table*}[t]
\setlength{\tabcolsep}{1.5pt}
\caption{Standard deviation of ImageCLEF results ($\%$) for ``Joint + a strategy'', ``Separate + a strategy'' (Sep.), CAL-$\alpha$ (C.-$\alpha$), CAL-FA (C.-FA), and CAL. We train these models using different seeds and report the standard deviation of the accuracy. }
\label{tab:imageclef_std}
\scriptsize
\begin{center}
\begin{tabular}{l|ccccc|ccccc|ccccc|ccccc|ccccc}
\toprule[1.5pt]
Method         & Joint  & Sep.  & C.-$\alpha$  & C.-FA  & CAL & Joint  & Sep.  & C.-$\alpha$  & C.-FA  & CAL & Joint  & Sep.  & C.-$\alpha$  & C.-FA  & CAL & Joint  & Sep.  & C.-$\alpha$  & C.-FA  & CAL  & Joint  & Sep.  & C.-$\alpha$  & C.-FA  & CAL  \\ \midrule
Query & \multicolumn{5}{c|}{+Random} & \multicolumn{5}{c|}{+Margin} & \multicolumn{5}{c|}{+BADGE} & \multicolumn{5}{c|}{+Cluster-Margin} & \multicolumn{5}{c}{+Energy} \\ \midrule
Round 0       &2.7&2.7&2.71&4.05&1.68&2.7&2.7&2.64&4.05&1.68&2.7&2.7&2.71&4.05&1.68&2.7&2.7&2.64&4.05&1.68&2.7&2.7&2.56&4.05         &1.67\\ 
Round 1       &4.01&2.12&2.48&2.75&1.51&2.76&2.4&2.48&3.15&7.09&2.85&3.45&3.92&0.79&3.09&5.06&3.17&3.12&1.97&6.16&1.91&3.88&2.51&4.82&4.42\\ 
Round 2       &4.57&0.35&0.17&1.3&1.09&1.02&4.55&2.34&1.62&3.19&3.16&4.79&3.67&1.25&2.38&2.39&0.87&5.44&1.19&2.38&2.13&1.62&2.13&0.69&2.02\\ 
Round 3       &2.77&2.51&3.0&2.65&3.73&1.02&2.31&1.87&1.19&2.67&0.84&4.34&3.14&0.86&1.89&1.25&1.12&5.06&2.25&2.51&2.05&1.06&0.85&1.03&0.93\\ 
Round 4       &1.12&2.29&2.45&1.97&2.84&0.55&1.35&2.23&0.78&2.72&1.79&1.99&1.89&2.26&2.13&0.69&1.84&1.28&0.9&1.92&0.64&0.68&0.45&1.89&1.28\\ 
Round 5       &2.69&2.32&0.86&2.65&2.74&1.13&2.72&3.9&0.45&1.45&1.62&2.73&1.03&1.43&2.22&1.82&1.18&1.18&1.51&0.51&1.88&1.61&1.13&1.19&0.26\\  \midrule
Average &2.76&0.82&0.37&0.72&0.86&0.87&2.38&2.42&1.21&2.32&1.44&3.22&1.6&0.92&1.23&2.01&1.39&1.6&1.05&1.54&1.23&1.01&1.17&0.76       &0.99\\ 
\bottomrule[1.5pt]
\end{tabular}
\end{center}
\end{table*}

\begin{table*}[!t]
\setlength{\tabcolsep}{1.5pt}
\caption{Office-Caltech results ($\%$) for ``Joint + a strategy'', ``Separate + a strategy'' (Sep.), CAL-$\alpha$ (C.-$\alpha$), CAL-FA (C.-FA), and CAL. We mark the best results with \textbf{bold face} and the second-best results with \underline{underline}.}
\label{tab:office-caltech_mean}
\scriptsize
\begin{center}
\begin{tabular}{l|ccccc|ccccc|ccccc|ccccc|ccccc}
\toprule[1.5pt]
Method         & Joint  & Sep.  & C.-$\alpha$  & C.-FA  & CAL & Joint  & Sep.  & C.-$\alpha$  & C.-FA  & CAL & Joint  & Sep.  & C.-$\alpha$  & C.-FA  & CAL & Joint  & Sep.  & C.-$\alpha$  & C.-FA  & CAL  & Joint  & Sep.  & C.-$\alpha$  & C.-FA  & CAL  \\ \midrule
Query & \multicolumn{5}{c|}{+Random} & \multicolumn{5}{c|}{+Margin} & \multicolumn{5}{c|}{+BADGE} & \multicolumn{5}{c|}{+Cluster-Margin} & \multicolumn{5}{c}{+Energy} \\ \midrule
Round 0       &50.0&50.0&50.2&\textbf{54.8}&\underline{53.0}&50.0&50.0&50.2&\textbf{54.8}&\underline{53.0}&50.0&50.0&50.2&\textbf{54.8}&\underline{53.0}&50.0&50.0&50.2&\textbf{54.8}&\underline{54.2}&50.0&50.0&50.2&\textbf{54.8}&\underline{51.7}\\
Round 1       &73.7&77.6&\underline{78.6}&78.3&\textbf{81.4}&72.1&72.5&75.0&\underline{77.3}&\textbf{78.1}&74.5&68.8&72.3&\underline{75.4}&\textbf{78.7}&70.9&69.0&70.2&\underline{75.0}&\textbf{77.5}&70.5&74.5&71.6&\underline{78.8}&\textbf{81.1}\\
Round 2       &79.4&80.9&\underline{83.3}&82.5&\textbf{86.1}&79.0&81.7&\underline{84.8}&84.2&\textbf{88.3}&80.3&78.9&80.4&\underline{83.7}&\textbf{87.9}&77.5&81.1&82.6&\underline{83.2}&\textbf{87.4}&82.5&82.9&\underline{85.0}&81.9&\textbf{86.3}\\
Round 3       &85.4&85.3&87.0&\textbf{88.4}&\underline{88.2}&84.6&86.3&89.8&\underline{90.7}&\textbf{91.4}&86.8&87.7&89.1&\underline{91.9}&\textbf{92.5}&83.7&\underline{87.8}&85.1&87.8&\textbf{93.2}&88.1&87.9&89.7&\underline{90.8}&\textbf{91.6}\\
Round 4       &\textbf{89.3}&88.4&88.7&88.7&\underline{89.1}&89.6&91.5&\underline{92.6}&92.4&\textbf{93.4}&89.0&91.7&91.4&\textbf{92.6}&\underline{92.5}&88.6&90.4&89.4&\underline{91.7}&\textbf{92.0}&90.4&90.3&\underline{92.2}&91.5&\textbf{92.3}\\
Round 5       &90.6&90.4&89.7&\underline{92.9}&\textbf{93.0}&90.3&91.0&91.3&\underline{93.3}&\textbf{93.6}&89.5&92.4&91.2&\underline{93.6}&\textbf{93.9}&\underline{91.3}&89.3&91.3&90.8&\textbf{94.3}&87.7&91.0&92.3&\underline{93.7}&\textbf{94.1}\\ \midrule
Average &78.1&78.8&79.6&\underline{81.0}&\textbf{81.8}&77.6&78.8&80.6&\underline{82.1}&\textbf{83.0}&78.3&78.3&79.1&\underline{82.0}&\textbf{83.1}&77.0&77.9&78.1&\underline{80.5}&\textbf{83.1}&78.2&79.4&80.2&\underline{81.9}&\textbf{82.9}\\ \midrule
\bottomrule[1.5pt]
\end{tabular}
\end{center}
\end{table*}

\section{{More Clarification}} \label{sec:clarify}

\begin{table}[!t]
\setlength{\tabcolsep}{1.5pt}
\caption{Standard deviation of Office-Caltech results ($\%$) for ``Joint + a strategy'', ``Separate + a strategy'' (Sep.), CAL-$\alpha$ (C.-$\alpha$), CAL-FA (C.-FA), and CAL. We train these models using different seeds and report the standard deviation of the accuracy. }
\label{tab:office-caltech_std}
\scriptsize
\begin{center}
\begin{tabular}{l|ccccc|ccccc|ccccc|ccccc|ccccc}
\toprule[1.5pt]
Method         & Joint  & Sep.  & C.-$\alpha$  & C.-FA  & CAL & Joint  & Sep.  & C.-$\alpha$  & C.-FA  & CAL & Joint  & Sep.  & C.-$\alpha$  & C.-FA  & CAL & Joint  & Sep.  & C.-$\alpha$  & C.-FA  & CAL  & Joint  & Sep.  & C.-$\alpha$  & C.-FA  & CAL  \\ \midrule
Query & \multicolumn{5}{c|}{+Random} & \multicolumn{5}{c|}{+Margin} & \multicolumn{5}{c|}{+BADGE} & \multicolumn{5}{c|}{+Cluster-Margin} & \multicolumn{5}{c}{+Energy} \\ \midrule
Round 0       &4.04&4.04&4.17&4.47&0.93&4.04&4.04&4.17&4.47&0.93&4.04&4.04&4.17&4.47&0.93&4.04&4.04&4.17&4.47&3.66&4.04&4.04&4.17&4.47&4.23\\ 
Round 1       &1.54&1.88&2.33&1.97&1.11&3.55&2.13&2.55&3.27&3.51&2.91&3.42&4.54&3.14&2.66&3.05&0.44&8.16&5.13&5.95&3.96&4.05&2.16&6.35&1.27\\ 
Round 2       &1.53&2.1&2.02&4.23&2.44&2.3&2.06&1.25&4.2&1.72&4.31&0.58&1.21&2.15&0.8&1.95&1.24&3.0&4.99&3.26&2.19&2.7&2.0&2.85&3.33\\ 
Round 3       &2.79&1.02&0.38&2.06&2.29&3.4&3.92&1.42&2.08&1.44&1.09&1.25&0.49&0.51&1.25&1.37&2.13&3.74&3.23&1.04&2.19&1.46&2.01&0.74&2.23\\ 
Round 4       &0.61&2.74&0.77&2.3&2.54&2.16&1.22&0.82&1.03&0.36&1.28&1.03&1.23&0.51&0.82&0.74&2.27&3.17&0.63&0.04&2.0&0.62&0.53&1.39&1.02\\ 
Round 5       &1.63&0.85&1.54&0.04&1.67&1.64&0.16&0.41&0.57&1.33&0.68&0.57&0.24&0.27&0.43&1.45&0.65&0.95&0.87&0.93&1.47&0.6&0.93&1.21&0.42\\  \midrule
Average &1.7&0.65&0.44&1.24&1.37&2.7&1.74&0.51&2.54&0.7&1.8&0.64&1.3&0.5&0.75&0.16&1.38&3.69&2.48&2.16&2.0&1.1&1.48&1.74&1.84\\ 
\bottomrule[1.5pt]
\end{tabular}
\end{center}
\end{table}

\subsection{Our method: Composite Active Learning (CAL)} \label{sec:clarify_cal}


\textbf{Domain-Level Selection Is Agnostic to Instance-Level Selection and Model Architectures.} Training procedure of CAL is summarized in~\algref{alg:cadol}. 
In CAL, we consider it an advantage that domain-level selection is agnostic to, and therefore compatible with, instance-level selection methods, i.e., query strategies such as Random \cite{al_survey}, Margin \cite{margin}, and BADGE \cite{badge}. Specifically, note that CAL works in two independent steps:
\begin{compactitem}
\item In the domain-level selection, CAL jointly learns the encoder and estimates similarity weights $\alpha$ to reduce domain gaps between original domains and surrogate domains. The similarity weights $\alpha$ will determine the label budget for each domain in the query instance-level selection.
\item Then, in the instance-level selection, CAL leverages a certain query strategy, e.g., BADGE \cite{badge}, to select informative samples from each domain according to the estimated label budget from the domain level selection.
\end{compactitem}

Moreover, due to the independence between these two steps, query strategies are also agnostic to the architectures of deep learning models trained in the domain-level step. In other words, \emph{query strategies, e.g., BADGE \cite{badge}, can be used in the same way either with or without an encoder} since query strategies mainly utilize output or gradient embeddings of the model’s last layer, which is \emph{after} the encoder. Specifically, for uncertainty-based methods such as Random \cite{al_survey} and Margin \cite{margin}, they use the final output to estimate the uncertainty of each sample. For BADGE, uncertainty is measured as gradient magnitude with respect to parameters of the final (output) layer (see Page 2 and 3 in \cite{badge} for details). 

\textbf{Multi-Domain Training and Multi-Domain Testing}. Regarding the reasonableness of our "multi-domain training and multi-domain testing" setting, note that:
\begin{compactitem}
\item The problem of  "multi-domain training and single-domain testing" is a subset of our "multi-domain training and multi-domain testing". This is because "multi-domain training and multi-domain testing" aims to improve the performance of \emph{any} single domain contained in the "multi-domain". Therefore "multi-domain training and multi-domain testing" is a more challenging problem of great significance in the general ML community.
\item "Multi-domain training and multi-domain testing" is more natural and reasonable than  "multi-domain training and single-domain testing" under the active learning settings. The reason is as follows. Our proposed method is based on active learning settings, where both the training set and testing set are randomly selected from a total set of all data $T$. Therefore,  if $T$ contains multiple domains, training and testing should be conducted on multi-domain data. Accordingly, in our \emph{multi-domain} active learning setting, algorithms should also be trained and tested on \emph{multiple domains}.
\end{compactitem}
Therefore, our setting is natural. Typical active learning focuses on a \emph{single-domain} setting, where algorithms select samples from the \emph{single domain} to label, hoping to maximize the average accuracy in that \emph{single domain}. In our multi-domain active learning setting, algorithms select samples from \emph{multiple domains} to label, hoping to maximize the average accuracy over \emph{all domains}. Such a multi-domain setting is prevalent in practice. For example, to train an object recognition model that detects and classifies wildlife animals in different environments, where images from each environment constitute one domain, one needs to carefully decide how to spend the labeling budget among the different domains to achieve the highest average accuracy across all domains. In such cases, it is sub-optimal to directly perform single-domain active learning separately for each domain. Please kindly refer to Sec. 1 of the paper for more details and motivation.

\subsection{More Clarification on Related Work}
\textbf{Multi-Source Domain Adaptation (MSDA) and Multi-Source Domain Generalization (MSDG).} Note that our multi-domain active learning setting is \emph{very different} from MSDA and MSDG. Below are some key differences: 
\begin{compactitem}
\item Multi-source domain adaptation/generalization distinguishes between source domains and the target domain, while multi-domain active learning does not. 
\item Multi-source domain adaptation/generalization assumes access to\emph{all labels} in the source domains even from the beginning, while multi-domain active learning starts with \emph{only unlabeled data in all domains} (except that in active learning’s initial round, one would randomly sample a few data points to label). 
\item Multi-source domain adaptation/generalization aims to improve performance (e.g., accuracy) only on the \emph{target domain}, while multi-domain active learning aims to improve the average performance of \emph{all domains}. 
\end{compactitem}
It is also worth noting that blindly adapting multi-source domain adaptation/generalization for the active learning setting will not work and may even hurt performance (see \tabref{tab:ada_baseline} for more empirical results and details).

\textbf{Multi-Task Learning.} Note that we focus on a  multi-domain setting rather than a multi-task setting. In our setting, different domains share the same task (e.g., in Office-Home, different domains share the same label space of 65 classes); therefore, one single classifier would suffice.






\end{document}